\documentclass[runningheads]{llncs}
\usepackage{eccv}
\usepackage{eccvabbrv}

\usepackage{graphicx}
\usepackage{booktabs}
\usepackage{multirow}
\usepackage{tcolorbox}

\usepackage[accsupp]{axessibility}  

\usepackage[colorlinks]{hyperref}

\usepackage{orcidlink}

\newcommand{\ourwork}{AID-AppEAL}
\newcommand{\IPA}{ICAA}

\definecolor{darkgreen}{RGB}{30,150,30}
\definecolor{darkblue}{RGB}{0,0,127}
\definecolor{darkyellow}{RGB}{171,133,0}
\definecolor{darkred}{RGB}{180,20,20}
\definecolor{darkmagenta}{RGB}{127,0,127}
\definecolor{darkcyan}{RGB}{0,127,127}

\newcommand{\set} [1] {\mathbb{#1}}

\begin{document}

\title{AID-AppEAL: \underline{A}utomatic \underline{I}mage \underline{D}ataset and Algorithm for Content \underline{App}eal \underline{E}nhancement\\and \underline{A}ssessment \underline{L}abeling} 

\titlerunning{AID-AppEAL}

\author{Sherry X. Chen\inst{1}\thanks{Corresponding author email: xchen774@ucsb.edu}\orcidlink{0000-0002-4964-5286} \and Yaron Vaxman\inst{2}\orcidlink{0009-0000-7804-0473} \and Elad Ben Baruch\inst{2}\orcidlink{0009-0004-6237-5526} \and David Asulin\inst{2}\orcidlink{0009-0000-3647-795X} \and \\Aviad Moreshet\inst{2}\orcidlink{0009-0003-1956-296X} \and Misha Sra\inst{1}\orcidlink{0000-0001-8154-8518} \and Pradeep Sen\inst{1}\orcidlink{0000-0002-8042-924X}}

\authorrunning{S.Chen et al.}

\institute{University of California, Santa Barbara \and Cloudinary}

\maketitle

\begin{abstract}

We propose Image Content Appeal Assessment (\IPA{}), a novel metric that quantifies the level of positive interest an image's content generates for viewers, such as the appeal of food in a photograph. This is fundamentally different from traditional Image-Aesthetics Assessment (IAA), which judges an image's artistic quality. While previous studies often confuse the concepts of ``aesthetics'' and ``appeal,'' our work addresses this by being the first to study \IPA{} explicitly. To do this, we propose a novel system that automates dataset creation and implements algorithms to estimate and boost content appeal. We use our pipeline to generate two large-scale datasets (70K+ images each) in diverse domains (food and room interior design) to train our models, which revealed little correlation between content appeal and aesthetics. Our user study, with more than 76\% of participants preferring the appeal-enhanced images, confirms that our appeal ratings accurately reflect user preferences, establishing \IPA{} as a unique evaluative criterion. Our code and datasets are available at \href{https://github.com/SherryXTChen/AID-Appeal}{https://github.com/SherryXTChen/AID-Appeal}.

\keywords{image assessment \and automated dataset creation \and\\ image manipulation}
\end{abstract}
\section{Introduction}

\begin{figure}
    \centering
\begin{tabular}{l@{}c@{}c@{}c@{\hspace{0.02in}}|@{\hspace{0.02in}}c@{}c@{}c@{}}
 & \includegraphics[width=0.1\linewidth]{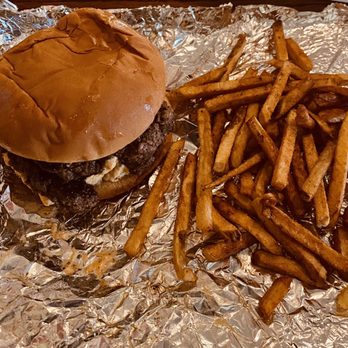} & \includegraphics[width=0.15\linewidth]{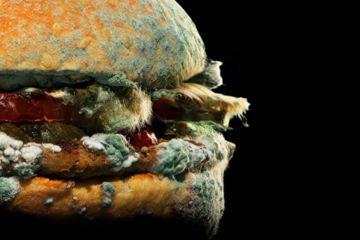} & \includegraphics[width=0.15\linewidth]{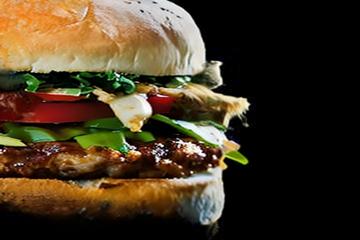} & \includegraphics[width=0.135\linewidth]{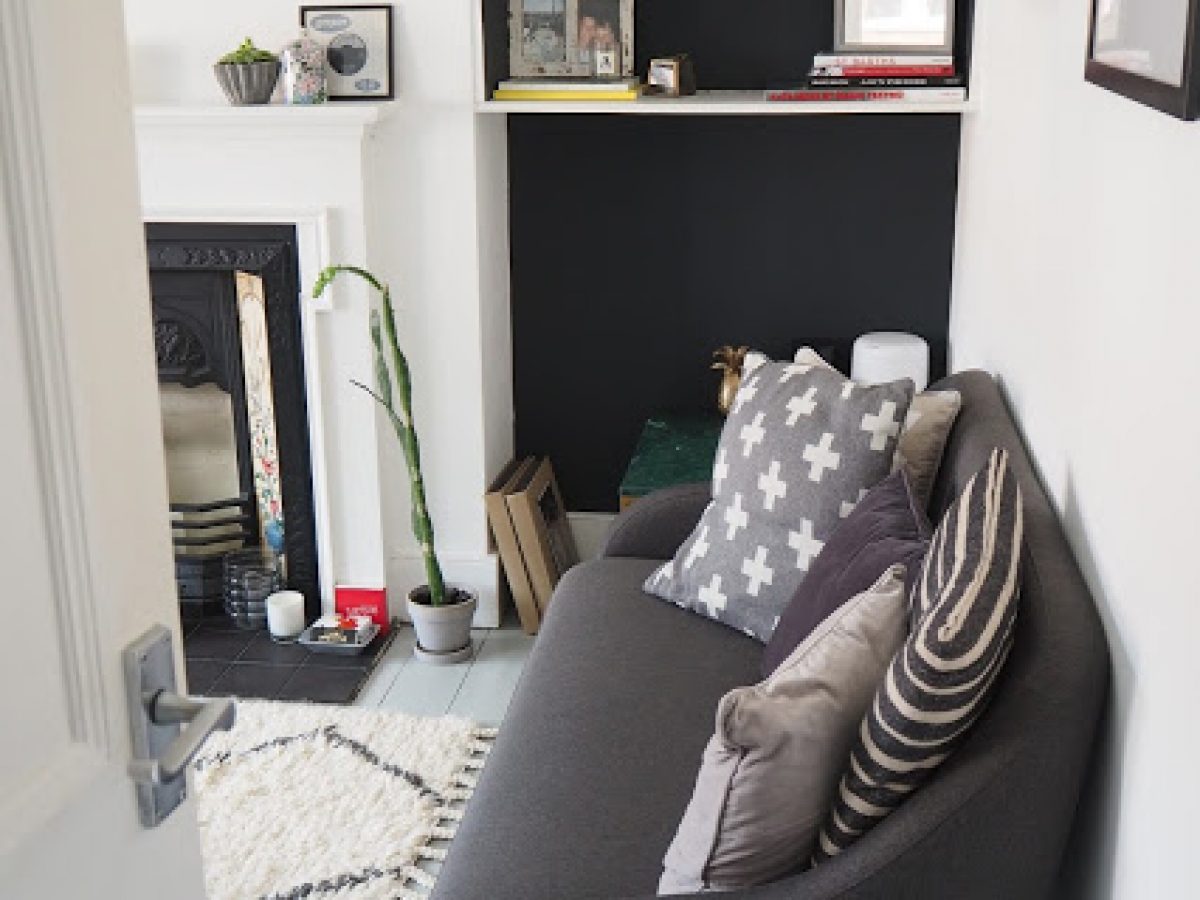} & \includegraphics[width=0.16\linewidth]{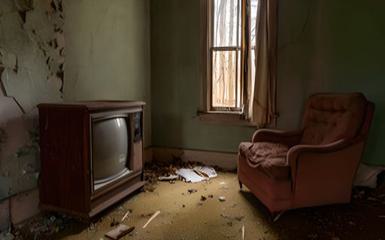} & \includegraphics[width=0.16\linewidth]{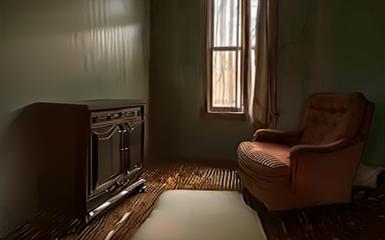} \\
DIAA & \underline{0.63} & 0.65 & \textbf{0.68} & \underline{0.34} & \textbf{0.42} & 0.36 \\
MPADA & \underline{0.44} & \textbf{0.61} & 0.49 & \underline{0.48} & 0.70 & \textbf{0.74} \\
NIMA & \underline{5.51} & 6.13 & \textbf{6.27} & \underline{5.68} & 5.95 & \textbf{5.98}  \\
Ours & 7.80 & \underline{5.52} & \textbf{8.08} (+2.56) & \textbf{7.70} & \underline{3.51} & 6.64 (+3.13) \\
\end{tabular}

\captionof{figure}{\textbf{Image-content appeal assessment (\IPA{}) and enhancement.} The $1^{st}$/$4^{th}$ columns show amateur photos lacking artistic appeal, while the $2^{nd}$/$5^{th}$ columns feature professionally taken images of less appealing content (a moldy burger and a dirty room). Because of their superior aesthetics, IAA baselines (DIAA~\cite{kong2016photo}, MPADA~\cite{sheng2018attention}, and NIMA~\cite{talebi2018nima}) rate them higher even though they have less appealing content (lowest scores underlined, highest in bold), while our \IPA{} estimator accurately assesses and enhances content appeal ($2^{nd}$/$5^{th}$ to $3^{rd}$/$6^{th}$ columns).
}

\label{fig:teaser}
\end{figure}

The accurate measure of perceptual image quality is an important problem in computer vision, since algorithms must often account for how humans actually perceive images. For this reason, researchers have developed algorithms focusing on distinct facets of image quality. For example, image-quality assessment (IQA) algorithms aim to estimate the perceptual impact of distortions~\cite{xue2013gradient, nafchi2016mean, prashnani2018pieapp, li2020norm, fang2020perceptual, ying2020patches, hosu2020koniq, lao2022attentions, he2021latentkeypointgan, ke2021musiq}, while image aesthetics assessment (IAA) evaluates an image's aesthetics based on principles of art and photography~\cite{lu2014rapid, ma2017lamp, Ren_2017_ICCV, sheng2018attention, talebi2018nima, hosu2019effective, Zhu2020Personalized, zhang2021image}.

Beyond IQA and IAA, we identify a critical yet overlooked aspect of perceptual image quality: image-content appeal assessment (\IPA{}). This concept becomes evident when comparing professional photographs that are highly aesthetic (\cref{fig:teaser}, Cols.2,5) but feature unappealing subjects (a moldy burger and a dirty room). As these images receive high scores from existing IAA methods, including some designed to measure ``image appeal,'' it underscores the need to evaluate a quality fundamentally different from existing metrics. We call it ``image content appeal.''

To devise a formal definition of image-content appeal, we take inspiration from the photography literature, where image appeal is defined as ``the interest that a picture generates when viewed by third-party observers''~\cite{savakis2000evaluation}. Our focus, however, shifts from the image itself to the content it portrays, emphasizing the amount of \textbf{positive interest in the content of a picture when viewed by generic third-party observers}. This distinction allows us to assess how much a viewer might desire to engage with the content, such as eating the food shown in the picture or staying in a depicted room. A metric like this would benefit sectors like food services, online retail, and vacation rentals, for example.

Image-content appeal assessment (\IPA{}) emerges as a compelling research avenue with significant practical implications. However, the absence of dedicated datasets for \IPA{} research presents a challenge, as existing image aesthetics assessment (IAA) datasets only broadly cover ``interesting content''~\cite{kong2016aesthetics,kong2016photo,hosu2019effective,Zhu2020Personalized} or ``interest-ness''~\cite{dhar2011high,lu2014rapid,yang2022personalized}, not specifically targeting the positive interest \IPA{} focuses on. Another option is to create our own \IPA{} dataset, but manually annotating large image assessment datasets (IQA~\cite{sheikh2006statistical, ponomarenko2009tid2008, ponomarenko2013color, ghadiyaram2015massive, ma2016waterloo, jinjin2020pipal, hosu2020koniq, fang2020perceptual, ying2020patches}; IAA~\cite{datta2008algorithmic, murray2012ava, Ren_2017_ICCV, chang2017aesthetic, he2022rethinking, yang2022personalized}) can become an expensive and time-consuming bottleneck. 

To bridge this gap, we present \ourwork{}, an automated dataset generation pipeline as well as algorithms for estimating and enhancing content appeal. We used our system to generate two large-scale datasets (food and room interior design), totaling over 70,000 images each, which enable the training of specialized content appeal estimators and enhancers. Our content appeal scores show little correlation with traditional aesthetics scores, underscoring the distinct nature of \IPA{}. User studies further validate our approach, with over 76\% of participants favoring the appeal-enhanced images, affirming the effectiveness of our system in accurately capturing and enhancing image-content appeal.

In summary, the main contributions of our work are:
\begin{enumerate}
    \item Recognition of image-content appeal (\IPA{}) as distinct from traditional image-aesthetic and appeal assessments that have been previously studied
    \item Development of a universal automated \IPA{} dataset creation pipeline
    \item Creation of two domain-specific \IPA{} datasets
    \item Introduction of accurate \IPA{} estimators for each of the two datasets
    \item Implementation of content appeal enhancers for each dataset, improving \IPA{} while maintaining visual integrity, validated by a user study.
\end{enumerate}

\section{Related Work\label{sec:related_work}}

\subsection{Image aesthetics and content appeal assessment}

Previous research in image aesthetics and appeal, collectively termed image aesthetic assessment (IAA), aims to evaluate an image's quality and attractiveness. In those works, the term ``aesthetic appeal'' has been used to denote ``the subjective notions of ``beauty'' in the image~\cite{Redi2013CrowdsourcingbasedMS}, ``what makes an image aesthetically pleasing''~\cite{Siahaan2016ARM}, and ``appeal'' is the quality of the image ``being attractive or interesting''~\cite{goring2023image}. `Appeal'' in these contexts refers to how appealing images are from an artistic point of view.

Since different factors such as lighting, contrast, color harmony, and composition all play a role in assessing image aesthetics, prior IAA methods often design different branches in their neural networks that either take different crops of each image~\cite{lu2014rapid,ma2017lamp} or estimate various IAA attribute scores~\cite{Ren_2017_ICCV,sheng2018attention,zhang2021image}. Convolution neural networks (CNNs) followed by fully connected layers (FCs) are commonly architectural components used as the backbone of these algorithms~\cite{lu2014rapid, ma2017lamp, Ren_2017_ICCV, sheng2018attention, talebi2018nima, hosu2019effective, Zhu2020Personalized, zhang2021image}. Some work also adapts variations of pre-trained visual models to facilitate their tasks~\cite{talebi2018nima,Zhu2020Personalized,zhang2021image}.

\subsection{Image assessment dataset creation}

Creating a good dataset is often one of the most critical steps for image assessment research. In image-quality assessment (IQA), the goal is to evaluate the perceived technical quality of an image after it is distorted. IQA datasets can either be full-reference (FR-IQA) or no-reference (NR-IQA), depending on whether the original pristine reference images are available in the dataset. 

FR-IQA datasets are created from a set of pristine images, where various distortion operations are applied to create different distorted versions of them~\cite{sheikh2006statistical,ponomarenko2009tid2008,ponomarenko2013color,ma2016waterloo,prashnani2018pieapp,lin2019kadid,jinjin2020pipal}. While these datasets are mostly human-annotated, this dataset creation process does offer some leeway to annotate images automatically based on the distortion operations being applied. On the other hand, samples in these datasets are heavily correlated. Furthermore, the distortion may not fully reflect the characteristics of distorted image ``in the wild.''

In contrast, NR-IQA datasets contain distorted images where their pristine counterparts are unknown, often because we only have the final distorted images from the internet~\cite{ghadiyaram2015massive,hosu2020koniq,fang2020perceptual,ying2020patches}. As a result, these datasets are usually much larger, more diverse, and contain more realistic samples. On the other hand, extensive laboratory subjective study~\cite{fang2020perceptual} or crowdsourcing~\cite{ghadiyaram2015massive,hosu2020koniq,ying2020patches} is required for dataset annotation, which is expensive and time-consuming.

Another type of image assessment is image aesthetics assessment (IAA), which is concerned with human perception of beauty. Although this is largely a subjective manner, some prior work~\cite{datta2008algorithmic,murray2012ava,kong2016photo,chang2017aesthetic} annotated images by having multiple annotators labeling one image so ``the average score can be thought of as an estimator for its intrinsic aesthetic quality''~\cite{datta2008algorithmic}. On the other hand, more recent work~\cite{Ren_2017_ICCV, he2022rethinking, yang2022personalized} acknowledge the subjective nature of aesthetics assessment and focus on not just the average score, but scores from each annotator to study and estimate human personal preference.

\subsection{Image generation and appeal enhancement}
Both our \IPA{} dataset creation pipeline and our content appeal enhancement module are made possible due to recent advances in image generation, in particular with GAN-based models~\cite{liu2021towards,sauer2021projected,sauer2022stylegan,sauer2023stylegan}, diffusion-based models~\cite{rombach2021highresolution,dhariwal2021diffusion,nichol2021glide,brooks2022instructpix2pix, bar2022text2live, parmar2023zero}, and even some combinations of them in between~\cite{wang2022diffusion}. There is also an increasing interest in using these models to create training datasets~\cite{Shrivastava_2017_CVPR,brooks2022instructpix2pix,Carlini2023ExtractingTD,Bansal2023LeavingRT}.

Although there isn't any work dedicated specifically to content appeal enhancement to the best of our knowledge, there are a lot of image manipulation methods that have the potential to enable such applications, where including text-based image editing~\cite{brooks2022instructpix2pix, bar2022text2live, parmar2023zero}, local inpainting with additional user-defined mask~\cite{meng2022sdedit, Ramesh2022HierarchicalTI, avrahami2022blended}, invert and finetuning pre-trained models on an image-text pair and edit the text to enable localized editing~\cite{dhariwal2021diffusion, mokady2022null, SongME21}. Our content appeal enhancement method has borrowed components from these methods and opted to use an automated generated mask and textual inversion~\cite{gal2022textual} for image editing. 

\section{Dataset Creation Pipeline\label{sec:dataset_creation_pipeline}}

To develop effective \IPA{} algorithms, it's crucial to assemble a dataset that meets specific criteria: it should align with human perception of content appeal, span a broad spectrum of appeal levels and content variety to avoid overfitting, and include a significant volume of high-resolution images for robust machine learning model training. Our initial research indicated that consumer photos typically exhibit a limited range of appeal, often skewing towards the lower end. To counteract potential bias, we exclusively incorporated professional images to ensure both high aesthetic and quality standards, enabling our model to accurately assess content appeal. The trained models can also generalize to consumer-taken pictures, as shown in \cref{sec:experiments}.

Creating such a dataset poses substantial challenges, including the costly manual labeling process, the difficulty in accessing a large volume of high-quality professional images due to stock photo website restrictions, and the inherent bias towards appealing content on these platforms. For instance, a search for ``delicious burger'' yields hundreds of thousands of results on major stock image sites, whereas ``moldy burger'' returns vastly fewer, risking biased model training.

To address these issues, we introduce an automated pipeline for generating domain-specific \IPA{} datasets, which ensure domain consistency to maintain relevance (e.g., food, rooms, scenery). Our approach involves collecting a base set of domain-matching images from stock websites, processing them to highlight domain-specific elements, and creating a synthetic dataset through image manipulation to vary content appeal and background diversity. This dataset trains a relative content appeal comparator to label a vast collection of real images for the final dataset, facilitating the efficient creation of large-scale datasets (over 70K images per domain) without manual labeling. While our examples focus on food imagery, this methodology is adaptable to various image domains, as further detailed in the supplementary.

\subsection{Base image set collection and pre-processing}
\label{sec:synthetic_dataset_creation_prerequisites}

To construct a comprehensive \IPA{} dataset tailored to a specific application domain $D$ (e.g., food, room interiors, scenery, people), we start by gathering a modest collection of domain-specific real images to generate a synthetic dataset. This synthetic dataset is then utilized to train an automatic labeling system, which produces the final dataset.

\begin{figure*}[t]
    \centering
    \includegraphics[width=0.7\linewidth]{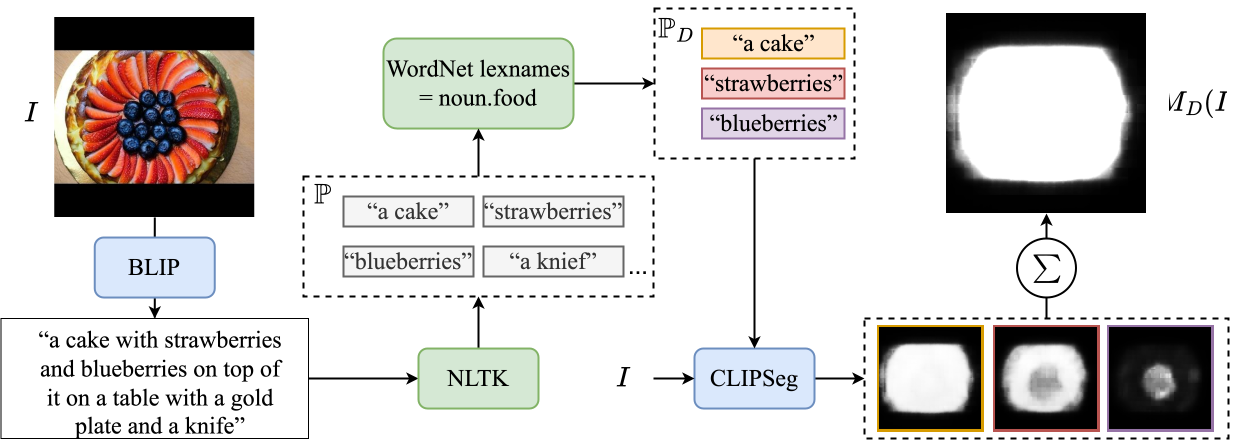}
    
    \caption{\textbf{Domain-relevancy map generation.} Given an image, we use BLIP~\cite{li2022blip} to estimate its description and extract all noun phrases $\set{P}$ using NLTK~\cite{bird2009natural}. For every phrase, we look up each of its words in WordNet~\cite{brown2005encyclopedia} to get their lexnames and keep the phrase if any of them matches the domain $D$ (e.g., if $D$ is food, then the phrase is kept only if at least one word's lexname is $noun.food$). The resulting set of phrases is $\set{P}_D$ and we use CLIPSeg~\cite{lueddecke22_cvpr} to create a segmentation map that locates objects described by each phrase in $\set{P}_D$. These maps collectively define the image region that contains objects from $D$, and we call it the domain-relevancy map.}
    
    \label{fig:image_relevancy_filter}
\end{figure*}

To automatically generate search queries for stock-image websites to retrieve suitable images, the process begins with defining a set of nouns $\set{N}_D$ that represent elements within the domain $D$, such as \{``burger,'' ``cake,'' ``fruit,'' \ldots \} for food. We then identify two sets of adjectives, $\set{A}_D^+$ or positive (appealing) descriptors (e.g., \{``delicious,''  ``gourmet,'' ``tasty,'' \ldots \} for food) and $\set{A}_D^-$ for negative (unappealing) descriptors (e.g., \{``disgusting,'' ``burnt,'' ``moldy,'' \ldots \} for food). Search queries are then created by randomly juxtaposing an adjective $a$ in either $\set{A}_D^+$ or $ \set{A}_D^-$ with a noun $n \in \set{N}_D$, creating a set of appealing search queries $\set{Q}_D^+$ and a set of unappealing search queries $\set{Q}_D^-$ respectively. 

Using $\set{Q}_D^+$ and $\set{Q}_D^-$, we gather low-resolution thumbnails from stock-image websites, to form positive $\set{I}_D^+$ and $\set{I}_D^-$ image sets respectively. Given the potential mismatch between the thumbnails and their search queries—particularly for unappealing images due to the scarcity of such content on stock websites—we select only the top matches from the search results to improve relevance.

To further refine the dataset and ensure its relevancy to the specific domain $D$, we implement a two-stage filtering process (\cref{fig:image_relevancy_filter}). Initially, we use the BLIP~\cite{li2022blip} to produce a text description for each image, discarding any image whose description does not contain words related to $D$. This relevancy check is performed by comparing each word in the image's description against WordNet~\cite{brown2005encyclopedia} to find their lexical categories. Images are retained only if at least one word from the description is classified under a lexname that matches the domain. This approach minimizes the inclusion of irrelevant/out-of-domain images, enhancing the dataset's quality and relevance to the application domain.

To ensure that objects from $D$ occupy enough space in the image to be relevant (an image of a room with a small apple is not a food image), we assess this by generating a domain-relevancy map to identify and measure the extent of domain-related objects within an image as follows (\cref{fig:image_relevancy_filter}).

The process starts by extracting noun phrases $\set{P}$ from the image's text description~\cite{bird2009natural}. From these, we identify phrases related to $D$ as $\set{P}_D$ and for each $p \in \set{P}_D$, we employ CLIPSeg~\cite{lueddecke22_cvpr} to segment the image $I$ based on these phrases, resulting in a map $\text{CLIPSeg}(I, p)$ that highlights domain-relevant objects. By aggregating these maps and normalizing the combined map, we obtain the domain-relevancy map $M_D(I)$. Image are discarded when the pixel value sum of $M_D(I)$ is less than $\gamma \cdot w_I \cdot h_I$, where $\gamma$ is a filtering threshold. To maintain dataset balance, we equalize the number of positive and negative images by removing excess ones from the larger subset.

Lastly, because all queried images are thumbnails and are fairly small (around $200 \times 200$), we apply ESRGAN~\cite{wang2018esrgan} to upscale and zero-pad images to make them a reasonable size ($512 \times 512$), making the final base image set $\set{I}_D$.

\subsection{Synthetic dataset image creation}
\label{sec:synthetic_dataset_creation}

\begin{figure*}[t]
\centering
\includegraphics[width=0.7\linewidth]{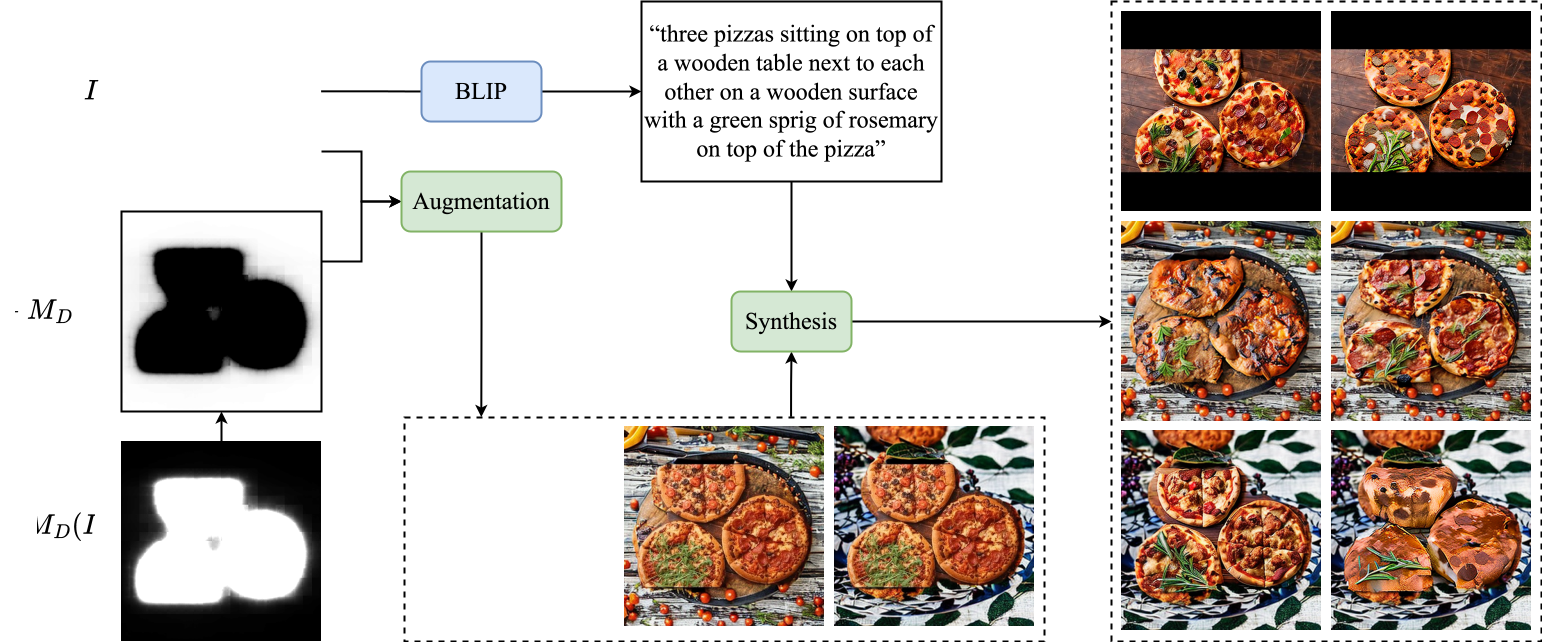}

\caption{\textbf{Synthetic dataset creation.} Given an image $I$, its text description, and its domain-relevancy map $M_D(I)$, we first locate ``background'' regions $1-M_D(I)$ that should have minimal effect on content appeal. The image is first augmented using Stable Diffusion~\cite{rombach2021highresolution}(\cref{eq:DiversifyingBackgroundSynthesis}). We then use Textual Inversion~\cite{gal2022textual} to generate appealing/unappealing-content embeddings, which can change image content appeal with respect to $M_D(I)$ (\cref{eq:VaryingAppealSynthesis}).}

\label{fig:synthesis_pipeline}
\end{figure*}

One limitation of $\set{I}_D$ is the binary nature of our search queries yield images at the extreme ends of content appeal without capturing the subtle variations in between that are essential for training an accurate appeal score estimator. 

To address this, we propose generating synthetic images with a nuanced spectrum of content appeal using a generative model like Stable Diffusion(SD)~\cite{rombach2021highresolution}. The process involves creating embeddings that encapsulate the characteristics of both appealing and unappealing content from our base image set, which can be linearly interpolated to produce all content appeal levels in between. 

We start by selecting images in $\set{I}_D$ that best represent the highest and lowest appeal levels by gathering the top search results for search queries since early search results tend to be most relevant to queries themselves. Selected images $\set{T}_D^+ \subset \set{I}_D^+$ and $\set{T}_D^- \subset \set{I}_D^-$  are used to get ``appealing'' and ``unappealing'' embeddings $z_D^+$ and $z_D^-$ with Textual Inversion~\cite{gal2022textual}, which capture the essence of content appeal at both ends of the spectrum. To represent any content appeal level between these two extremes we simply linearly blend these vectors $f(\alpha) = \alpha z_D^+ + (1 - \alpha) z_D^-$, where $\alpha \in [0, 1]$ controls the level of content appeal.

To adjust content appeal of an image $I$, we focus exclusively on the areas identified by the domain-relevancy map $M_D(I)$, leaving the background or non-domain content unchanged. This is crucial, especially in datasets where the subject's appeal, such as food on a table or incidental items in room interiors, should not be influenced by their surroundings. While background contexts can affect perception, we initially set aside these influences for simplicity. 

Specifically, we use the inpainting function of SD to adjust image content appeal. This function is denoted as as $\text{SD}(I, p, M, \text{seed}())$, which takes the input image $I$ and the text prompt $p$ to change masked region $M$ with randomization from the seed $\text{seed}()$. The adjusted image $I'$ with the intended content appeal level $\alpha \in [0, 1]$ is produced through the equation:
\begin{equation}
I' = \text{SD}(I, \text{BLIP}(I) + f(\alpha), M_D(I), \text{seed}()). \\
\label{eq:VaryingAppealSynthesis}
\end{equation}

To enrich our dataset with diverse background elements and mitigate the risk of overfitting our labeling algorithm, we employ the same inpainting function to freely generate any content for the background by using an empty string as the prompt, effectively allowing SD to introduce variability. The mask applied for this operation is the inverse of the domain-relevancy map ($1 - M_D(I)$), targeting the modification of non-domain areas in the image:
\begin{equation}
I'' = \text{SD}(I', \text{`` ''}, 1-M_D(I), \text{seed}()), \\
\label{eq:DiversifyingBackgroundSynthesis}
\end{equation}
where $I'$ is the image that has been adjusted previously for content appeal. 

While our procedural explanation first discusses adjusting content appeal and then background modification, the actual implementation first alters the background before the domain-specific content appeal. This sequence, depicted in our synthesis pipeline figure, is not expected to influence the outcome.

\subsection{Relative content appeal estimation and final dataset annotation}
\label{sec:relative_estimate_dataset_annotation}

To circumvent the laborious process of manually annotating a vast number of images, we propose employing an automatic labeling algorithm trained on synthetically generated data. This algorithm operates as a relative content appeal comparator, assessing the appeal difference between two images instead of determining an absolute appeal value, which demands a broader and more varied dataset for accurate estimation.

For this purpose, a synthetic dataset $\set{S}$ is created, producing $N$ variations for each base image with differing content appeal levels $\alpha$'s. and backgrounds, as previously outlined. We posit that the appeal difference between any two such variations correlates directly with the difference in their $\alpha$ values. Thus, for image $I'_1$ and $I'_2$ derived from the same original image $I$ with content appeal parameters $\alpha_1$ and $\alpha_2$, we assume their content appeal difference is $\hat{A}(I'_1, I'_2) = \alpha_1 - \alpha_2$.

\begin{figure}[t]
    \centering
    \subfloat[]{\includegraphics[width=0.5\linewidth]{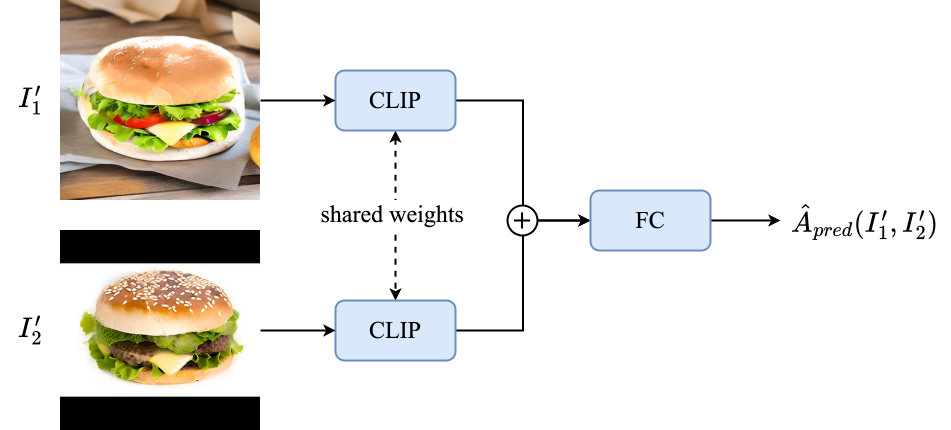}\label{fig:comparator_pipeline}}
    \hfill
     \subfloat[]{\includegraphics[width=0.4\linewidth]{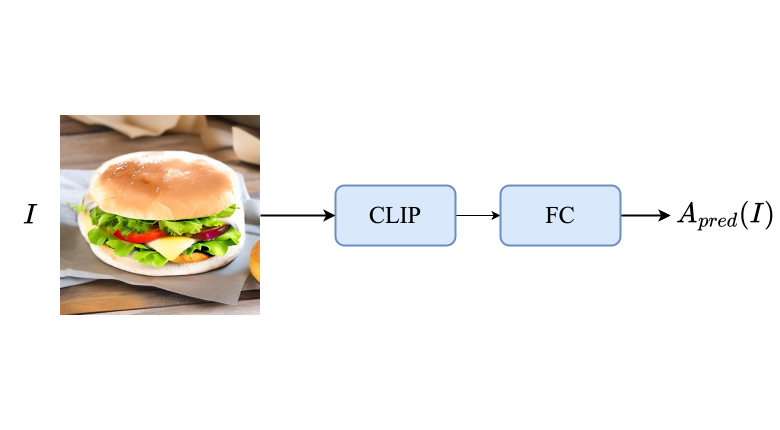}\label{fig:appeal_score_predictor}}
     
    \caption{\textbf{Relative and absolute content appeal estimation.} We use CLIP~\cite{radford2021learning} image encoder, followed by several fully connected (FC) layers to predict the image relative content appeal difference (\cref{fig:comparator_pipeline}) and absolute appeal (\cref{fig:appeal_score_predictor}).}
    
\end{figure}

A Siamese network architecture~\cite{prashnani2018pieapp, Lee2019ImageAA}, leveraging dual CLIP~\cite{radford2021learning} image encoders with shared weights for feature extraction, serves as our relative content appeal comparator (\cref{fig:comparator_pipeline}). This setup processes pairs of images, concatenates their features, and forwards these through fully connected layers to predict the appeal difference $\hat{A}_{pred}(I'_1, I'_2)$.  The network is trained to minimize the discrepancy between predicted and assumed appeal differences $|\hat{A}_{pred}(I'_1, I'_2) - \hat{A}(I'_1, I'_2)|$.

After training, this comparator is tasked with labeling a comprehensive set of real images $\set{I} = \set{I}_D^+ \cup \set{I}_D^-$ gathered from our initial queries, post domain-relevance filtering. To assign content appeal scores in the absence of absolute benchmarks, we employ a voting mechanism using a subset of exemplar images $\set{V}_D \subseteq \set{I}$ as reference points. Each image's appeal score is determined by averaging the comparator's outcomes against these exemplars, subsequently scaling these scores to a 1-10 range to represent the spectrum of content appeal. This methodology facilitates the creation of our final \IPA{} dataset.

\section{Absolute score estimation and enhancement\label{sec:image_appeal_enhancement}}

While the relative content appeal comparator has proven invaluable for dataset generation, our ultimate aim is to develop an estimator capable of assessing the content appeal of a single image in absolute terms. This leads to the creation of an absolute content appeal estimator, which evaluate the absolute content appeal of individual images $A(\cdot)$.

This estimator incorporates the same CLIP-based feature extraction mechanism used in the comparator, followed by a series of fully connected layers that culminate in a predicted content appeal score $A_{pred}(i)$. Training involves minimizing the discrepancy between the predicted appeal scores and the actual scores assigned during the dataset creation process $|A_{pred}(I) - A(I)|$.

An advantage of this absolute content appeal estimator is its utility in identifying and enhancing areas within an image that detract from its overall appeal (\cref{fig:teaser} Col. 3, 6). Instead of applying enhancement across the entire image, which risks altering already appealing or irrelevant regions, our goal is to specifically uplift areas deemed unappealing. A straightforward approach might involve applying a universal enhancement via Stable Diffusion, targeting maximum appeal. However, this method fails to discriminate between content that already meets or exceeds appeal thresholds and areas genuinely in need of improvement.

\begin{figure}

    \centering
    \includegraphics[width=0.5\columnwidth]{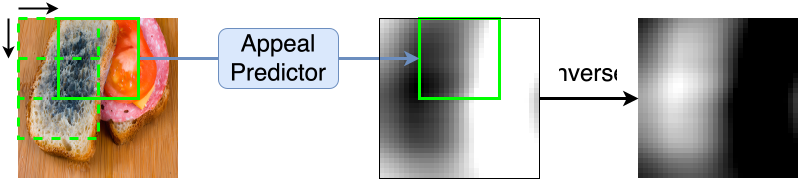}
    
    \caption{\textbf{Image content appeal heatmap generation.} We define a sliding window to capture overlapping patches of an image, where we use the content appeal estimator to estimate the content appeal score of each patch. The value of the heatmap for each pixel is averaged over all patches that include the pixel; we normalize all values and take their inverse, so a lighter color means the content in that region is more unappealing.}
    \label{fig:image_appeal_heatmap_generation}
    
\end{figure}

Although we can use $M_D(I)$ as a mask, this only resolves the first problem. So we generate a content appeal heatmap $M_D^H(I)$ that indicates the unappealing level at each image pixel (\cref{fig:image_appeal_heatmap_generation}) to control the location and magnitude of enhancement. Given image $I$, we define a window $w$ that slides over $I$ to get overlapping image patches $w(I)$ every $t$ pixels. The content appeal value of each pixel $p \in i$ is 
\begin{equation}
    \Bar{A}(p) = \text{mean}_{p \in w(i)}(A(w(I))),
\end{equation}
and the value of the image content appeal heatmap $M_D^H(I)$ for pixel $p$ is $1 - n(\Bar{A}(p))$, where $n(\cdot)$ normalizes all $\Bar{A}(p)$ to the range $[0, 1]$. Lastly, we enhance the content appeal of $I$ through
\begin{equation}
    \text{SD}(I, \text{BLIP}(I) +  z_D^+, M_D^H(I), \text{seed}()).
\end{equation}

\section{Experiments\label{sec:experiments}}

\subsection{Dataset creation}
To show \ourwork{} generalizes across different domains, we created two datasets with food and room interior images, both of which were automatically assigned appeal labels by our trained relative content appeal estimator:

\noindent \textsc{\textbf{Food}}: Search queries were generated from the following sets of words:
\begin{itemize}
    \item $\set{N}_F = \{$``burger,'' ``cake,'' ``chicken,'' ``cookie,'' ``food,'' ``rice,'' ``pizza,'' ``pasta,'' ``salad,'' ``steak,'' ``yogurt''$\}$
    \item $\set{A}_F^+ = \{$``delicious''$\}$
    \item $\set{A}_F^- = \{$``burnt,'' ``moldy,'' ``rotten''$\}$,
\end{itemize}
and use for \href{https://stock.adobe.com}{stock.adobe.com} and \href{https://shutterstock.com}{shutterstock.com}. We generated 18,000 images for $\set{S}_F$ and 78,917 images for $\set{I}_F$. 

\noindent \textsc{\textbf{Room}}: Search queries were generated from the following sets of words:
\begin{itemize}
    \item $\set{N}_R = \{`$`bathroom,'' ``bedroom,'' ``kitchen,'' ``living\ room,'' ``room''$\}$
    \item $\set{A}_R^+ = \{$``interior''$\}$
    \item $\set{A}_R^- = \{$``abandoned,'' ``dirty''$\}$.
\end{itemize}
and generated 15,000 images for $\set{S}_R$, 75,287 images for $\set{I}_R$.

\subsection{Model training}
We train our relative content appeal comparator in two stages. In the first stage, we freeze the CLIP backbone and train the comparator on $\set{S}_D$ for 10 epochs using PyTorch's AdamW optimizer with learning rate $1e^{-3}$ and batch size 16. In the second stage, we unfreeze the backbone and train the comparator for another 10 epochs with learning rate $1e^{-5}$. We then use the comparator to label $\set{I}_D$ and train the content appeal score estimator on it with the same training procedure as above. The final mean absolute error is $\text{MAE}_{i \in \set{I}_D}(|A_{pred}(I) - A(I)|) = 0.6756$ for the \textsc{Food} dataset and 0.6332 for the \textsc{Room} dataset. We use Stable Diffusion v2.1 inpainting with depth-guided ControlNet~\cite{zhang2023adding} for appeal enhancement.

\subsection{User study design}

\begin{figure}[t]
    \centering
    \subfloat[]{\includegraphics[width=0.45\linewidth]{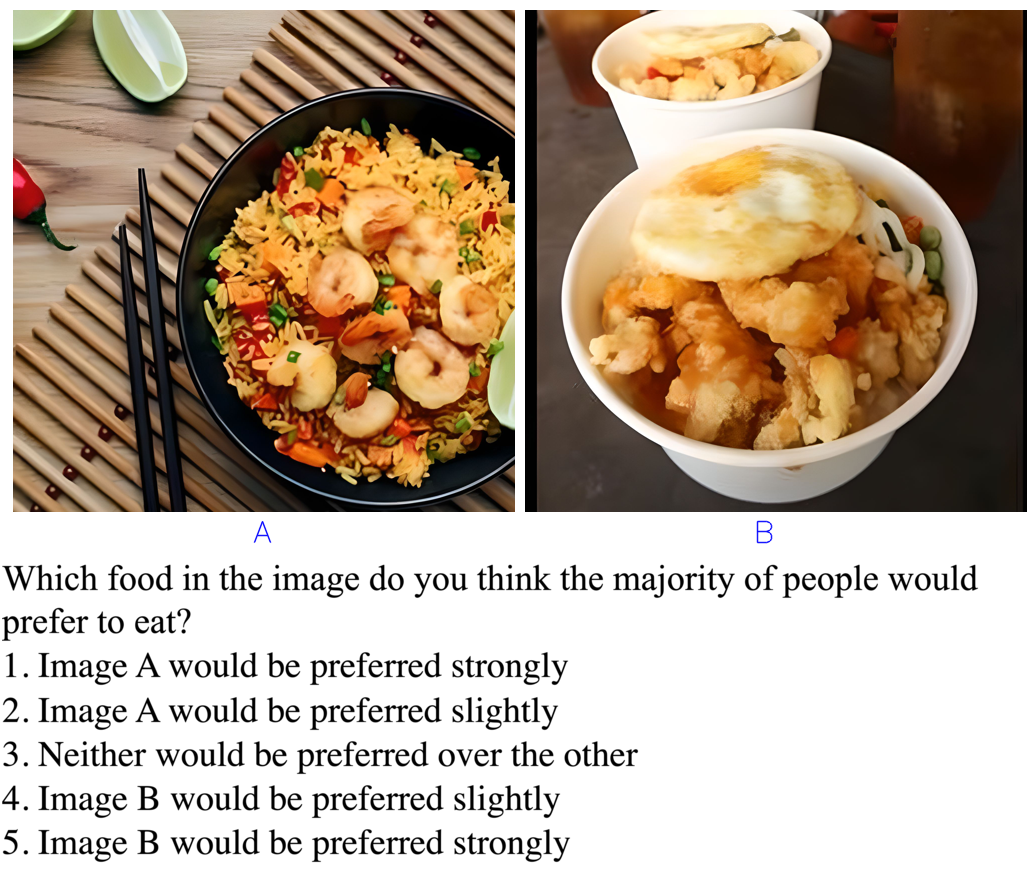}\label{fig:user_study_interface_food}}
    \hfill
     \subfloat[]{\includegraphics[width=0.45\linewidth]{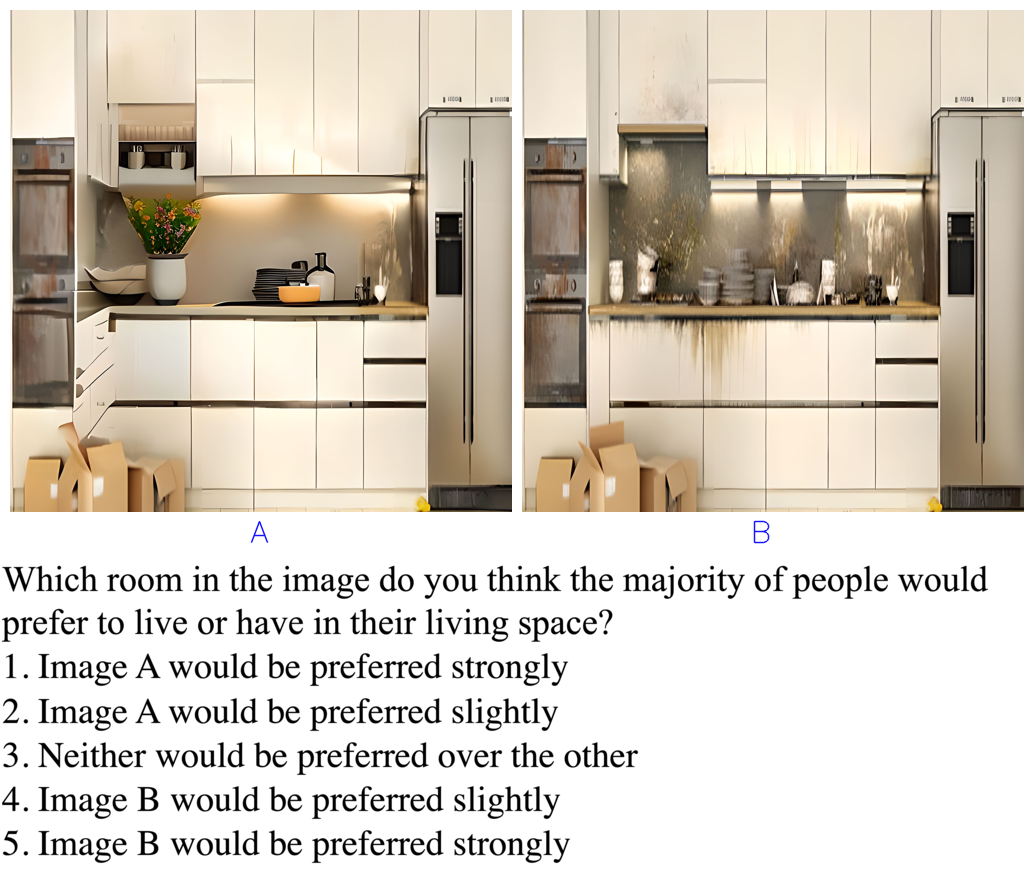}\label{fig:user_study_interface_room}\label{fig:user_study_interface_room}}
     
    \caption{\textbf{User study interface.} Participants are asked to answer questions by selecting one of the five options provided.}
    \label{fig:user_study_design}
    
\end{figure}

\begin{table}
\centering
\caption{\textbf{Quantitative evaluation between content appeal labels and three IAA baselines.} We evaluate the correlation coefficient between content appeal labels and three IAA baselines, and observe little to no correlation. RMSE metrics further suggests that our content appeal labels and IAA predictions are very different.}

\begin{tabular}{l|ccc|ccc}
\multirow{2}{*}{} & \multicolumn{3}{c|}{\textsc{Food}} & \multicolumn{3}{c}{\textsc{Room}} \\ \cline{2-7} 
 &  DIAA & MPADA & NIMA & DIAA & MPADA & NIMA \\ \hline
 PLCC & 0.168 & 0.005 & 0.01 & -0.123 & -0.012 & -0.147 \\
 SRCC & 0.162 & -0.015 & 0.003  & -0.121 & -0.017 & -0.149 \\
 KRCC & 0.109 & -0.009 & 0.002  & -0.082 & -0.013 & -0.098 \\
 RMSE & 6.463 & 6.711  & 2.009  & 6.262  & 5.899  &  1.79
\end{tabular}
\label{tab:appeal_vs_aesthetic}

\end{table}

We conducted a user study to validate the effectiveness of our content appeal estimator and enhancer by comparing them against human preference. We invited 28 volunteers (male = 14, female = 14, non-binary = 0), with ages ranging 18 - 44, to participate in our study. After providing informed consent (IRB protocol number \#anonymized), participants were asked to complete a survey on a computer screen in a lab setting.

The survey presents each user with pairs of images, each of which accompanied by a question tailored to the specific image domain: ``Which food in the image do you think the majority of the people would prefer \textbf{to eat}''  (\cref{fig:user_study_interface_food}) or ``Which room in the image do you think the majority of the people would prefer \textbf{to live} or have in their living space'' (\cref{fig:user_study_interface_room}), which are phrased to focus attention on the subject matter rather than the image as a whole and direct responses towards a collective preference, thereby reducing the impact of personal tastes, which we verify through supplementary analysis to have minimal influence on the outcomes.

To mitigate potential biases linked to cultural or personal predispositions towards certain subjects, we ensured that each image pair featured the same kind of domain-relevant object (e.g., fried rice in \cref{fig:user_study_interface_food} or a kitchen \cref{fig:user_study_interface_room}). The study comprised two sections, one for food and the other for room images, with a mixture of real image pairs and pairs consisting of a real image alongside its enhanced version, randomly selected from our datasets to cover a broad spectrum of content appeal levels.

Participants are asked to answer a total of 74 questions: 38 in the food section and 36 in the room section, which included both comparisons between real images and assessments of enhancements. The presentation order of questions and the left/right positioning of images were randomized for each participant to prevent any ordering effects. Through this methodology, we collected 2072 individual responses, providing a comprehensive dataset for evaluating our systems against human judgment. Further details and the results of this evaluation are documented in our supplementary materials.

\begin{figure}[t]
    \centering
    \captionsetup[subfigure]{labelformat=empty}
    \begin{subfigure}{0.45\linewidth}
        \centering
        \includegraphics[width=\linewidth]{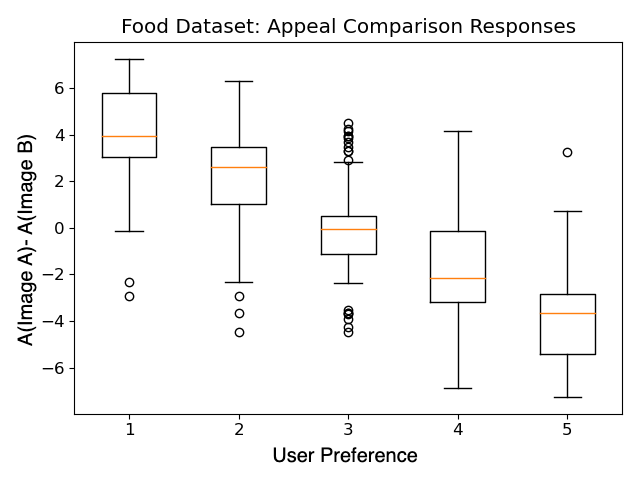}
    \end{subfigure}%
    \hfill
    \begin{subfigure}{0.45\linewidth}
        \centering
        \includegraphics[width=\linewidth]{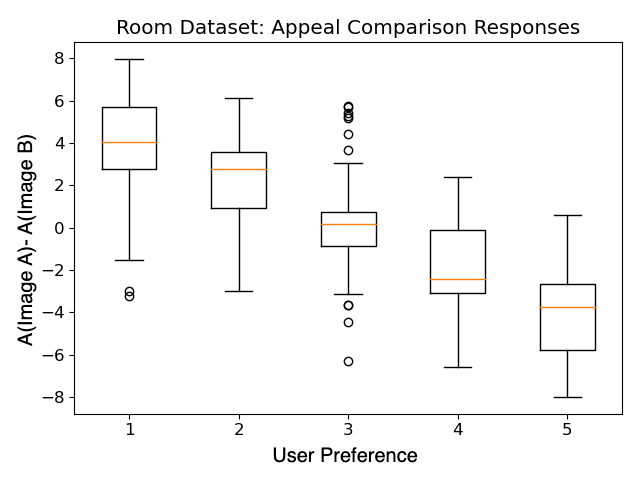}
    \end{subfigure}%
    \caption{\textbf{Content appeal labels versus user preferences.} Distribution of the difference in appeal labels \(A(\text{image A}) - A(\text{image B})\) for each preference option in our user study (see \cref{fig:user_study_design}). From response 1 to 5, we see a clear decrease in the mean of \(A(\text{image A}) - A(\text{image B})\): as people start preferring B over A more, image B also becomes more appealing.}
    \label{fig:appeal_compare_response}
\end{figure}

\begin{figure}[t]
    \centering
    \captionsetup[subfigure]{labelformat=empty}
    \includegraphics[width=0.5\linewidth]{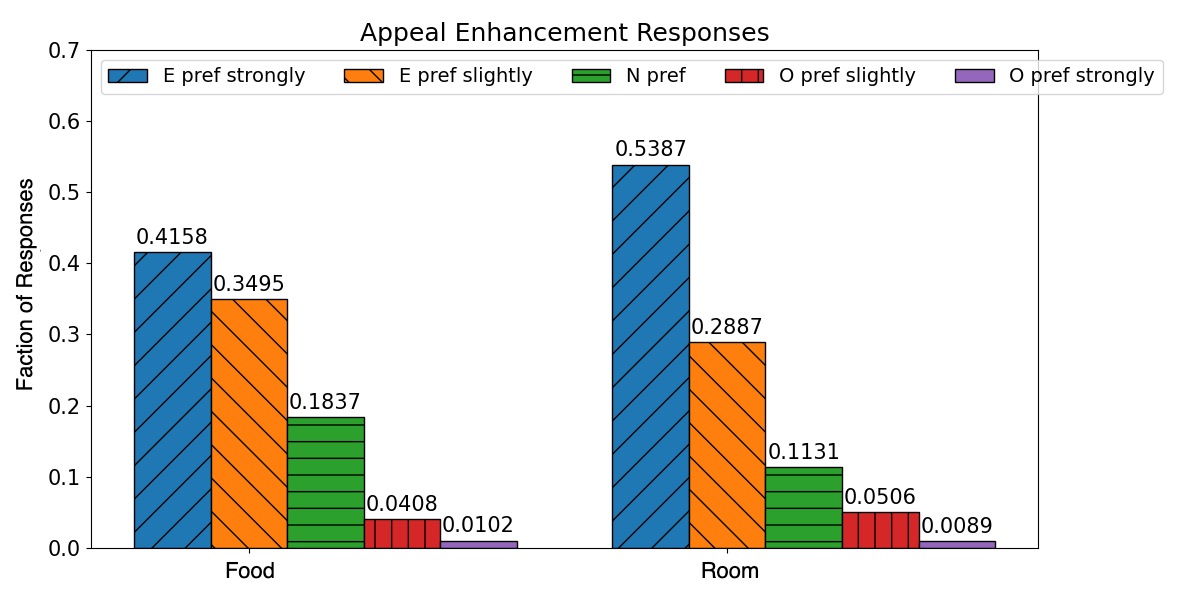}
    \caption{\textbf{Content appeal enhancement user responses.} Percentage of responses for each category, where $E$ represents the appeal-enhanced image, $O$ is the original image, $N$ is neither, and ``pref'' stands for ``preferred.'' We can see that 76.53\% and 82.74\% of the responses prefer the appeal enhanced images for the \textsc{Food} and \textsc{Room} dataset respectively.}
    \label{fig:appeal_enhancement_response}
\end{figure}

\subsection{IAA baseline comparison}

To show the difference between content appeal and image aesthetics, we uniformly stride every 1 out of every 100 images in each dataset and estimate their aesthetics scores with three popular, open-source IAA baselines~\cite{kong2016aesthetics,sheng2018attention,talebi2018nima}. We then compare to our appeal labels and observe little correlation as well as large value difference between the two (\cref{tab:appeal_vs_aesthetic}). Refer to the supplementary for more details.

\begin{figure*}[t]

    \centering
    \captionsetup[subfigure]{labelformat=empty}
    \begin{subfigure}{0.14\linewidth}
        \centering
        \caption{Input}
        \includegraphics[width=\linewidth]{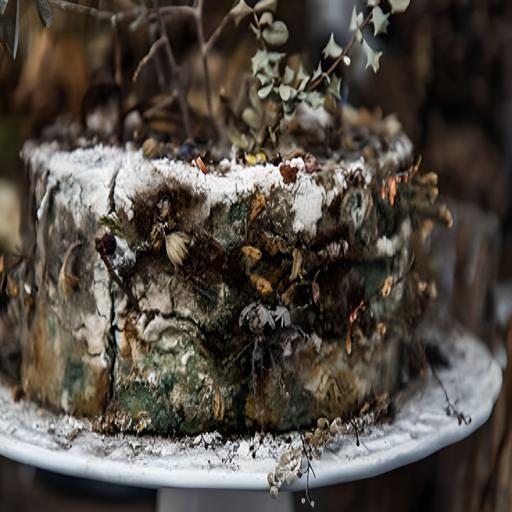}
    \end{subfigure}%
    \hfill
    \begin{subfigure}{0.14\linewidth}
        \centering
        \caption{N-TI}
        \includegraphics[width=\linewidth]{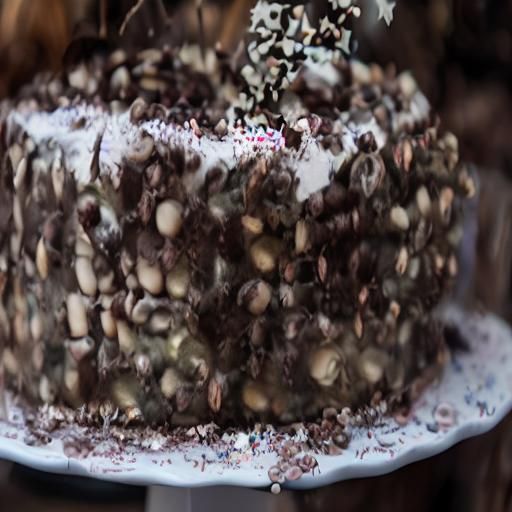}
    \end{subfigure}%
    \hfill
    \begin{subfigure}{0.14\linewidth}
        \centering
        \caption{P2P-0}
        \includegraphics[width=\linewidth]{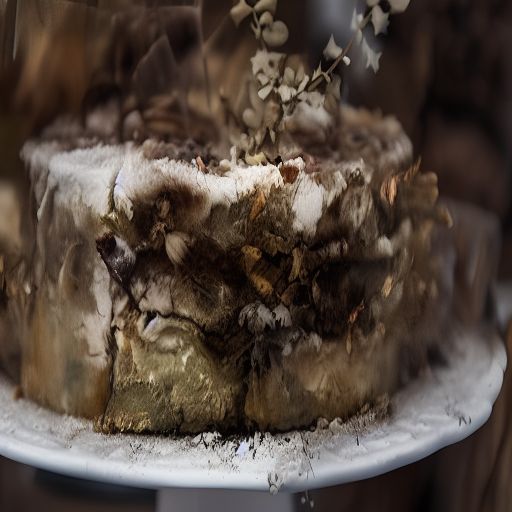}
    \end{subfigure}%
    \hfill
    \begin{subfigure}{0.14\linewidth}
        \centering
        \caption{T2L}
        \includegraphics[width=\linewidth]{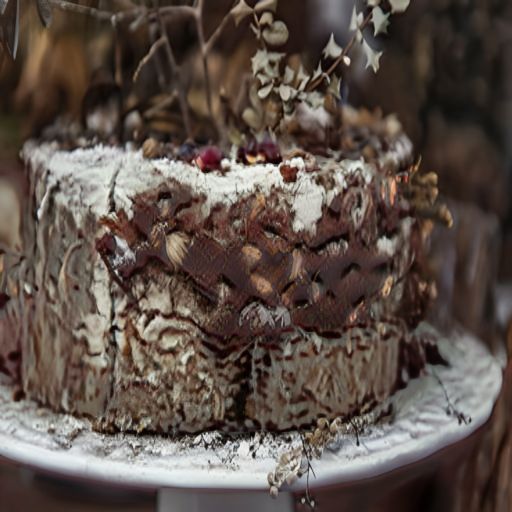}
    \end{subfigure}%
    \hfill
    \begin{subfigure}{0.14\linewidth}
        \centering
        \caption{IP2P}
        \includegraphics[width=\linewidth]{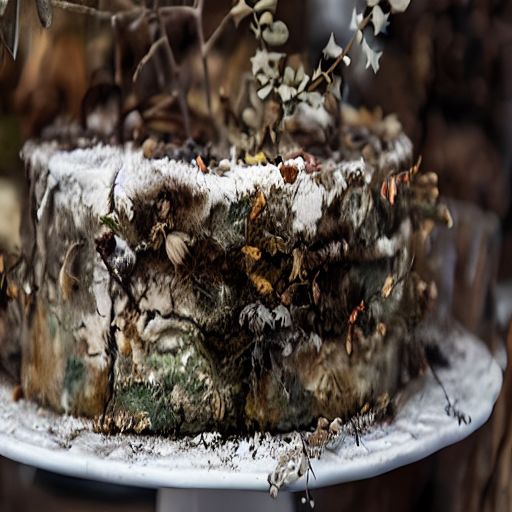}
    \end{subfigure}%
    \hfill
    \begin{subfigure}{0.14\linewidth}
        \centering
        \caption{Ours}
        \includegraphics[width=\linewidth]{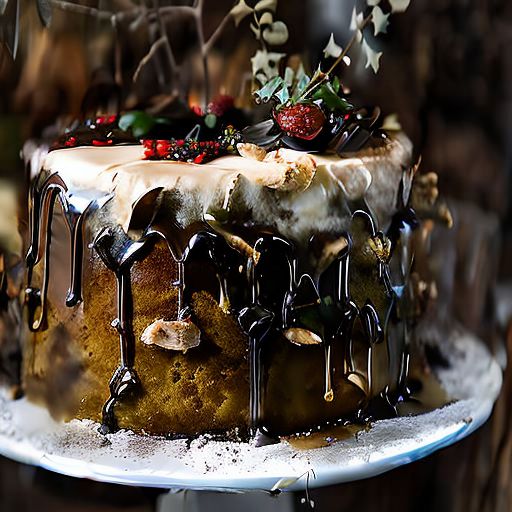}
    \end{subfigure}%
    \hfill
    \begin{subfigure}{0.14\linewidth}
        \centering
        \caption{Ours ($M_F^H$)}
        \includegraphics[width=\linewidth]{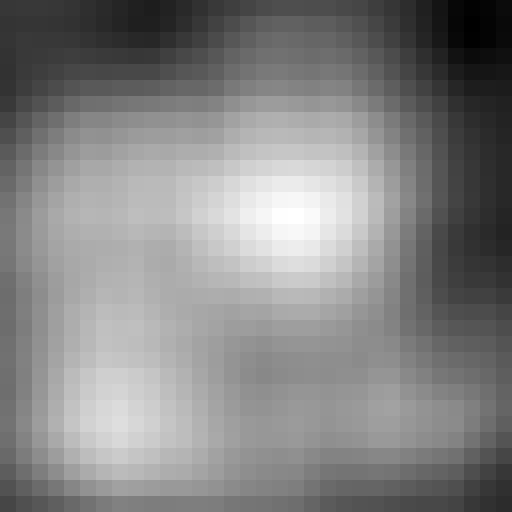}
    \end{subfigure}%

    \begin{subfigure}{0.14\linewidth}
        \centering
        \includegraphics[width=\linewidth]{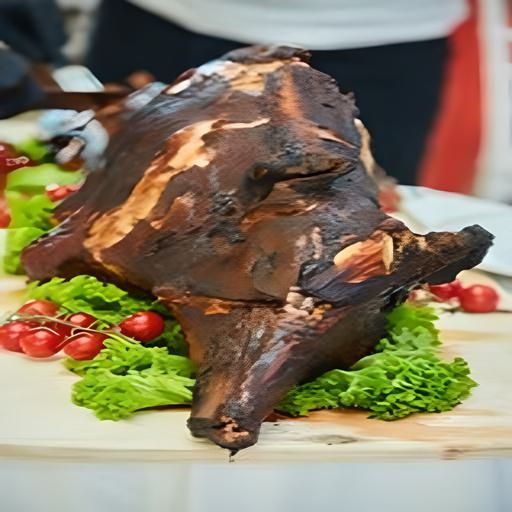}
    \end{subfigure}%
    \hfill
    \begin{subfigure}{0.14\linewidth}
        \centering
        \includegraphics[width=\linewidth]{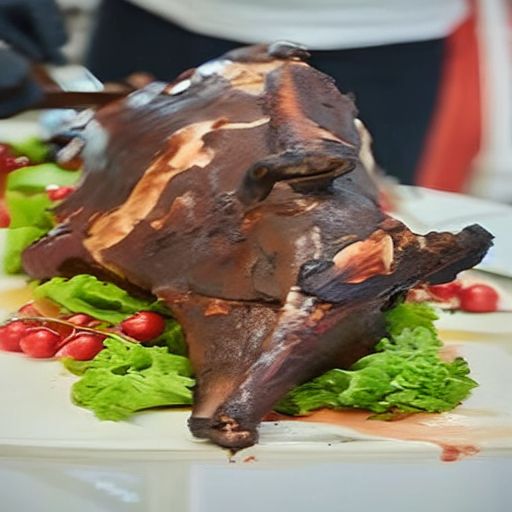}
    
    \end{subfigure}%
    \hfill
    \begin{subfigure}{0.14\linewidth}
        \centering
        \includegraphics[width=\linewidth]{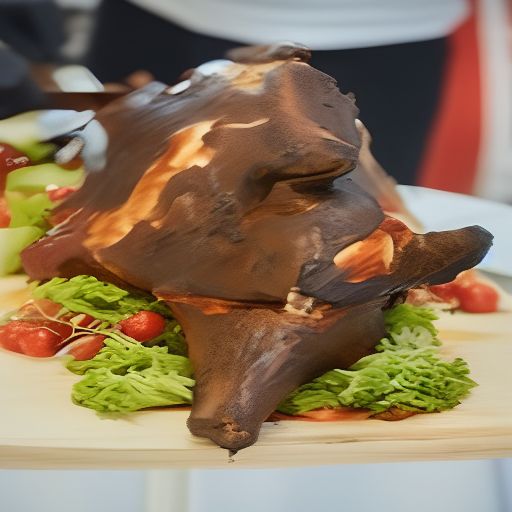}
    \end{subfigure}%
    \hfill
    \begin{subfigure}{0.14\linewidth}
        \centering
        \includegraphics[width=\linewidth]{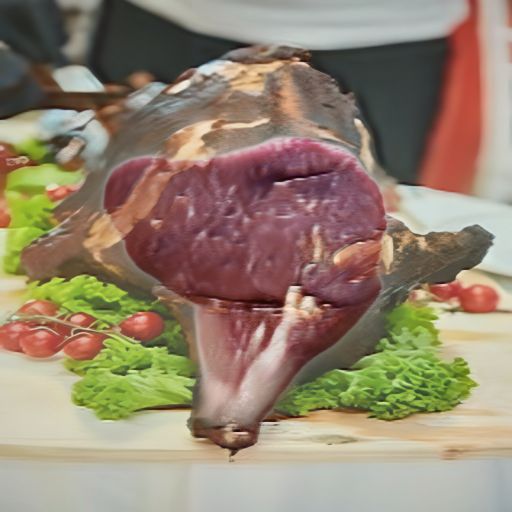}
    \end{subfigure}%
    \hfill
    \begin{subfigure}{0.14\linewidth}
        \centering
        \includegraphics[width=\linewidth]{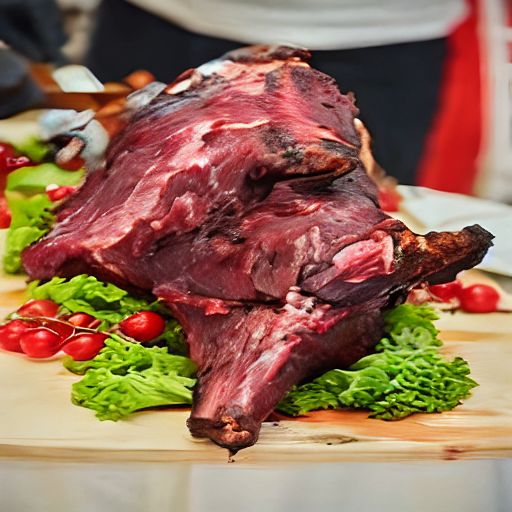}
    \end{subfigure}%
    \hfill
    \begin{subfigure}{0.14\linewidth}
        \centering
        \includegraphics[width=\linewidth]{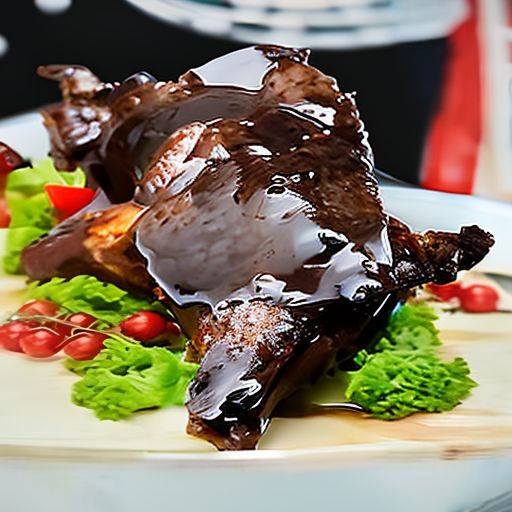}
    \end{subfigure}%
    \hfill
    \begin{subfigure}{0.14\linewidth}
        \centering
        \includegraphics[width=\linewidth]{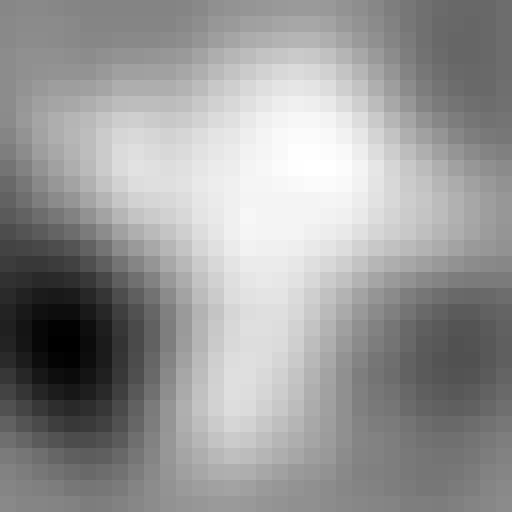}
    \end{subfigure}%

    \begin{subfigure}{0.14\linewidth}
        \centering
        \includegraphics[width=\linewidth]{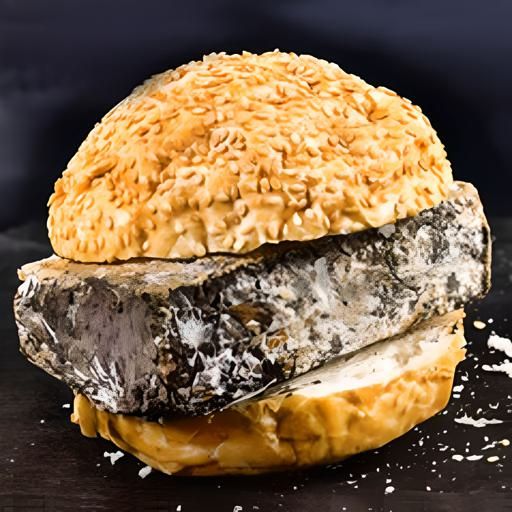}
    \end{subfigure}%
    \hfill
    \begin{subfigure}{0.14\linewidth}
        \centering
        \includegraphics[width=\linewidth]{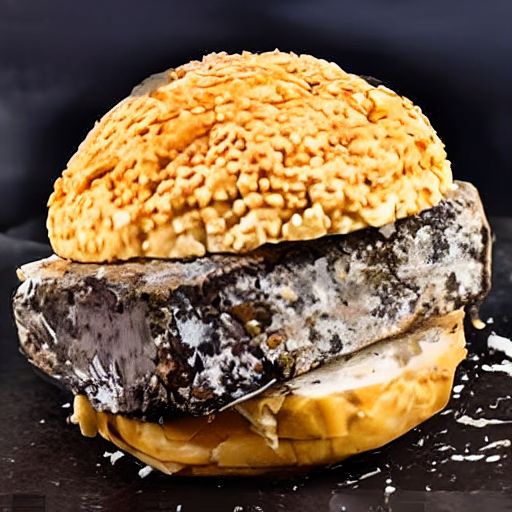}
    \end{subfigure}%
    \hfill
    \begin{subfigure}{0.14\linewidth}
        \centering
        \includegraphics[width=\linewidth]{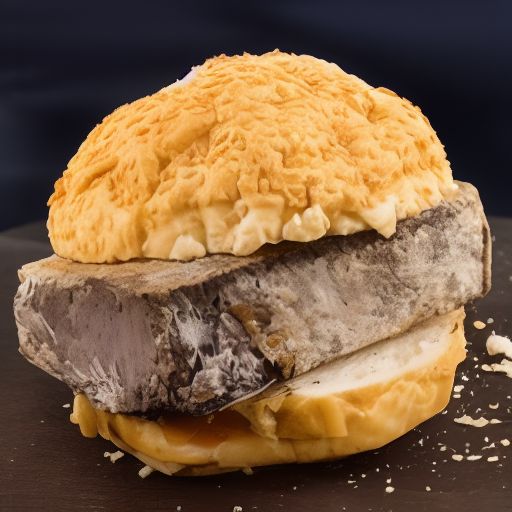}
    \end{subfigure}%
    \hfill
    \begin{subfigure}{0.14\linewidth}
        \centering
        \includegraphics[width=\linewidth]{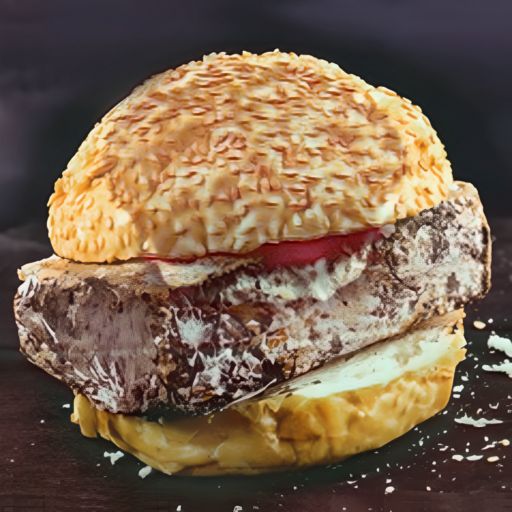}
    \end{subfigure}%
    \hfill
    \begin{subfigure}{0.14\linewidth}
        \centering
        \includegraphics[width=\linewidth]{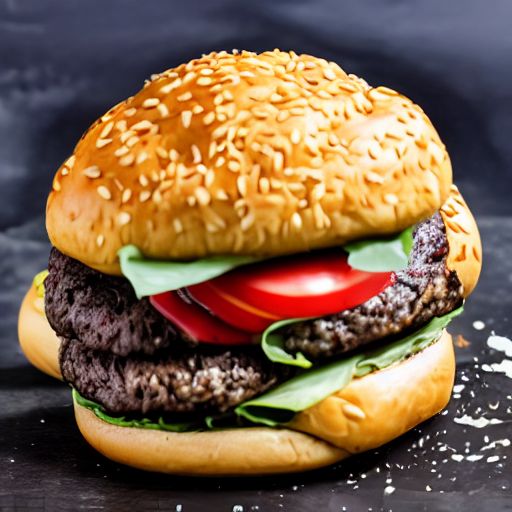}
    \end{subfigure}%
    \hfill
    \begin{subfigure}{0.14\linewidth}
        \centering
        \includegraphics[width=\linewidth]{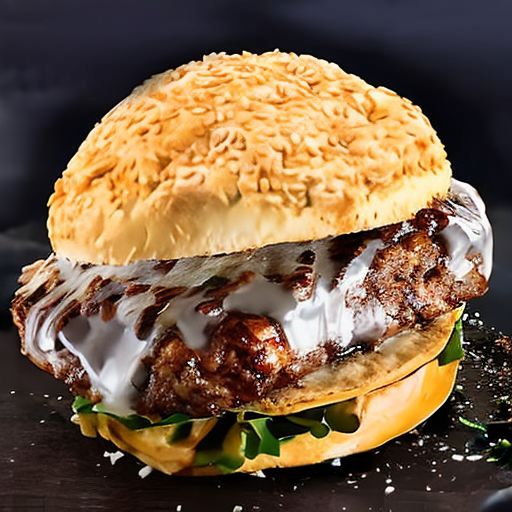}
    \end{subfigure}%
    \hfill
    \begin{subfigure}{0.14\linewidth}
        \centering
        \includegraphics[width=\linewidth]{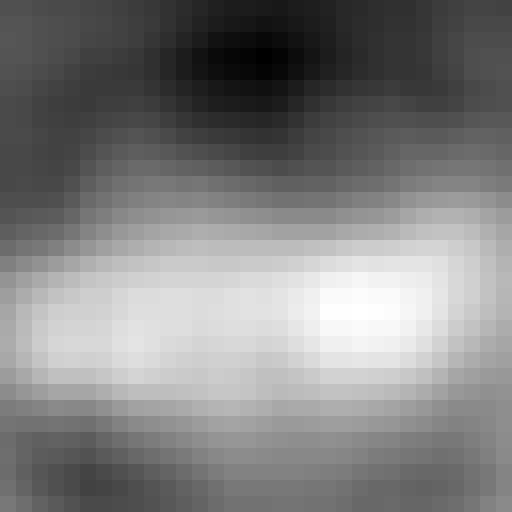}
    \end{subfigure}%

    \begin{subfigure}{0.14\linewidth}
        \centering
        \includegraphics[width=\linewidth]{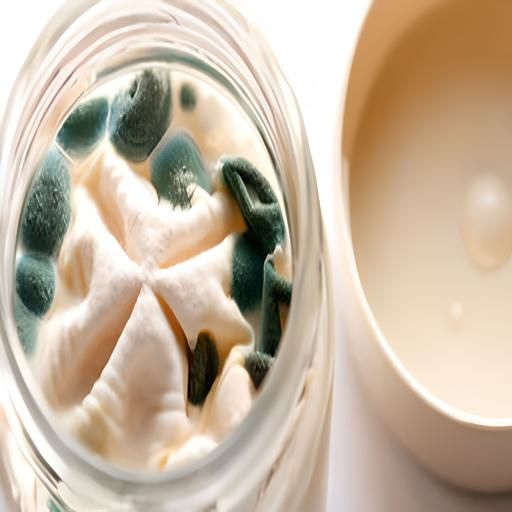}
    \end{subfigure}%
    \hfill
    \begin{subfigure}{0.14\linewidth}
        \centering
        \includegraphics[width=\linewidth]{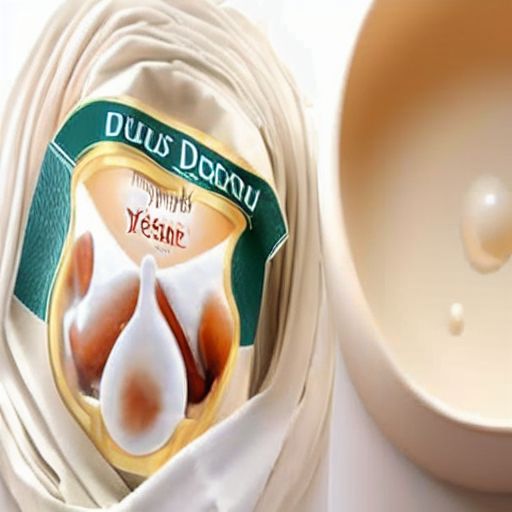}
    \end{subfigure}%
    \hfill
    \begin{subfigure}{0.14\linewidth}
        \centering
        \includegraphics[width=\linewidth]{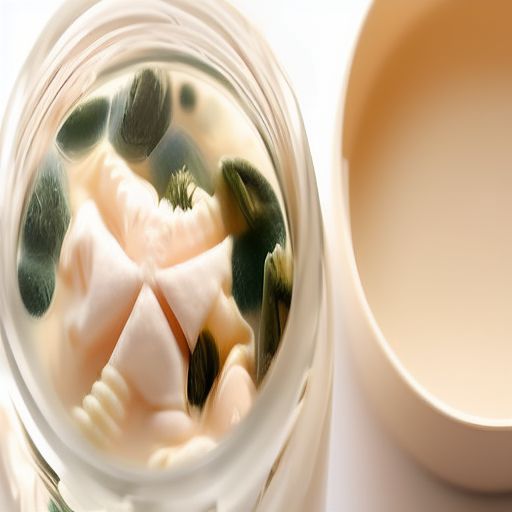}
    \end{subfigure}%
    \hfill
    \begin{subfigure}{0.14\linewidth}
        \centering
        \includegraphics[width=\linewidth]{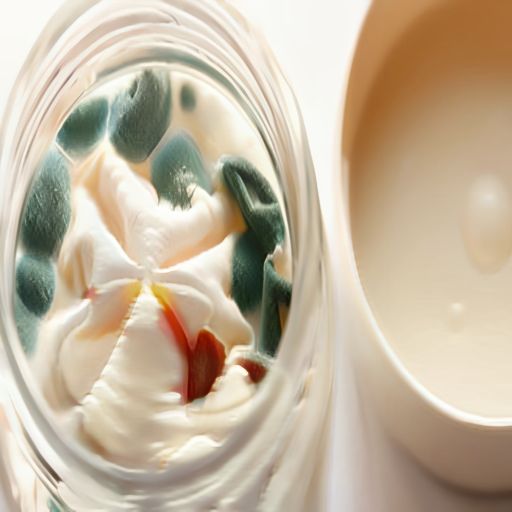}
    \end{subfigure}%
    \hfill
    \begin{subfigure}{0.14\linewidth}
        \centering
        \includegraphics[width=\linewidth]{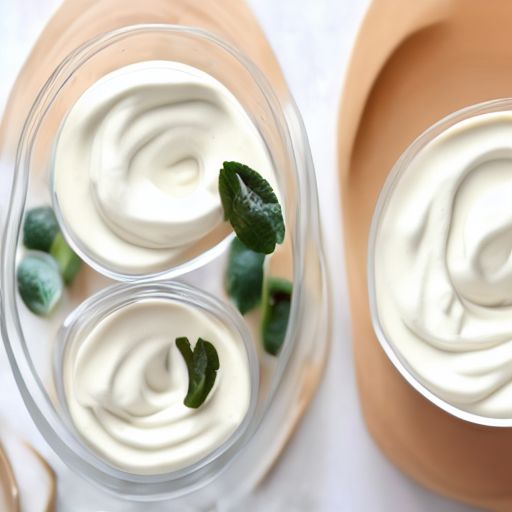}
    \end{subfigure}%
    \hfill
    \begin{subfigure}{0.14\linewidth}
        \centering
        \includegraphics[width=\linewidth]{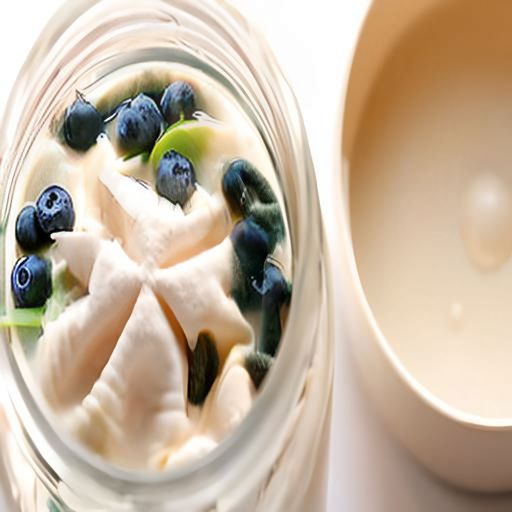}
    \end{subfigure}%
    \hfill
    \begin{subfigure}{0.14\linewidth}
        \centering
        \includegraphics[width=\linewidth]{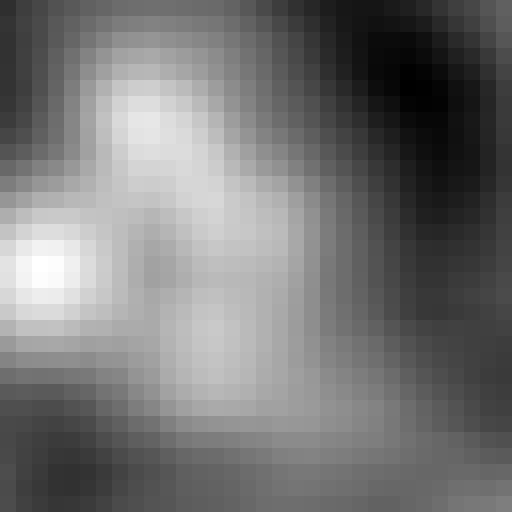}
    \end{subfigure}%

    \caption{\textbf{Food image content appeal enhancement comparison with baselines.} Compared with baselines, our enhancer respects the color, texture, and structures of the original images while effectively improving the appeal level of their content.}
    
\label{fig:appeal_enhancement_visual_compare_food}
\end{figure*}

\begin{figure*}[t]

    \centering
    \captionsetup[subfigure]{labelformat=empty}
    \begin{subfigure}{0.14\linewidth}
        \centering
        \caption{Input}
        \includegraphics[width=\linewidth]{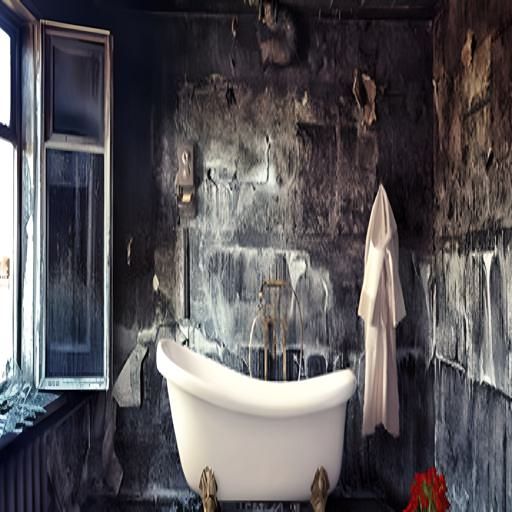}
    \end{subfigure}%
    \hfill
    \begin{subfigure}{0.14\linewidth}
        \centering
        \caption{N-TI}
        \includegraphics[width=\linewidth]{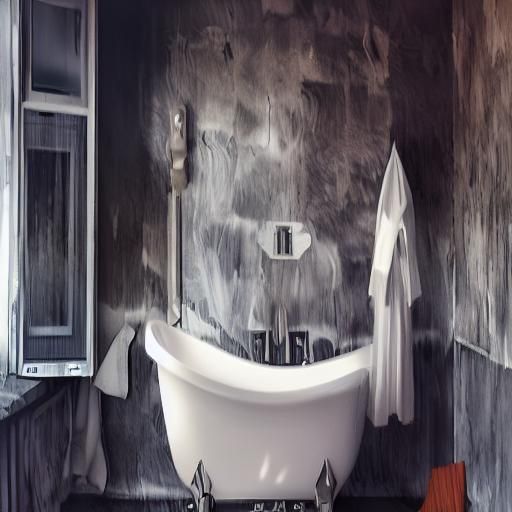}
    \end{subfigure}%
    \hfill
    \begin{subfigure}{0.14\linewidth}
        \centering
        \caption{P2P-0}
        \includegraphics[width=\linewidth]{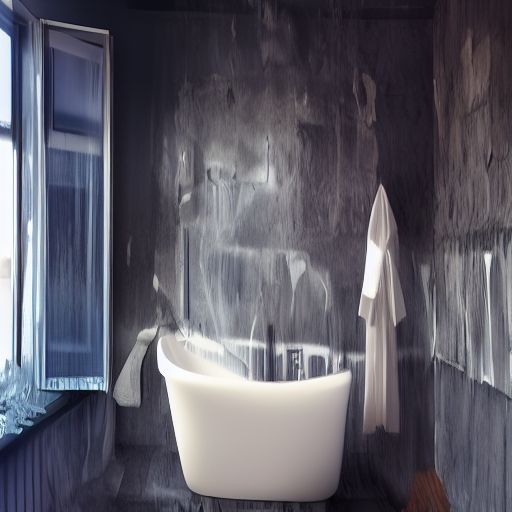}
    \end{subfigure}%
    \hfill
    \begin{subfigure}{0.14\linewidth}
        \centering
        \caption{T2L}
        \includegraphics[width=\linewidth]{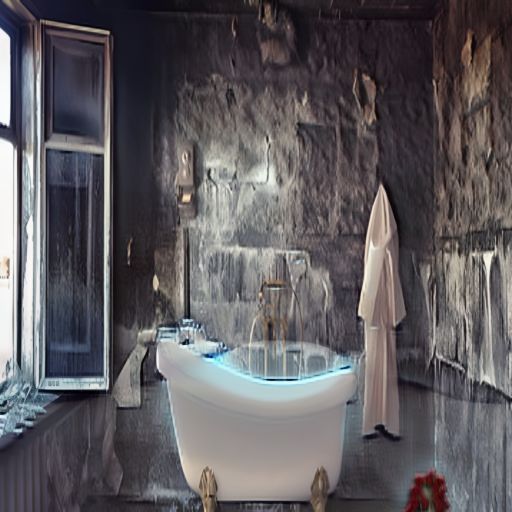}
    \end{subfigure}%
    \hfill
    \begin{subfigure}{0.14\linewidth}
        \centering
        \caption{IP2P}
        \includegraphics[width=\linewidth]{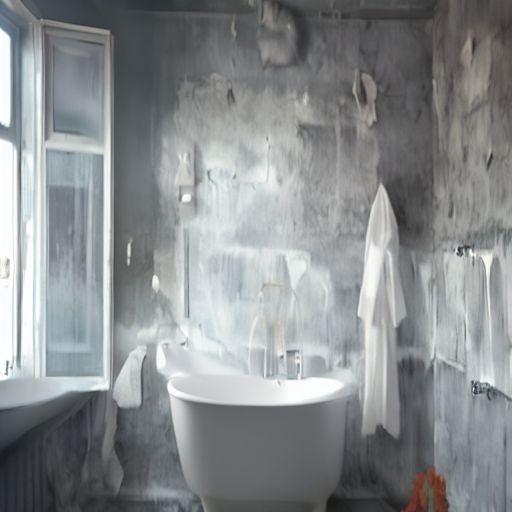}
    \end{subfigure}%
    \hfill
    \begin{subfigure}{0.14\linewidth}
        \centering
        \caption{Ours}
        \includegraphics[width=\linewidth]{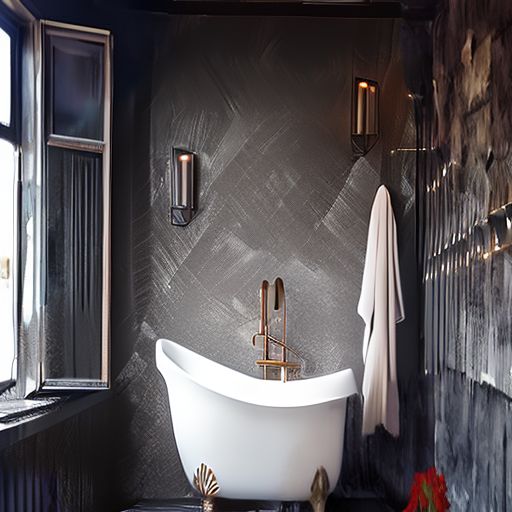}
    \end{subfigure}%
    \hfill
    \begin{subfigure}{0.14\linewidth}
        \centering
        \caption{Ours ($M_R^H$)}
        \includegraphics[width=\linewidth]{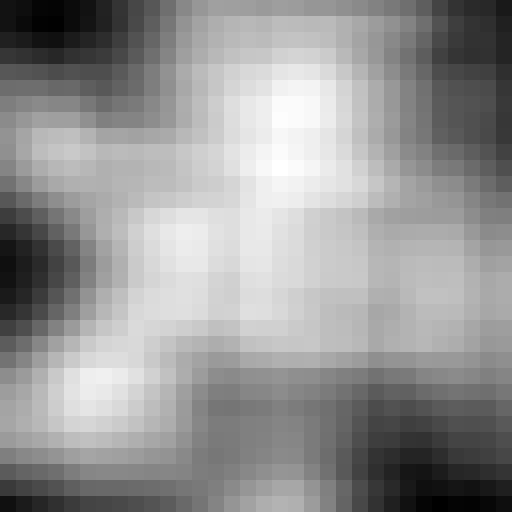}
    \end{subfigure}%

    \begin{subfigure}{0.14\linewidth}
        \centering
        \includegraphics[width=\linewidth]{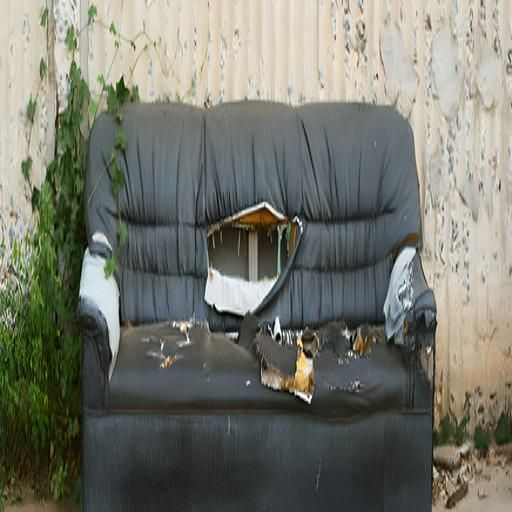}
    \end{subfigure}%
    \hfill
    \begin{subfigure}{0.14\linewidth}
        \centering
        \includegraphics[width=\linewidth]{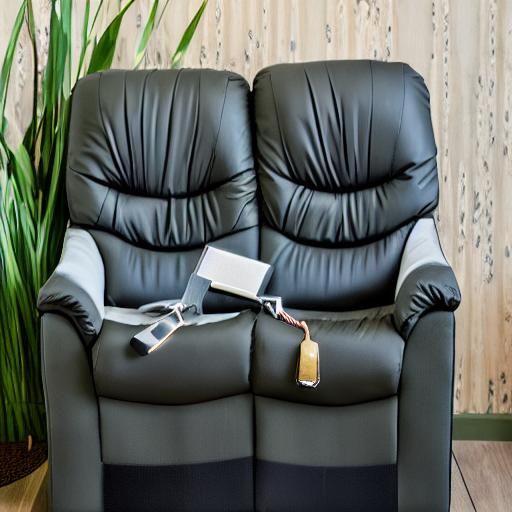}
    \end{subfigure}%
    \hfill
    \begin{subfigure}{0.14\linewidth}
        \centering
        \includegraphics[width=\linewidth]{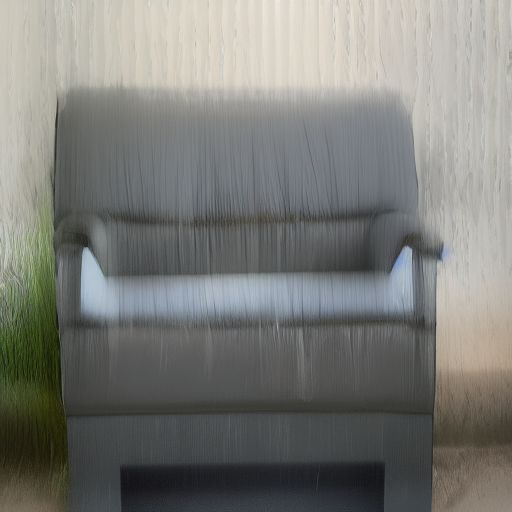}
    \end{subfigure}%
    \hfill
    \begin{subfigure}{0.14\linewidth}
        \centering
        \includegraphics[width=\linewidth]{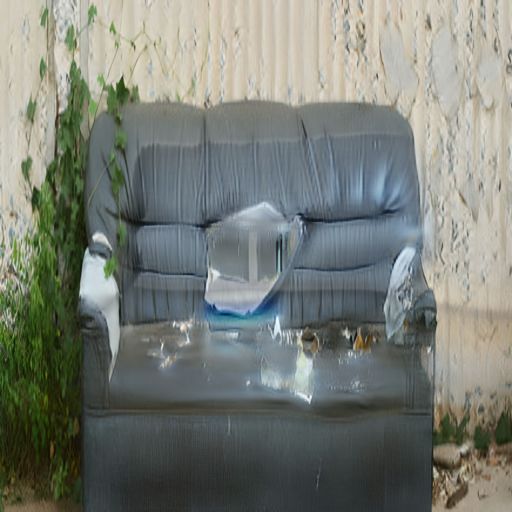}
    \end{subfigure}%
    \hfill
    \begin{subfigure}{0.14\linewidth}
        \centering
        \includegraphics[width=\linewidth]{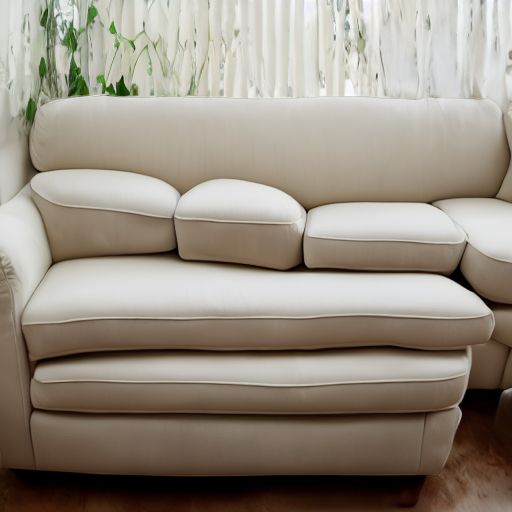}
    \end{subfigure}%
    \hfill
    \begin{subfigure}{0.14\linewidth}
        \centering
        \includegraphics[width=\linewidth]{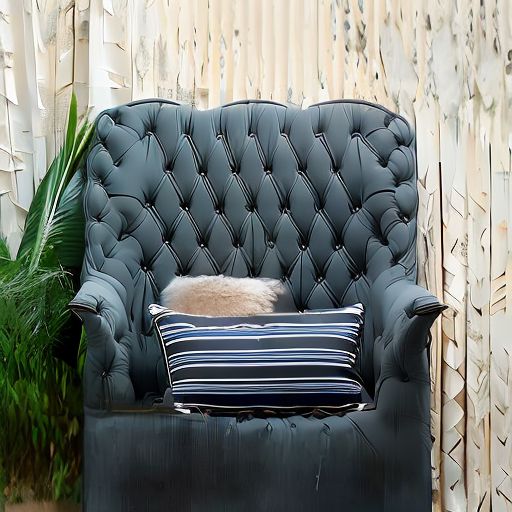}
    \end{subfigure}%
    \hfill
    \begin{subfigure}{0.14\linewidth}
        \centering
        \includegraphics[width=\linewidth]{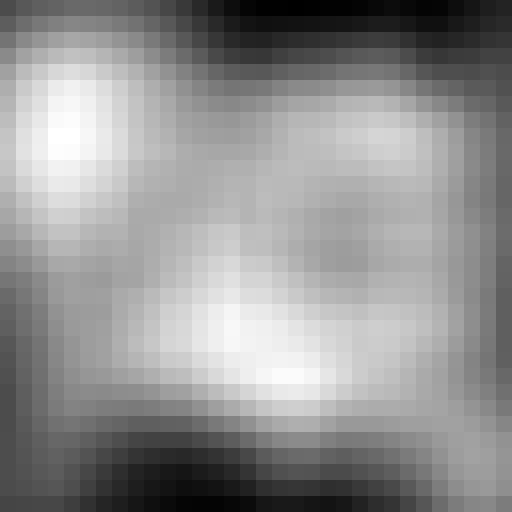}
    \end{subfigure}%

    \begin{subfigure}{0.14\linewidth}
        \centering
        \includegraphics[width=\linewidth]{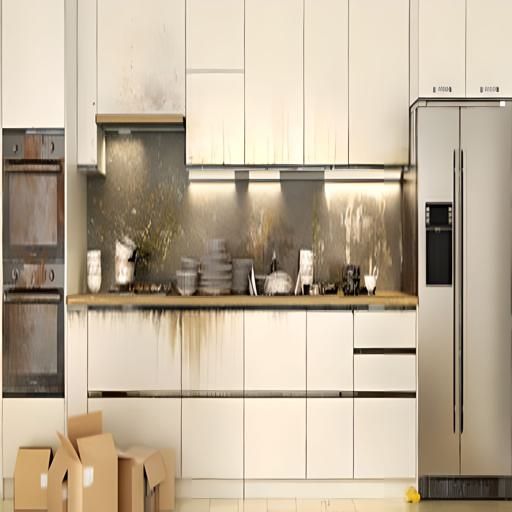}
    \end{subfigure}%
    \hfill
    \begin{subfigure}{0.14\linewidth}
        \centering
        \includegraphics[width=\linewidth]{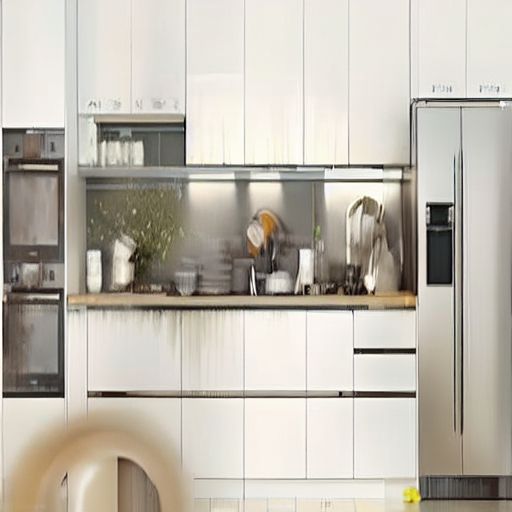}
    \end{subfigure}%
    \hfill
    \begin{subfigure}{0.14\linewidth}
        \centering
        \includegraphics[width=\linewidth]{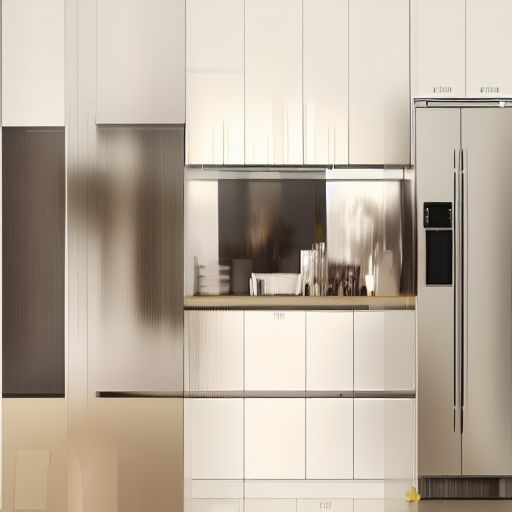}
    \end{subfigure}%
    \hfill
    \begin{subfigure}{0.14\linewidth}
        \centering
        \includegraphics[width=\linewidth]{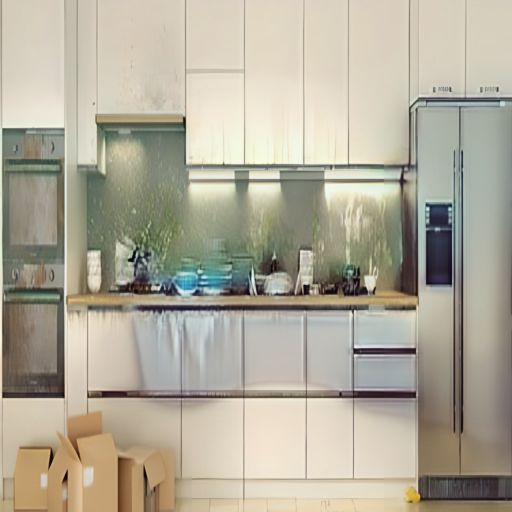}
    \end{subfigure}%
    \hfill
    \begin{subfigure}{0.14\linewidth}
        \centering
        \includegraphics[width=\linewidth]{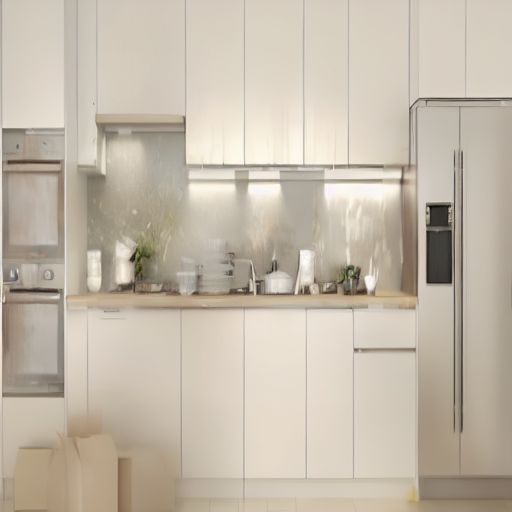}
    \end{subfigure}%
    \hfill
    \begin{subfigure}{0.14\linewidth}
        \centering
        \includegraphics[width=\linewidth]{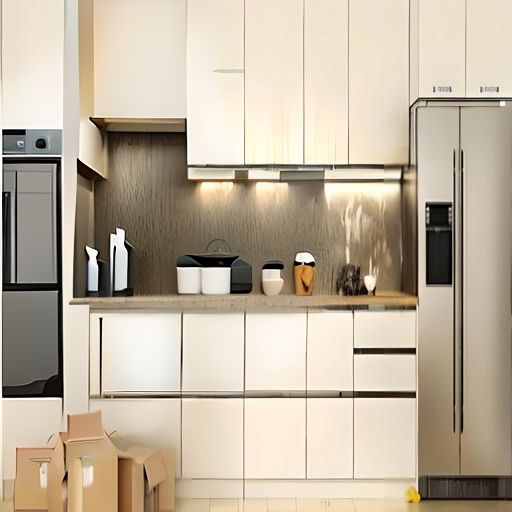}
    \end{subfigure}%
    \hfill
    \begin{subfigure}{0.14\linewidth}
        \centering
        \includegraphics[width=\linewidth]{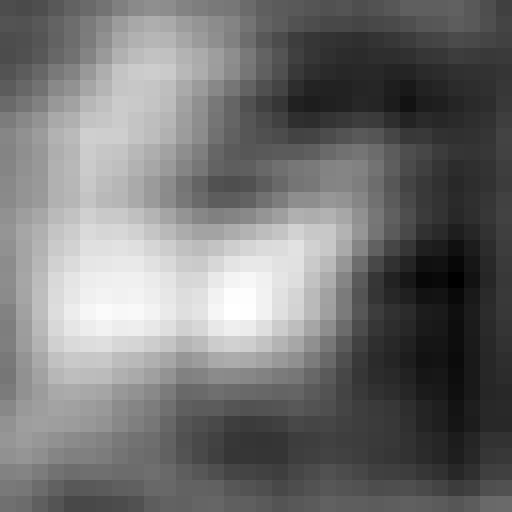}
    \end{subfigure}%

    \begin{subfigure}{0.14\linewidth}
        \centering
        \includegraphics[width=\linewidth]{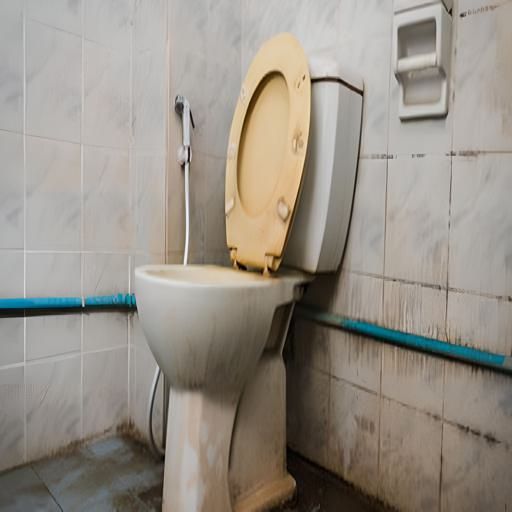}
    \end{subfigure}%
    \hfill
    \begin{subfigure}{0.14\linewidth}
        \centering
        \includegraphics[width=\linewidth]{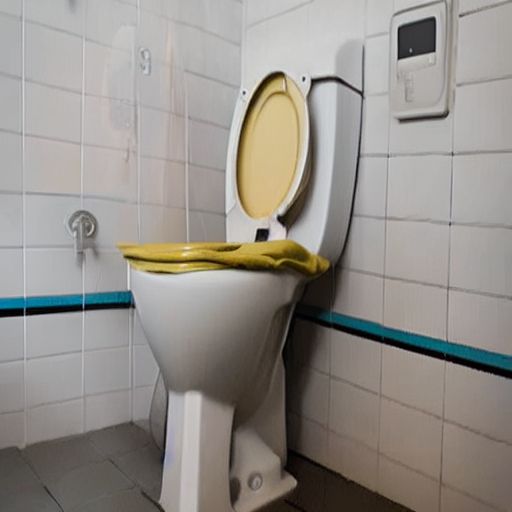}
    \end{subfigure}%
    \hfill
    \begin{subfigure}{0.14\linewidth}
        \centering
        \includegraphics[width=\linewidth]{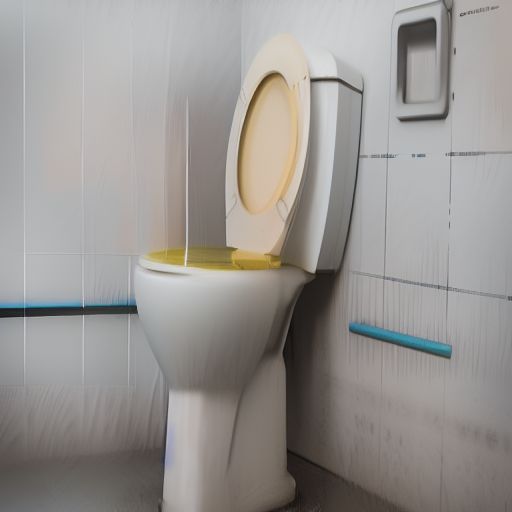}
    \end{subfigure}%
    \hfill
    \begin{subfigure}{0.14\linewidth}
        \centering
        \includegraphics[width=\linewidth]{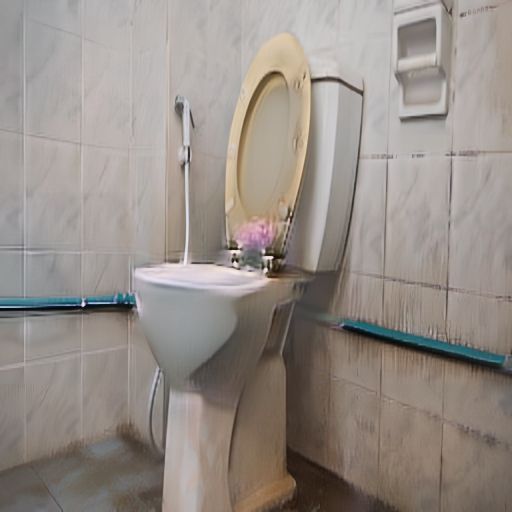}
    \end{subfigure}%
    \hfill
    \begin{subfigure}{0.14\linewidth}
        \centering
        \includegraphics[width=\linewidth]{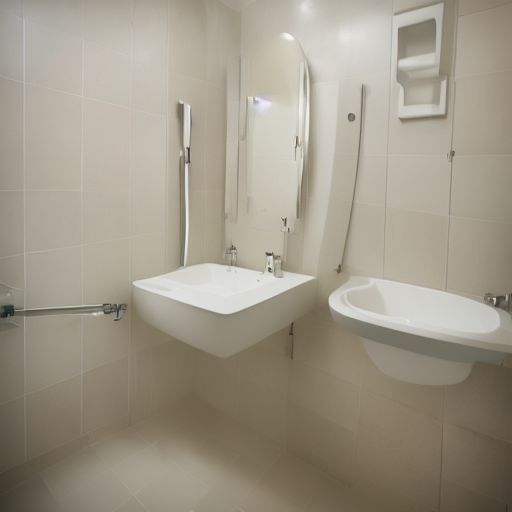}
    \end{subfigure}%
    \hfill
    \begin{subfigure}{0.14\linewidth}
        \centering
        \includegraphics[width=\linewidth]{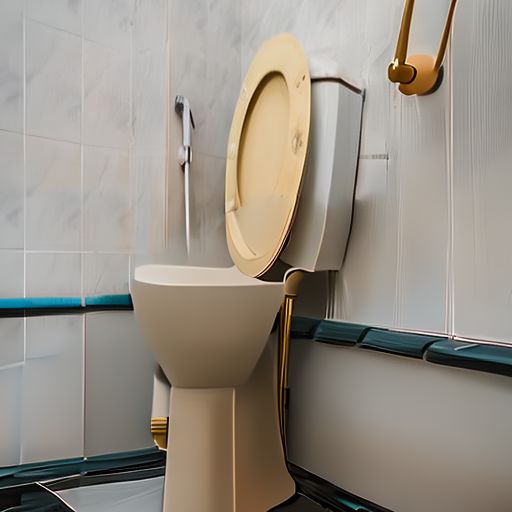}
    \end{subfigure}%
    \hfill
    \begin{subfigure}{0.14\linewidth}
        \centering
        \includegraphics[width=\linewidth]{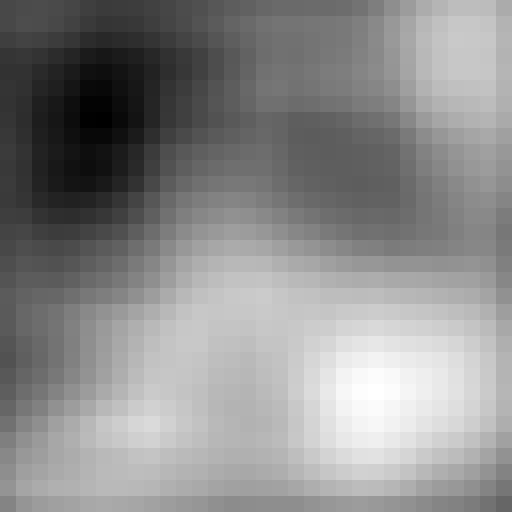}
    \end{subfigure}%

    \caption{\textbf{Room image content appeal enhancement comparison with baselines.} Our enhancer better respects the color, texture, and structures of the original images while effectively improving the appeal level of their content than baselines.}
\label{fig:appeal_enhancement_visual_compare_room}

\end{figure*}

\subsection{Human preference comparison}

Out of 2072 responses, 672 of them compared the appeal of real images in the \textsc{Food} and \textsc{Room} datasets, respectively. We first assess the accuracy of our labeling process by plotting in \cref{fig:appeal_compare_response} the distribution of appeal label differences, $A(\text{image A}) - A(\text{image B})$, versus user preference, and observe a clear decrease in mean values from response 1 to 5. This indicates that image B's appeal increases relative to image A aligns with user preferences and demonstrates that our content appeal labels are indeed accurate.

To compare content appeal before and after appeal enhancement (Fig.~\ref{fig:appeal_enhancement_response}), we received 392 responses for the \textsc{Food} dataset and \textsc{336} for the Room dataset out of the 2072 responses. For the former, 76.53\% of responses favored enhanced images, with 41.58\% strongly preferring them. In the \textsc{Room} dataset, 82.74\% preferred enhanced images, with 53.87\% showing strong preference. This demonstrates a clear preference for enhanced images from our methods across both datasets.

\subsection{Content appeal enhancer baseline comparison}


We have chosen existing diffusion-based image editing methods InstructPix2Pix (IP2P)~\cite{brooks2022instructpix2pix}, Null-text Inversion (NTI)~\cite{mokady2022null}, pix2pix-zero (P2P0)~\cite{parmar2023zero}, and Text2LIVE (T2L)~\cite{bar2022text2live}, which we think can be applied in a similar setting to ours. 

In \cref{fig:appeal_enhancement_visual_compare_food,fig:appeal_enhancement_visual_compare_room}, we provide a visual comparison between our method and these baselines, as well as our content appeal heatmaps $M_D^H$. As we can see, N-TI tends to enlarge objects in images (\cref{fig:appeal_enhancement_visual_compare_food}, Rows 1-3) without changing the content appeal level much. It may also produces drastic undesired change to the change (\cref{fig:appeal_enhancement_visual_compare_food}, Row 4; \cref{fig:appeal_enhancement_visual_compare_room}, Row 2). P2P-0 and T2L often blur objects and create shadowing artifacts with little effect on the content appeal; IP2P is prone to changing the images too much (\cref{fig:appeal_enhancement_visual_compare_food}, Rows 3-4; \cref{fig:appeal_enhancement_visual_compare_room}, Rows 1, 2, 4). In contrast, our method is able to constrain the location and the magnitude of image content appeal enhancement using the heatmap and produce results with improved content appeal while respecting the color and structure of the input images. Please refer to the supplementary for baseline details, more results, and the ablation study.

\section{Conclusion}
In this work, we explored a new area of image appeal assessment (\IPA{}) that evaluates the interest an image creates in observers. We highlight the challenge of manual labeling in dataset creation, and propose a fully automated pipeline to generate extensive datasets across domains. Our research illustrates how these datasets can be used to train an appeal estimator and facilitate appeal enhancement applications. Validation of our methods is conducted through a user study, confirming their effectiveness.

\bibliographystyle{splncs04}
\bibliography{main}

\clearpage
\appendix
\section*{Appendix}

Here we elaborate on our datasets in \cref{sec:dataset}, including the creation process and sample images (\cref{sec:dataset_creation_details}), as well as method generalizability across image domains (\cref{sec:dataset_creation_generalizability}). 

In \cref{sec:estimator_details}, we compare the content appeal labels in our dataset with aesthetic scores from IAA baselines (\cref{sec:iaa_baseline_comparison}), demonstrate the generalizability of the models on amateur-taken images (\cref{sec:amateur_generalizability}), and discuss the effect of technical distortions on content appeal (\cref{sec:technical_distortion_effect}).

\cref{sec:content_appeal_enhancer_details} outlines the configuration of our content appeal enhancer, followed by more enhancement results  and ablation studies in \cref{sec:image_content_appeal_enhancement}, while \cref{sec:enhancement_baselines_details} provides further setup details of enhancement baselines we compared in the paper.

Finally, \cref{sec:supp_user_study} furnishes more information regarding our user study, including the questionnaire and analysis of data collected from the participants.

\section{Dataset Details\label{sec:dataset}}

\subsection{Dataset creation details and samples\label{sec:dataset_creation_details}}

To show that \ourwork{} can be generalized across different image domains, we create two datasets, one with food images and the other with room interior images. Here we present them in detail.

\noindent \textsc{\textbf{Food}}: Search queries were generated from the following sets of words:
\begin{itemize}
    \item $\set{N}_F = \{$``burger,'' ``cake,'' ``chicken,'' ``cookie,'' ``food,'' ``rice,'' ``pizza,'' ``pasta,'' ``salad,'' ``steak,'' ``yogurt''$\}$
    \item $\set{A}_F^+ = \{$``delicious''$\}$
    \item $\set{A}_F^- = \{$``burnt,'' ``moldy,'' ``rotten''$\}$
\end{itemize}
We generated search queries and retrieved 189,477 image thumbnails from image hosting sites \href{https://stock.adobe.com}{stock.adobe.com} and \href{https://shutterstock.com}{shutterstock.com}. We used our filtering method with $\gamma = 0.4$, which gave us 80,067 images, all of which were upscaled and zero-padded to $512 \times 512$ resolution. 

We selected 50 ``delicious food'' images as $\set{T}_F^+$,  50 ``burnt food'' images as $\set{T}_{F_1}^{-}$, as well as a total of 50 ``moldy food'' and ``rotten food'' images as $\set{T}_{F_2}^{-}$ for textual inversion. We generated two $\set{T}_F^{-}$'s because burnt food and moldy/rotten food have distinctly different features (blackened food vs. hairy mold) that rarely appear in the same image in real life. Mixing them will generate images with both characteristics together, which is not very realistic. All selected images appear at the top of search results by search engines using the corresponding queries to ensure maximum relevance between image content and search queries. We train $z_F^+$, $z_{F_1}^{-}$, and $z_{F_2}^{-}$ with $\set{T}_F^+$, $\set{T}_{F_1}^{-}$, and $\set{T}_{F_2}^{-}$ respectively using Stable Diffusion with batch size 1 and learning rate $lr=5e^{-3}$. 

\begin{figure}[t]
    \centering
    \begin{tabular}{cc@{\hspace{0.1in}}c}
     & Food    & Room \\  
     HL & \includegraphics[width=0.1\linewidth]{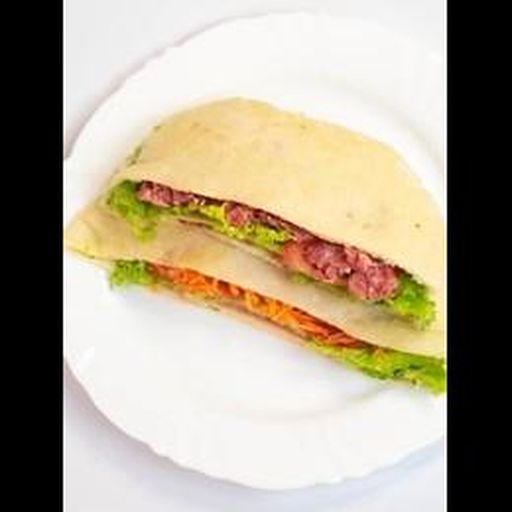}
    \includegraphics[width=0.1\linewidth]{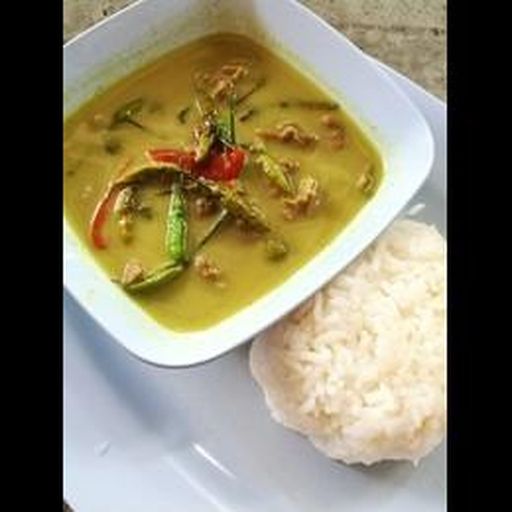}
    \includegraphics[width=0.1\linewidth]{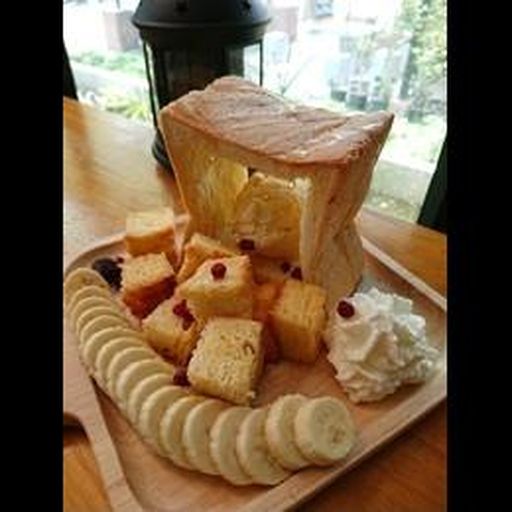}
    \includegraphics[width=0.1\linewidth]{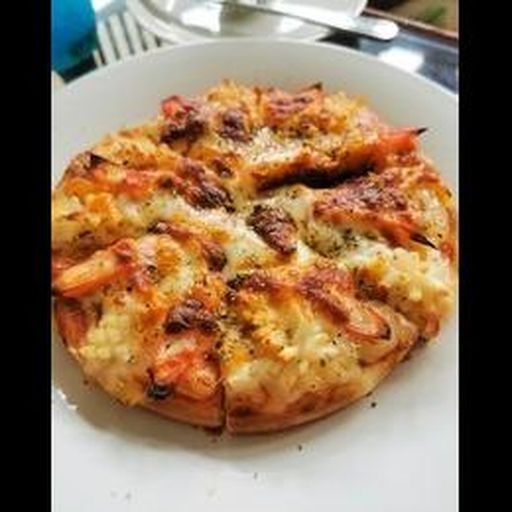} & \includegraphics[width=0.1\linewidth]{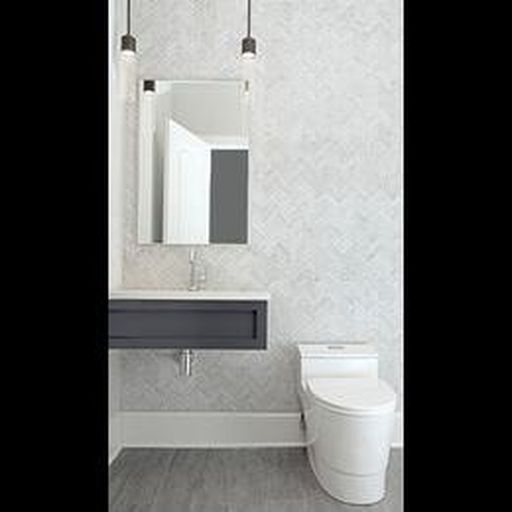}
    \includegraphics[width=0.1\linewidth]{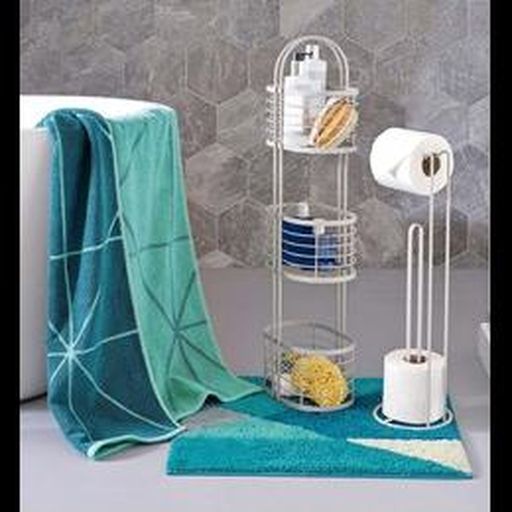}
    \includegraphics[width=0.1\linewidth]{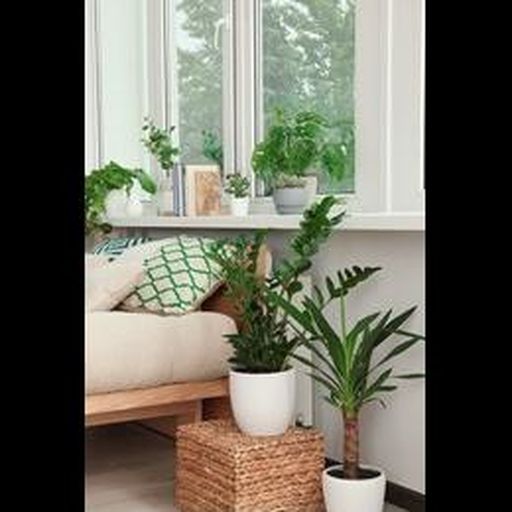}\includegraphics[width=0.1\linewidth]{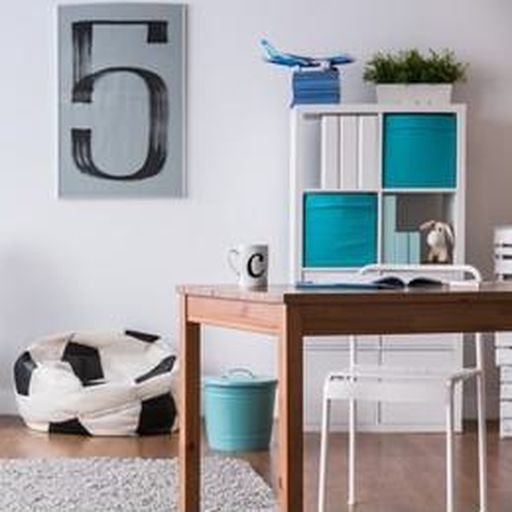} \\ 
    LH & \includegraphics[width=0.1\linewidth]{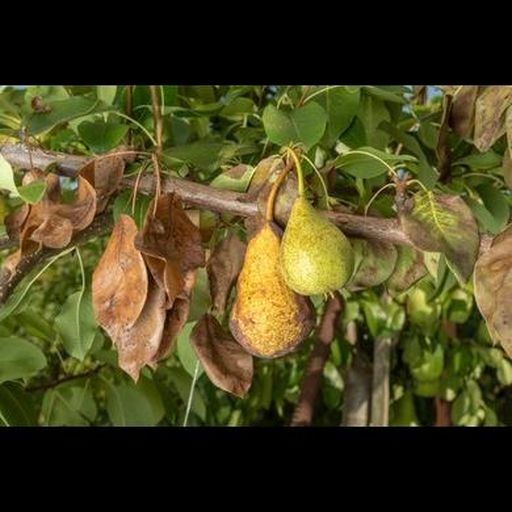}
    \includegraphics[width=0.1\linewidth]{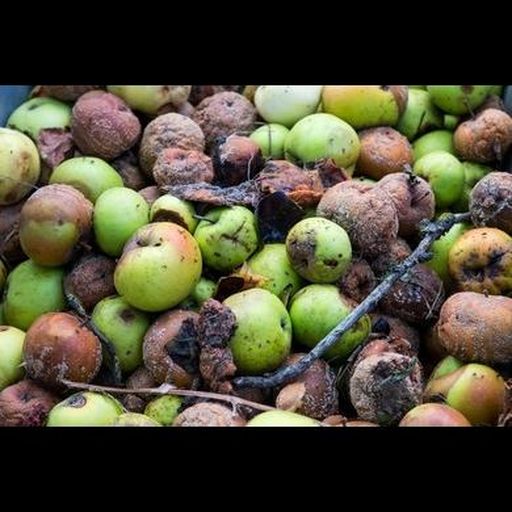}
    \includegraphics[width=0.1\linewidth]{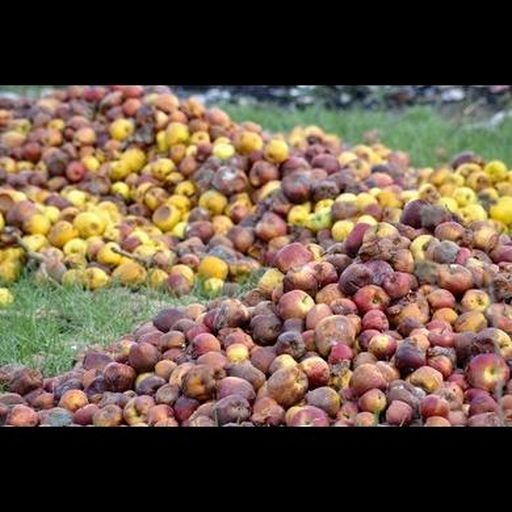}
    \includegraphics[width=0.1\linewidth]{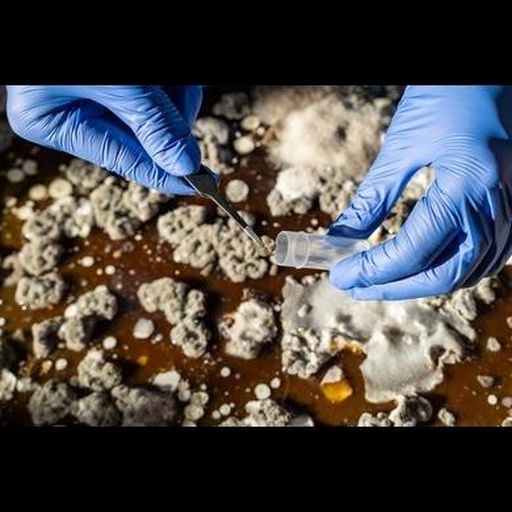} & \includegraphics[width=0.1\linewidth]{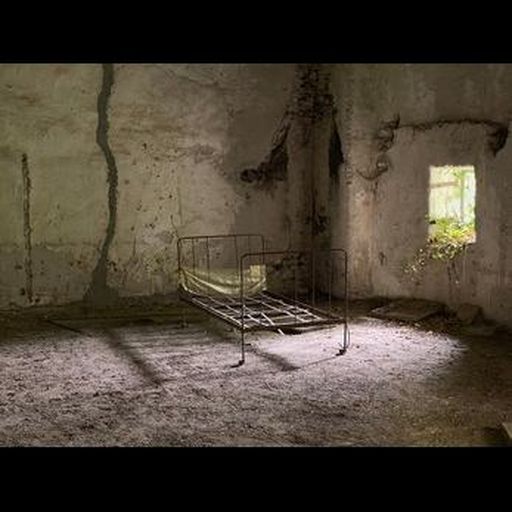}
   \includegraphics[width=0.1\linewidth]{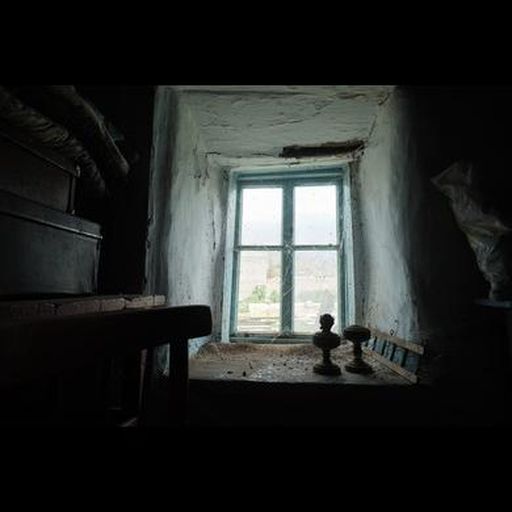}
   \includegraphics[width=0.1\linewidth]{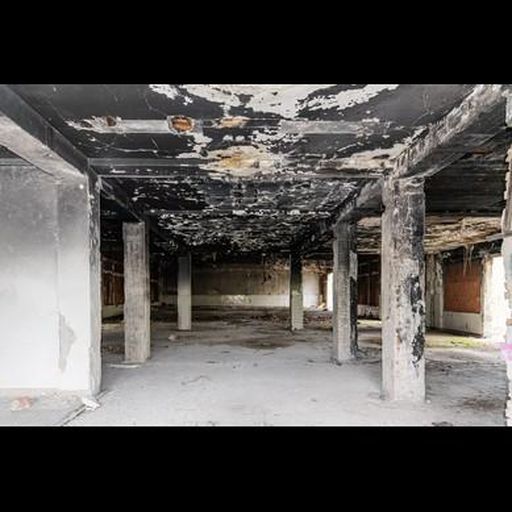}
   \includegraphics[width=0.1\linewidth]{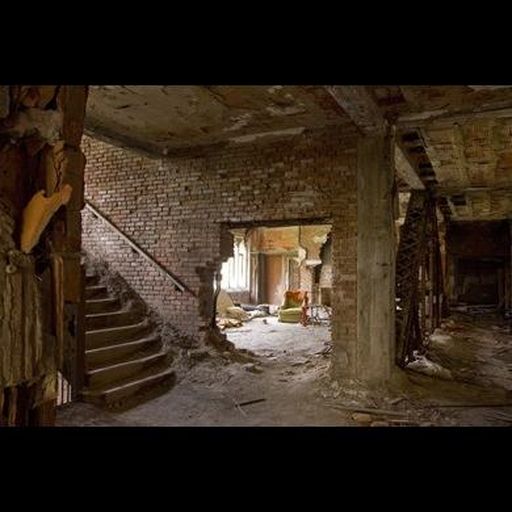} \\ 
    HH & \includegraphics[width=0.1\linewidth]{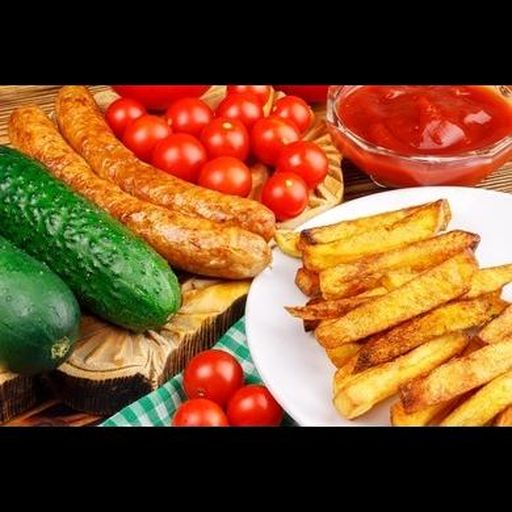}
    \includegraphics[width=0.1\linewidth]{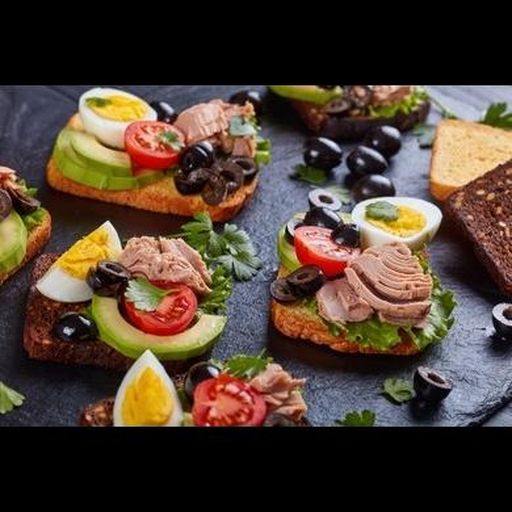}
    \includegraphics[width=0.1\linewidth]{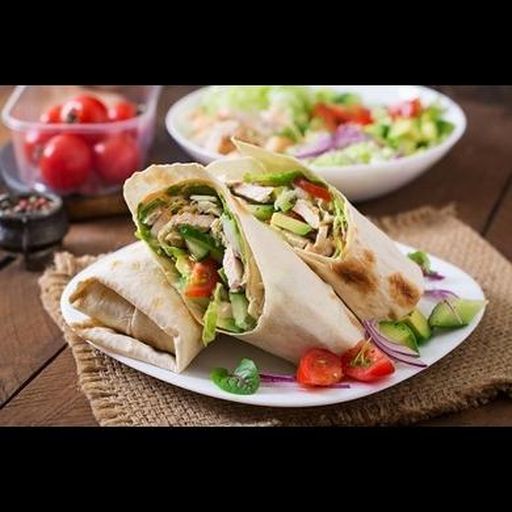}
    \includegraphics[width=0.1\linewidth]{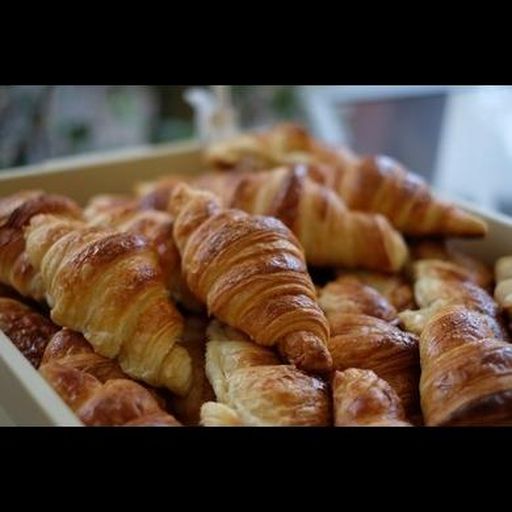} & \includegraphics[width=0.1\linewidth]{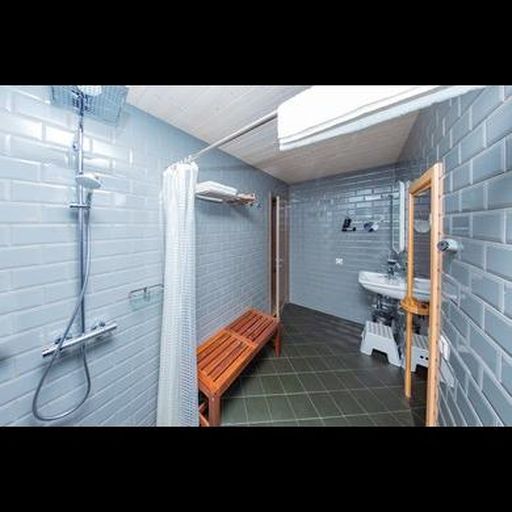}
    \includegraphics[width=0.1\linewidth]{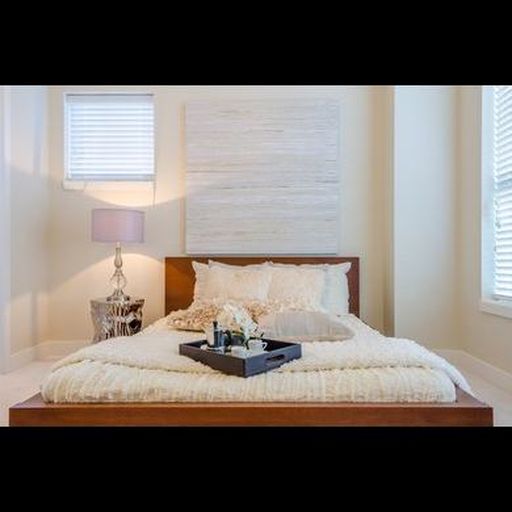}
    \includegraphics[width=0.1\linewidth]{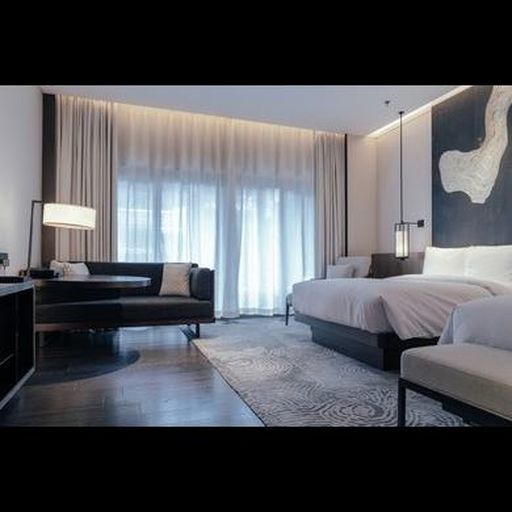}
    \includegraphics[width=0.1\linewidth]{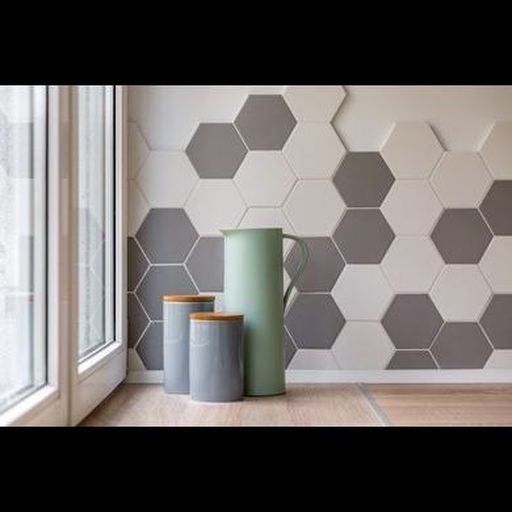} \\
    LL & \includegraphics[width=0.1\linewidth]{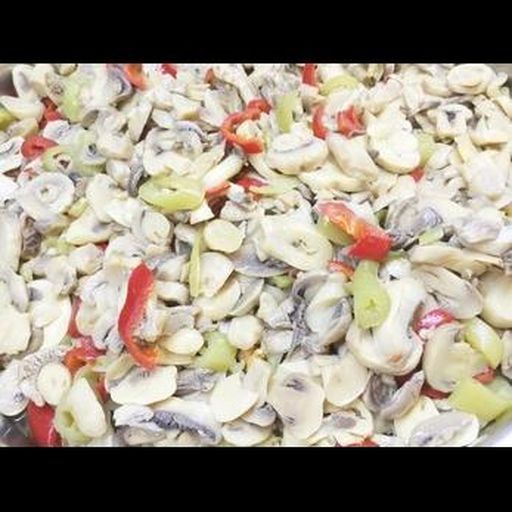}
    \includegraphics[width=0.1\linewidth]{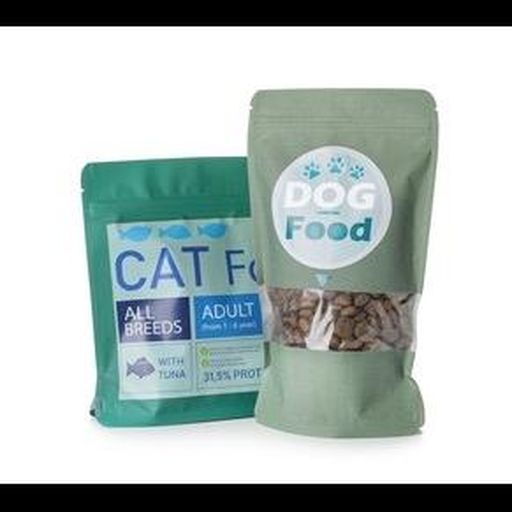}
    \includegraphics[width=0.1\linewidth]{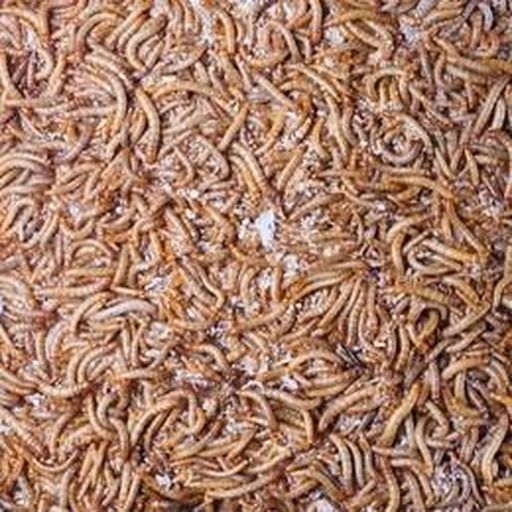}
    \includegraphics[width=0.1\linewidth]{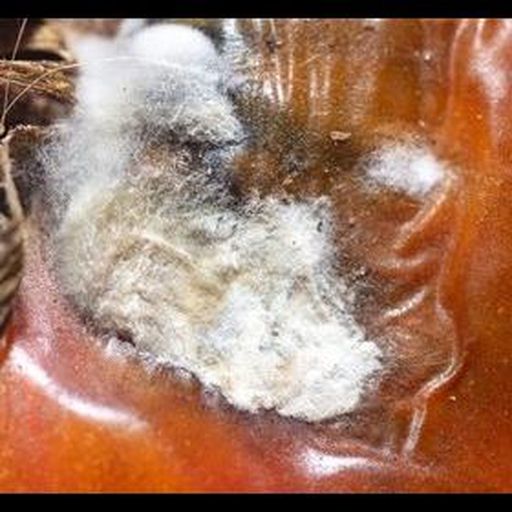} & \includegraphics[width=0.1\linewidth]{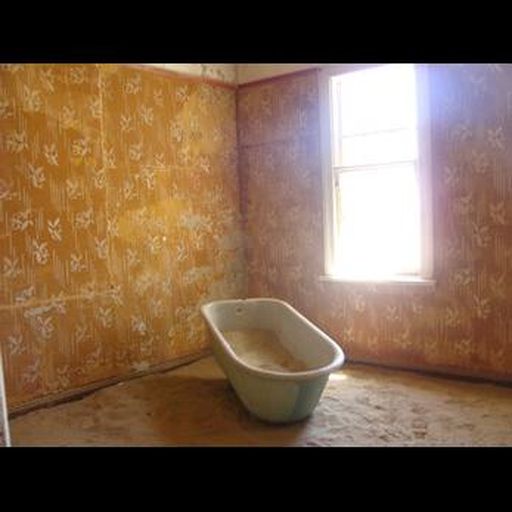}
    \includegraphics[width=0.1\linewidth]{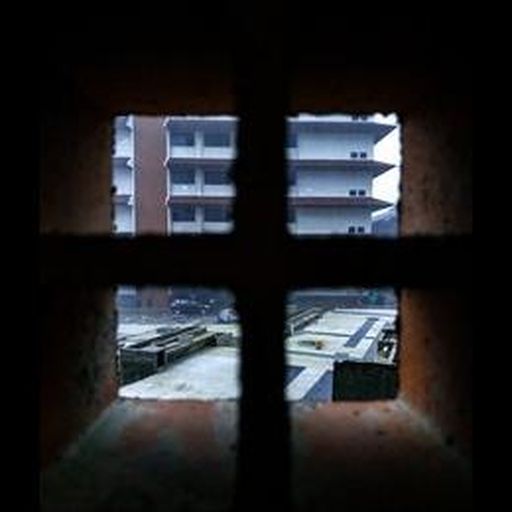}
    \includegraphics[width=0.1\linewidth]{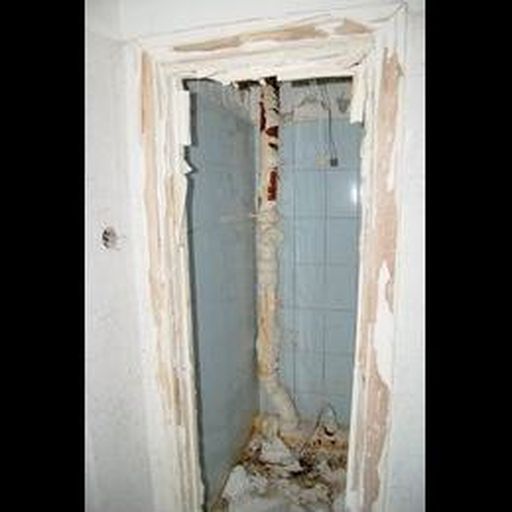}
    \includegraphics[width=0.1\linewidth]{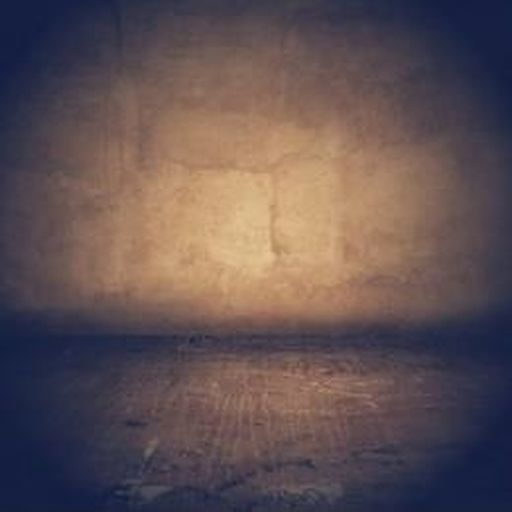}
    \end{tabular}
    \caption{\textbf{Dataset samples.} We show 4 sample images from each of the food and room interior dataset, where the label next to each row indicates the content appeal score and image aesthetic score level of images in the corresponding row. Images with scores above the $75^{th}$ percentile in each dataset or IAA baseline predictions are considered to have high (H) scores. Images with scores below the $25^{th}$ percentile in each dataset or IAA baseline predictions are considered to have low (L) scores.}
    \label{fig:dataset_samples}
\end{figure}

Following that, we select a different set of 1,000 images with balanced content appeal levels and food types as the starting point of our synthetic dataset. Specifically, we choose 50 images retrieved from each $q = a + n$ where $+$ means appending to $a \in \set{A}_F^+, n \in \set{N}_F - \{$``food''$\}$, which gives us 500 images with appealing content and balanced food types as $I_F^+$. Similarly, we choose 50 images retrieved using $a \in \set{A}_F^-$, which gives us a total of 500 images with unappealing content as $\set{I}_F^-$. All selected images appear at the top of search results by search engines using the corresponding queries to ensure maximum relevance between image content and search queries. We use $n \in \set{N}_F - \{$``food''$\}$ to help constrain object types and keep them balanced.  For each $i \in \set{I}_F^+ \cup \set{I}_F^-$, we first augment it to generate three versions of $I' = SD(I, \text{`` ''}, 1 - M_F(I), \text{seed}())$, where  $1- M_F(I)$ is the inverse of domain-relevancy map for the food domain. For each $I'$, we generate 6 final images $s = SD(I', BLIP(I) + f(\alpha), M_F(I), \text{seed}())$, where 
\begin{equation}
    \begin{aligned}
    \alpha &= max(min(k/2 + \delta, 1), 0) \\
    k &\in {0, 1, 2} \\
    \delta &\in uniform(-0.2, 0.2).
    \end{aligned}
\end{equation} 
Note that $k$ is used to ensure that 6 images generated from each $i$ span the entire content appeal spectrum. We use $\delta$ to add randomization and more variety in  $\hat{A}(\cdot, \cdot)$ to avoid over-fitting when training our relative content appeal comparator.  In the end, we generated 18,000 images as our synthetic dataset $\set{S}_F$ and 78,917 remaining images for the final dataset $\set{I}_F$.

\begin{figure}[t]
\tcbset{
    on line,
    colframe=black, 
    boxsep=0pt,     
    left=0pt,       
    right=0pt,      
    top=0pt,        
    bottom=0pt,     
    boxrule=0.2pt,     
    sharp corners   
}
    \centering
    \scriptsize
    \begin{tabular}{c@{\hspace{1pt}}c@{\hspace{1pt}}c@{\hspace{1pt}}c@{\hspace{2pt}}@{\hspace{2pt}}c@{\hspace{1pt}}c@{\hspace{1pt}}c}
     Mask & Input & 0.5 $z_V^+$ & 1.0 $z_V^+$ & Input & 0.5 $z_L^+$ & 1.0 $z_L^+$\\
    \tcbox{\includegraphics[width=0.115\linewidth]{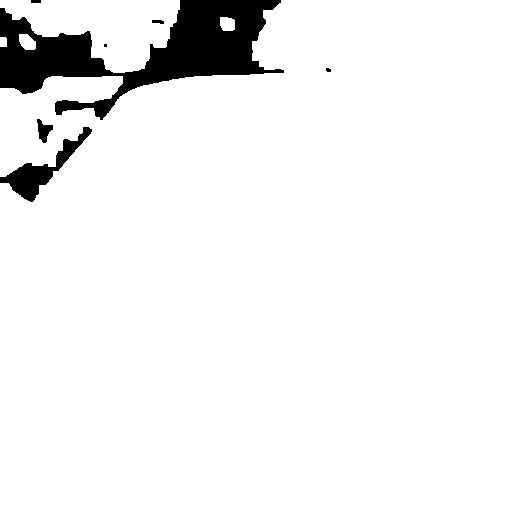}} &
     \includegraphics[width=0.115\linewidth]{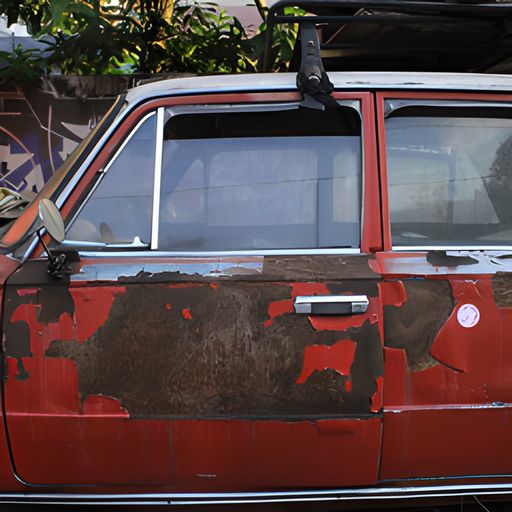} & \includegraphics[width=0.115\linewidth]{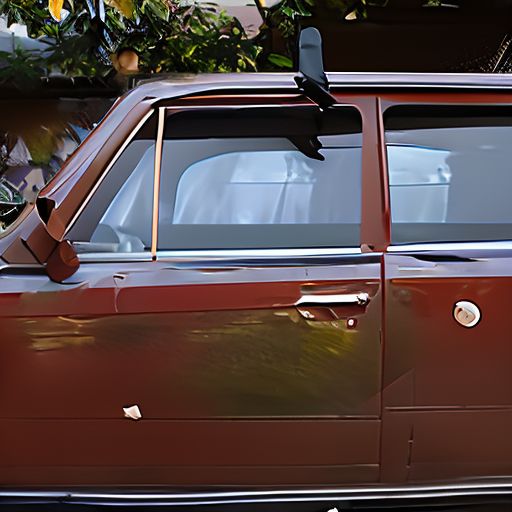} & \includegraphics[width=0.115\linewidth]{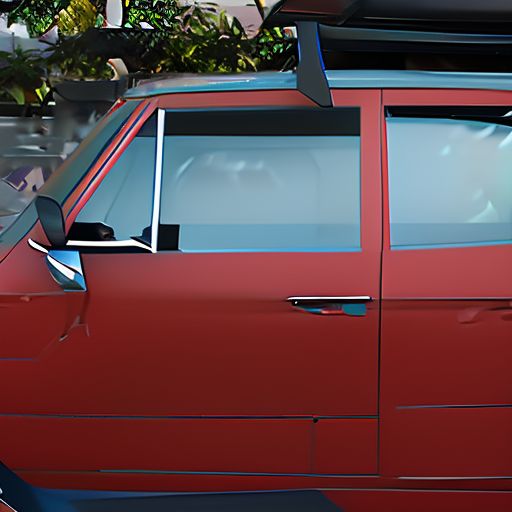} & \includegraphics[width=0.115\linewidth]{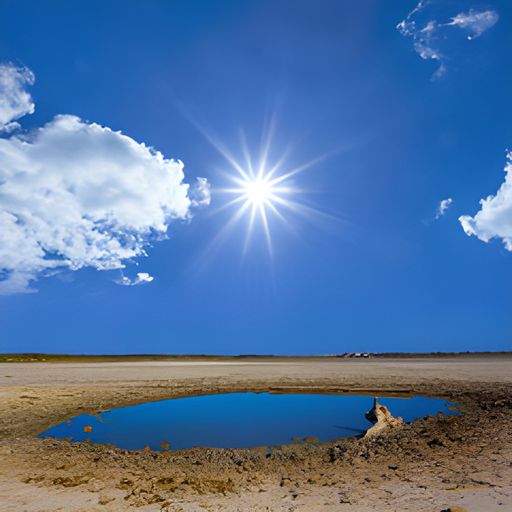} & \includegraphics[width=0.115\linewidth]{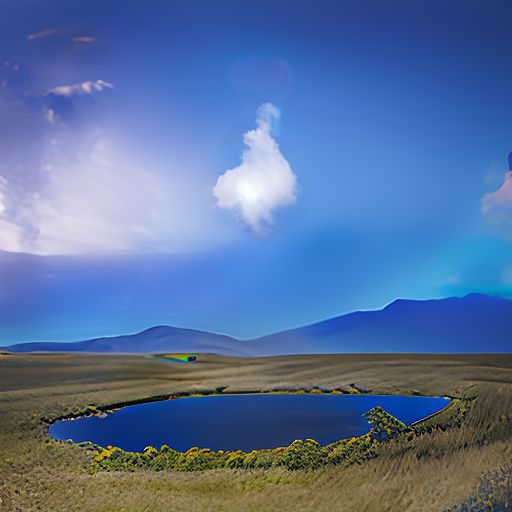} & \includegraphics[width=0.115\linewidth]{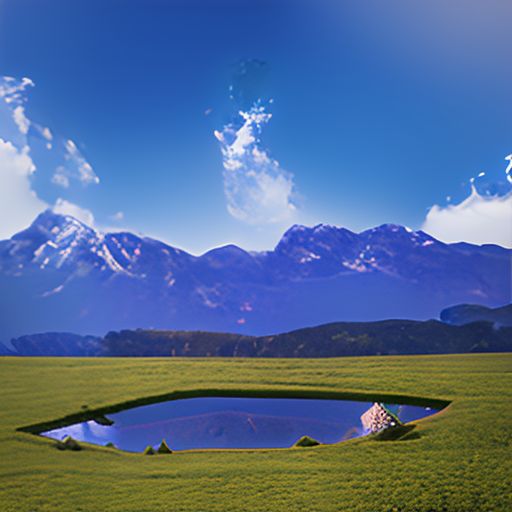} \\ 

     \includegraphics[width=0.115\linewidth]{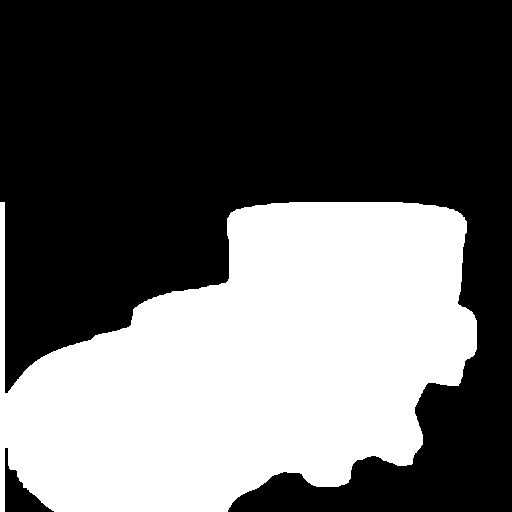} &
    \includegraphics[width=0.115\linewidth]{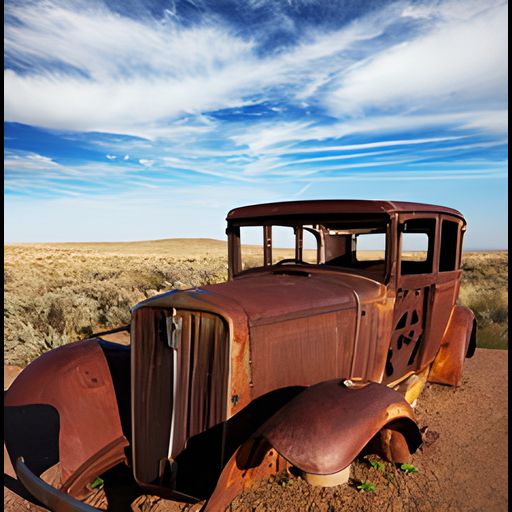} & \includegraphics[width=0.115\linewidth]{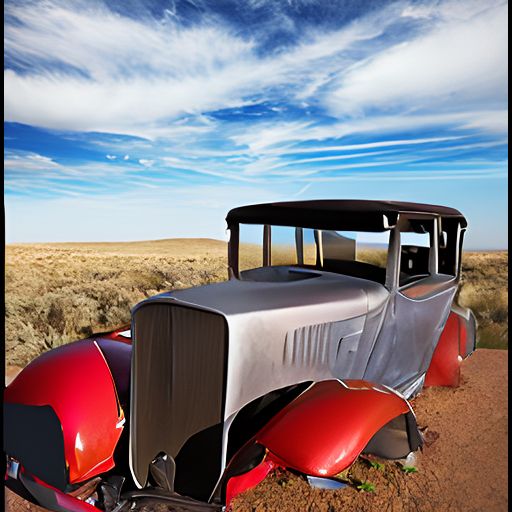} & \includegraphics[width=0.115\linewidth]{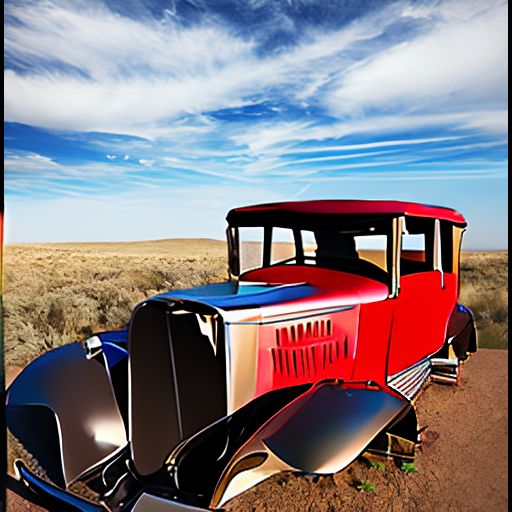} & \includegraphics[width=0.115\linewidth]{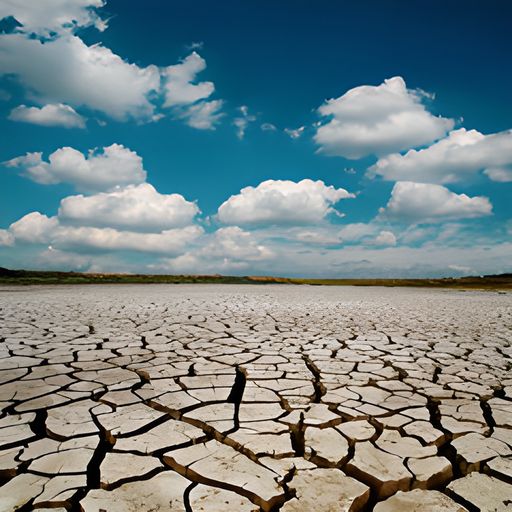} & \includegraphics[width=0.115\linewidth]{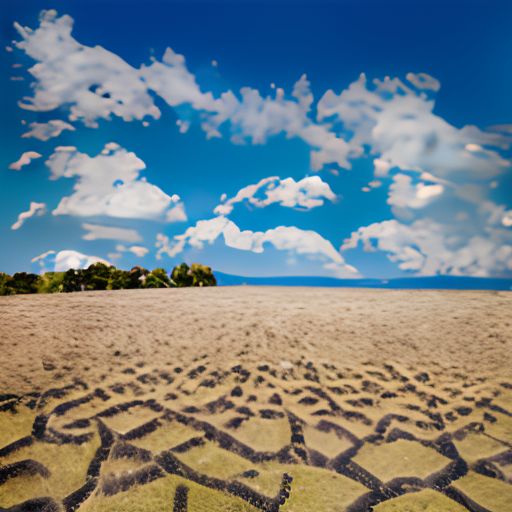} & \includegraphics[width=0.115\linewidth]{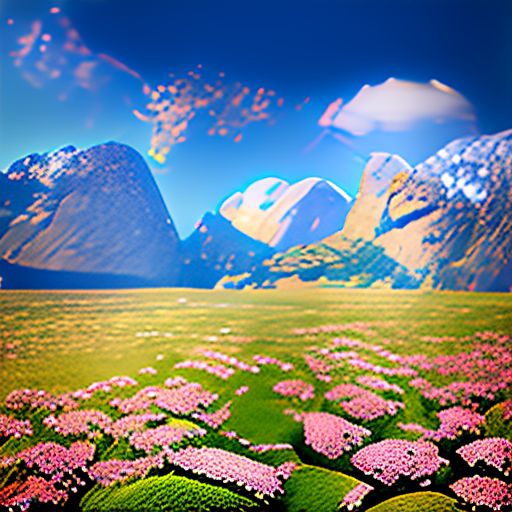} \\

    \multicolumn{4}{c}{low appeal $\rightarrow$ high appeal} & \multicolumn{3}{c}{low appeal $\rightarrow$ high appeal} \\ \hline\hline

    Mask & Input & 0.5 $z_V^-$ & 1.0 $z_V^-$ & Input & 0.5 $z_L^-$ & 1.0 $z_L^-$\\

    \includegraphics[width=0.115\linewidth]{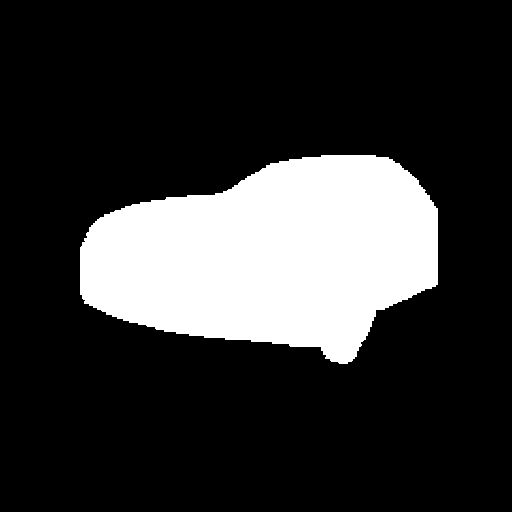} &
    \includegraphics[width=0.115\linewidth]{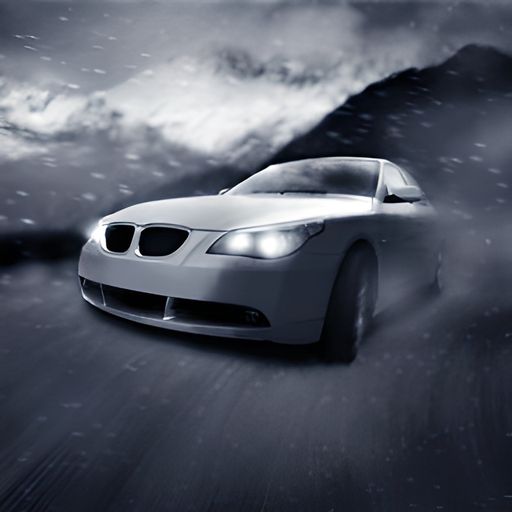} & \includegraphics[width=0.115\linewidth]{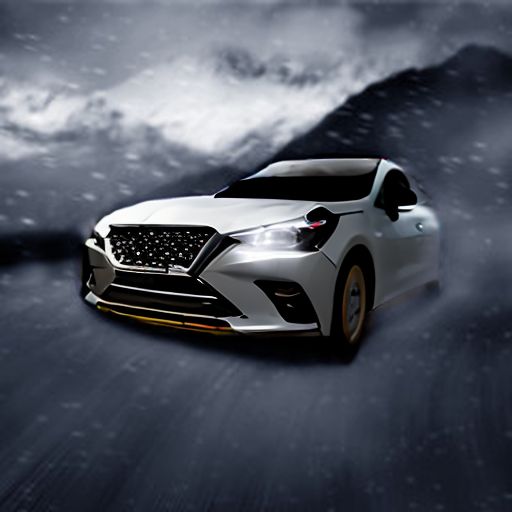} & \includegraphics[width=0.115\linewidth]{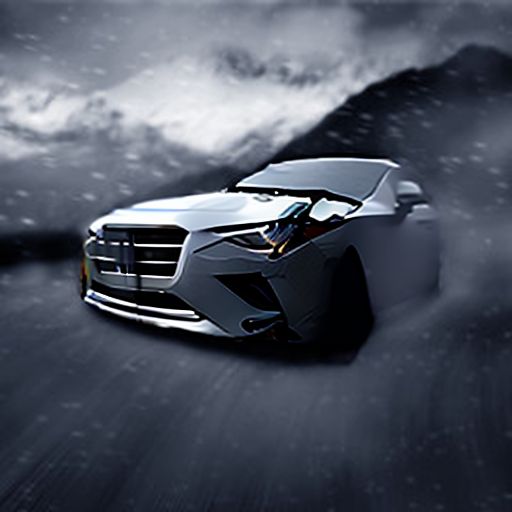} & \includegraphics[width=0.115\linewidth]{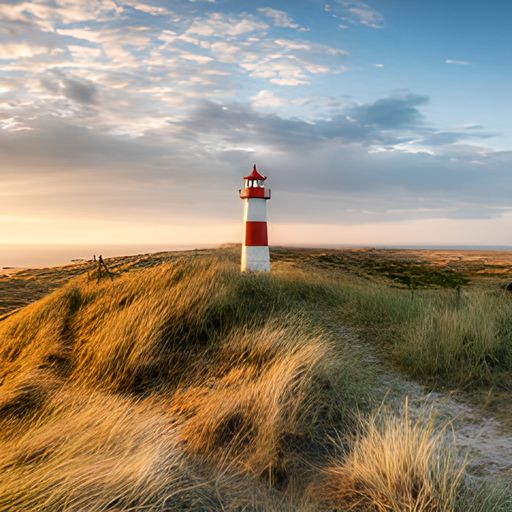} & \includegraphics[width=0.115\linewidth]{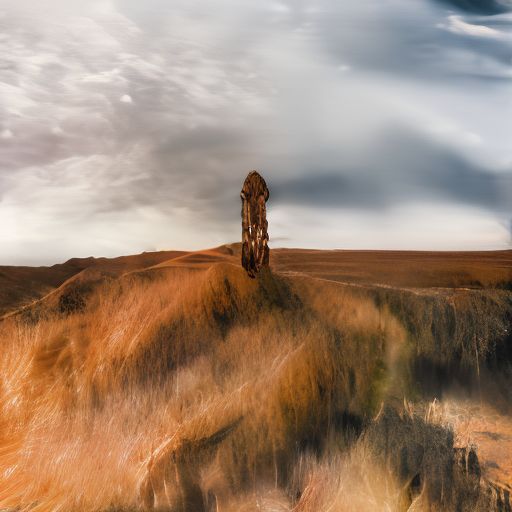} & \includegraphics[width=0.115\linewidth]{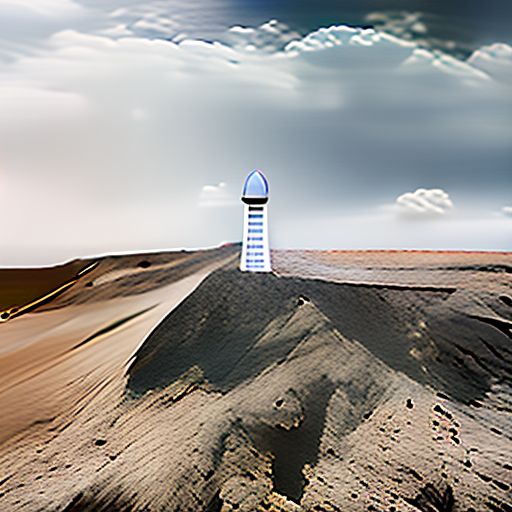} \\

    \includegraphics[width=0.115\linewidth]{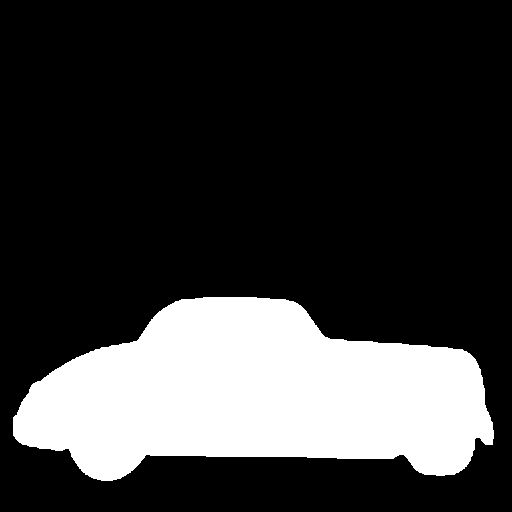} &
    \includegraphics[width=0.115\linewidth]{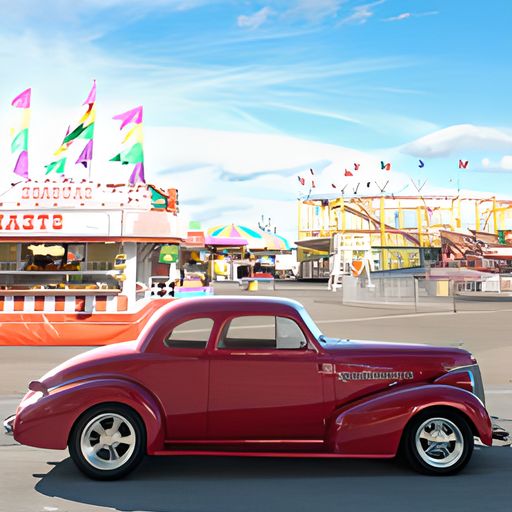} & \includegraphics[width=0.115\linewidth]{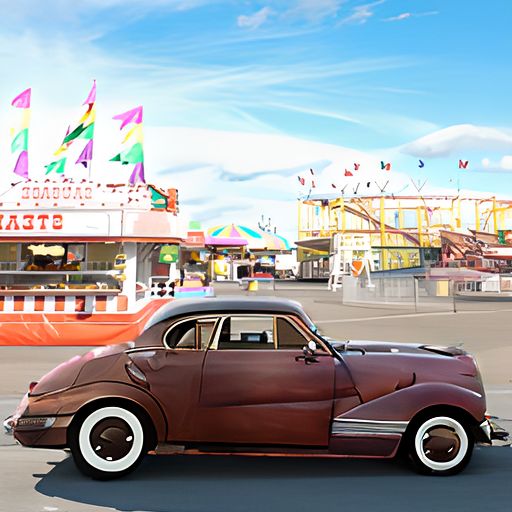} & \includegraphics[width=0.115\linewidth]{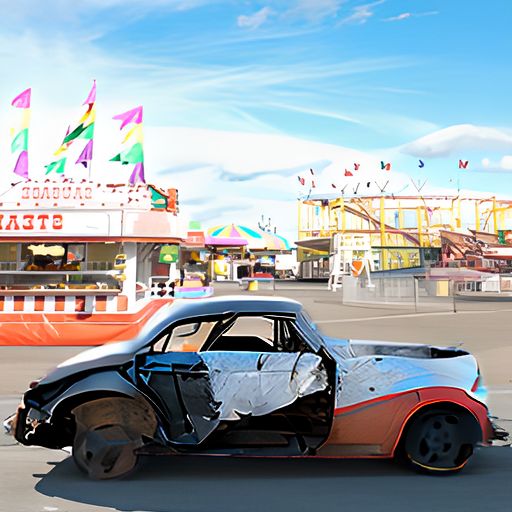} & \includegraphics[width=0.115\linewidth]{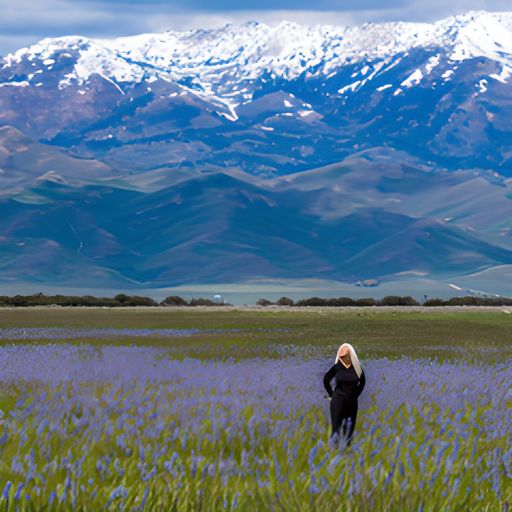} & \includegraphics[width=0.115\linewidth]{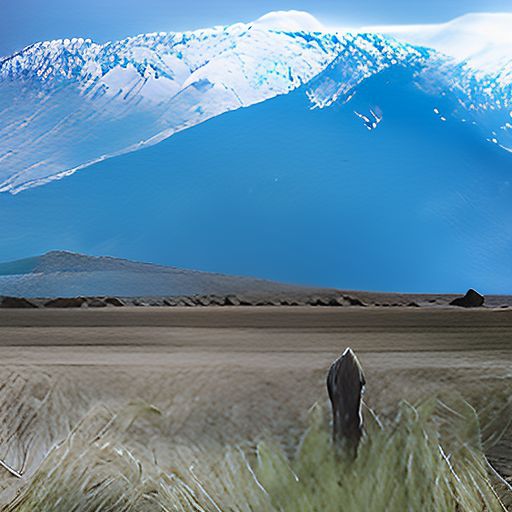} & \includegraphics[width=0.115\linewidth]{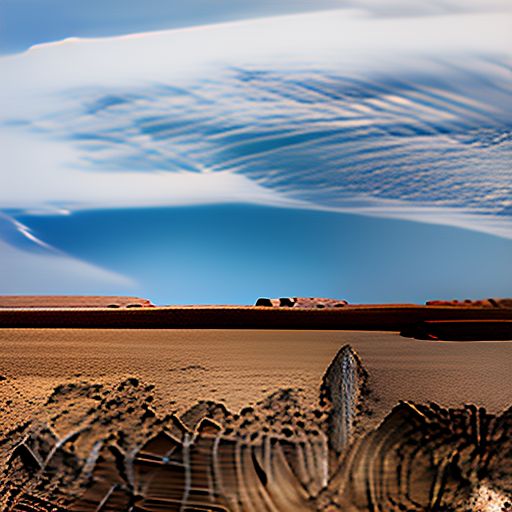} \\
     \multicolumn{4}{c}{high appeal $\rightarrow$ low appeal} & \multicolumn{3}{c}{high appeal $\rightarrow$ low appeal}
    \end{tabular}
    \caption{We trained embeddings $z_V^+$/$z_L^+$ and $z_V^-$/$z_L^-$ for vehicles (left) and landscapes (right) to adjust image appeal with different weights in the domain-relevant area (``Mask'' for vehicles; for landscapes, we consider all pixels to be relevant) and created synthetic datasets samples. Although these results are not equivalent to the final output of our image enhancer (which operates with respect to the appeal heatmaps from our predictor and generates more consistent results), we can observe successful appeal changes between images necessary for training our models.}
    \label{fig:generalize_to_vehicles_landscapes}
\end{figure}

\noindent \textsc{\textbf{Room}}: Search queries were generated from the following sets of words:
\begin{itemize}
    \item $\set{N}_R = \{`$`bathroom,'' ``bedroom,'' ``kitchen,`` ``living\ room,'' ``room''$\}$
    \item $\set{A}_R^+ = \{$``interior''$\}$
    \item $\set{A}_R^- = \{$``abandoned,'' ``dirty''$\}$
\end{itemize}
Note that we didn't include ``clean'' in $\set{A}_R^+$ because the word can be interpreted as a verb, so images focusing on people cleaning rooms will be returned, which is outside the room interior domain. We collect 261,907 image thumbnails and obtain 76,387 images of size $512 \times 512$ after filtering and preprocessing. Likewise, we select 100 images to generate embeddings using textual inversion, and 1,000 images with balanced content appeal levels and room types to create the synthetic dataset. For each image, we use it to generate five different augmentations. For each augmentation, we change its content appeal level and generate three different images. In the end, we generate 15,000 images for our synthetic dataset $\set{S}_R$, leaving us with 75,287 images for the final dataset $\set{I}_R$.

We present image examples from each dataset with various levels of content appeal and image aesthetics in \cref{fig:dataset_samples}. Specifically, we uniformly stride one out of each 100 images in each dataset we created by image indices and estimate their image aesthetics scores using three popular open-sourced IAA baselines: DIAA, MPADA, and NIMA. We denote images with appeal scores in the $25^{th}$ and $75^{th}$ percentile in their respective datasets to have low and high content appeal respectively. Images with aesthetics scores in the $25^{th}$ and $75^{th}$ percentiles across all three IAA baselines have low and high aesthetics respectively. We can see that the content appeal and image aesthetics of an image may be very different. 

\subsection{Dataset creation across image domains\label{sec:dataset_creation_generalizability}}

\begin{figure}[t]
    \centering
    \begin{tabular}{ccccc@{\hspace{0.2in}}c}
         &  DIAA & MPADA & NIMA & DIAA-IC \\
     Food    & \includegraphics[width=0.2\linewidth]{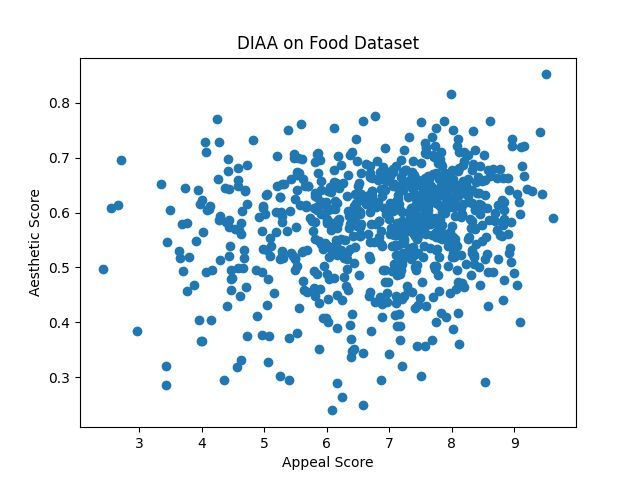}
     & 
    \includegraphics[width=0.2\linewidth]{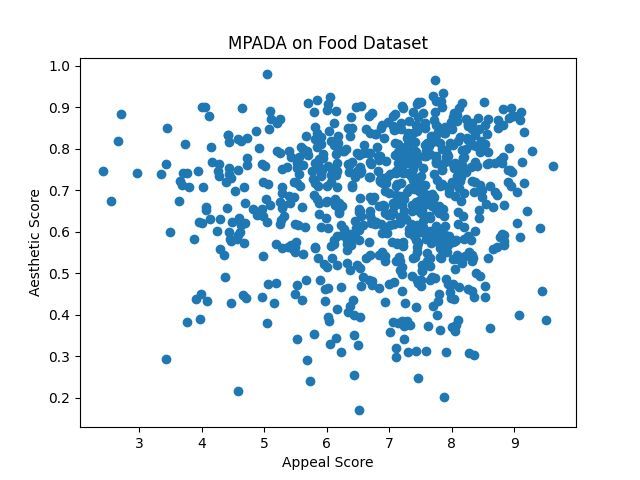}
    & 
    \includegraphics[width=0.2\linewidth]{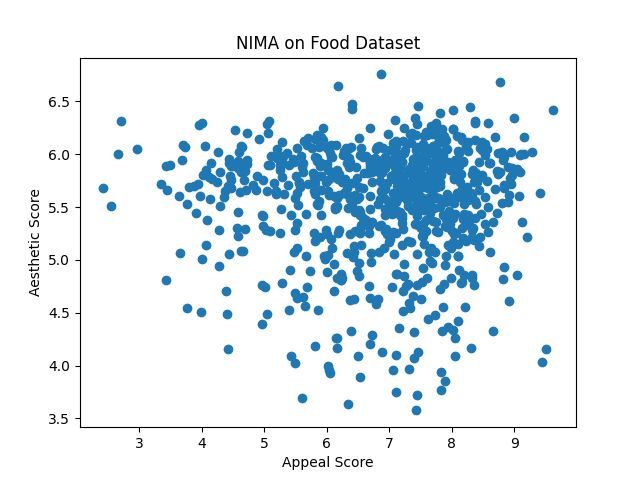} 
    & \includegraphics[width=0.2\columnwidth]{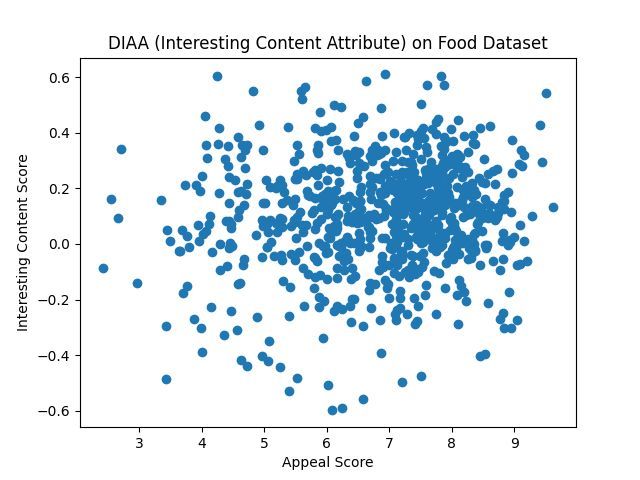} \\  
     Room    & \includegraphics[width=0.2\linewidth]{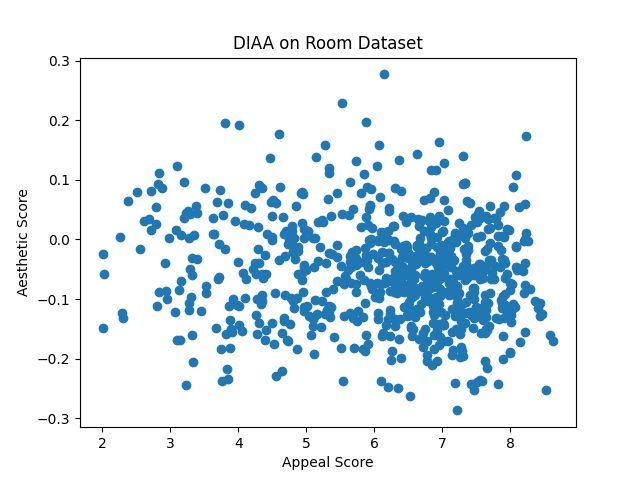}
     & 
    \includegraphics[width=0.2\linewidth]{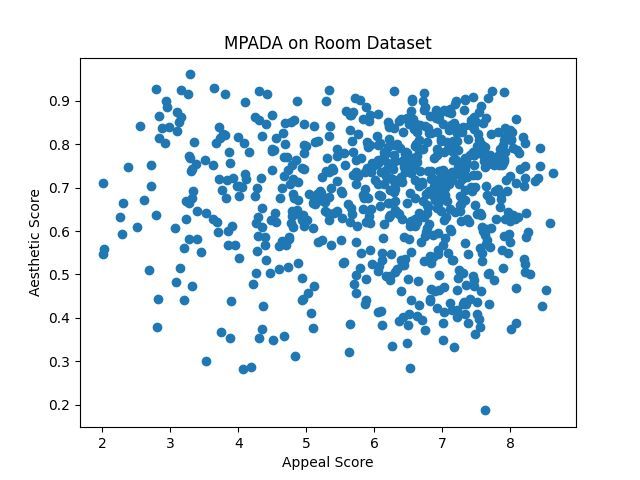}
    & 
    \includegraphics[width=0.2\linewidth]{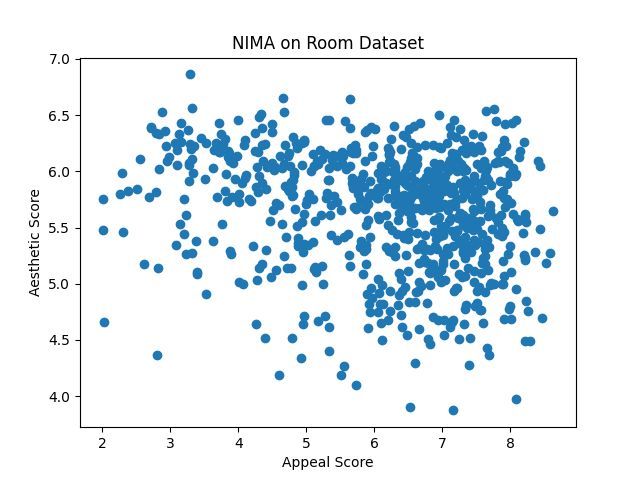} 
    & \includegraphics[width=0.2\columnwidth]{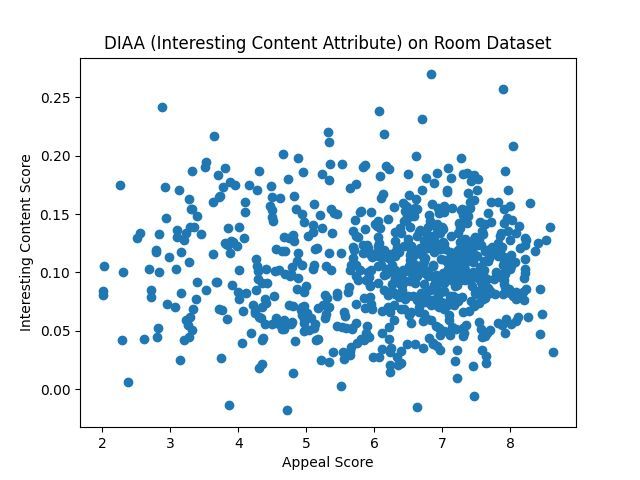}
    \end{tabular}
    \caption{\textbf{Correlation between content appeal and image aesthetics.} We visualize the relationship between predictions from our estimator and from three IAA models on subsets of our two datasets. We can see there is little correlation between content appeal and image aesthetics, suggesting they are indeed different image metrics. There is also little correlation between our content appeal predictions and DIAA ``interesting content'' (DIAA-IC) predictions, meaning that the latter cannot be readily substituted by the former.}
    \label{fig:appeal_vs_aesthetic}
\end{figure}

\begin{figure*}
    \centering
    \includegraphics[width=0.95\textwidth]{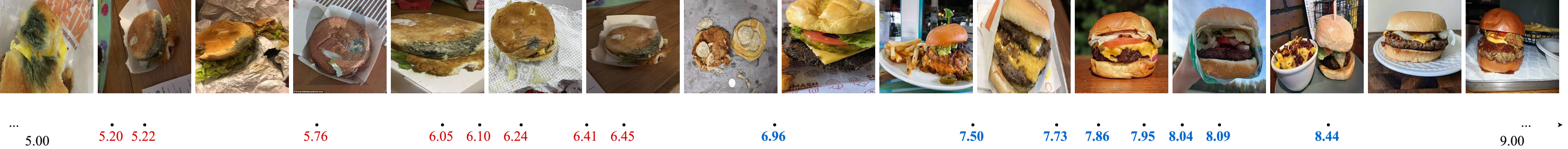}
    \vfill
    \includegraphics[width=0.95\textwidth]{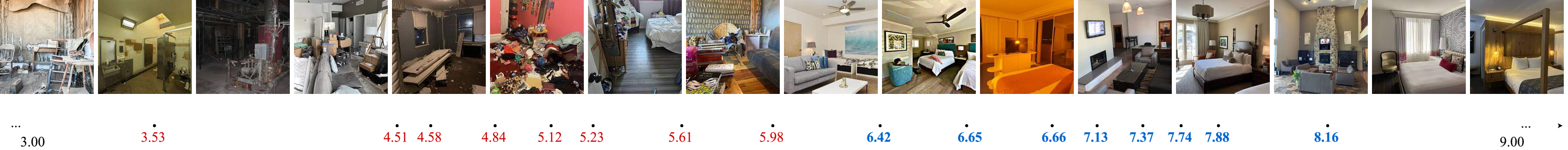}
    \caption{\textbf{Generalizability of content appeal estimator on amateur-taken images.} Although being trained on professionally-taken images, the estimator can be generalized to amateur-taken images during run time and accurately distinguish appealing (predicted scores in blue and bold) and unappealing (predicted scores in red and boxes) images.}
    \label{fig:generalize_to_amateur}
\end{figure*}

\begin{figure}
    \centering
    \scriptsize
    \begin{tabular}{cc@{\hspace{0.2in}}cc}
     w/o distortion &  w/ distortion & w/o distortion &  w/ distortion \\
     \includegraphics[width=0.115\linewidth]{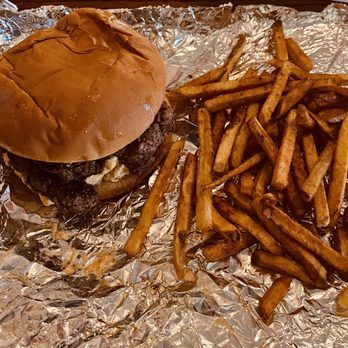} & \includegraphics[width=0.115\linewidth]{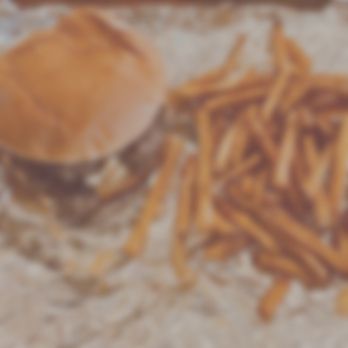} & \includegraphics[width=0.17\linewidth]{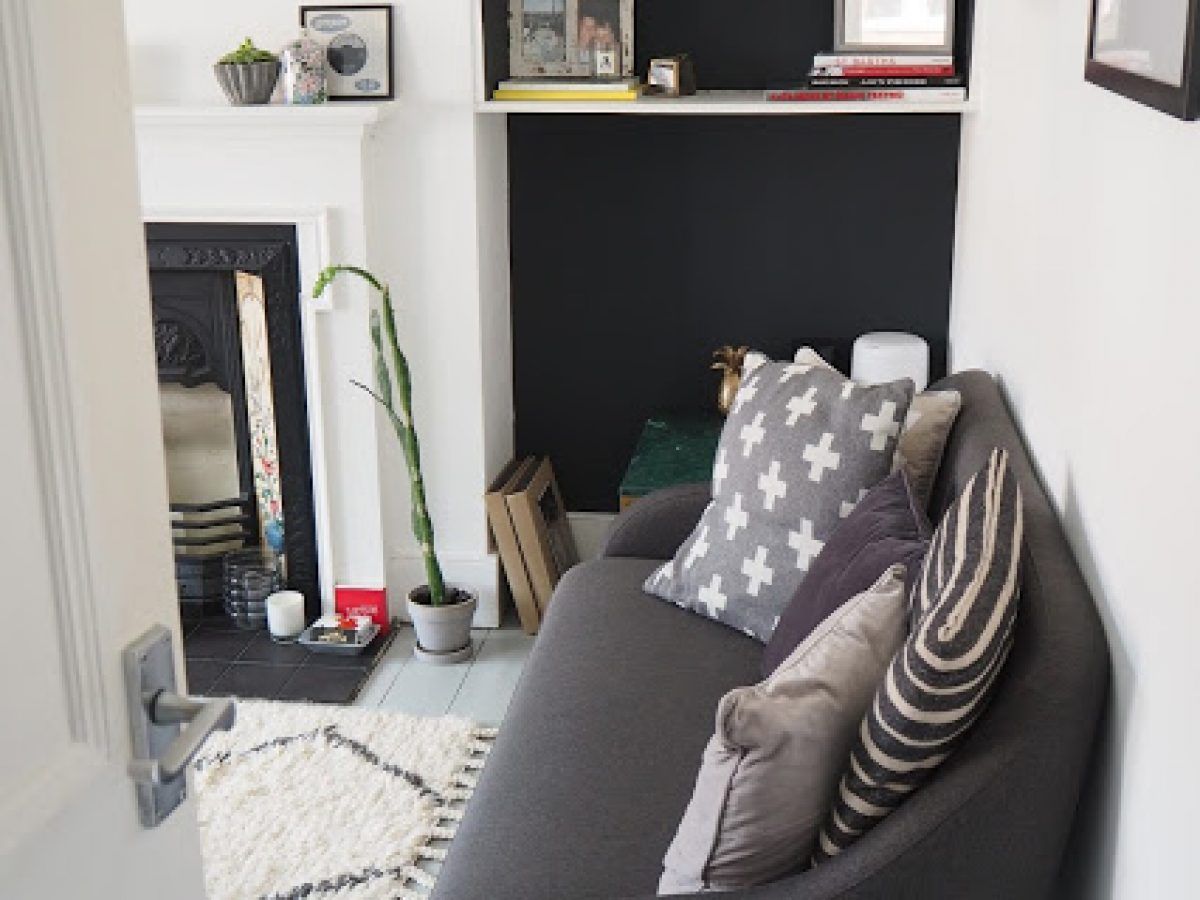} & \includegraphics[width=0.17\linewidth]{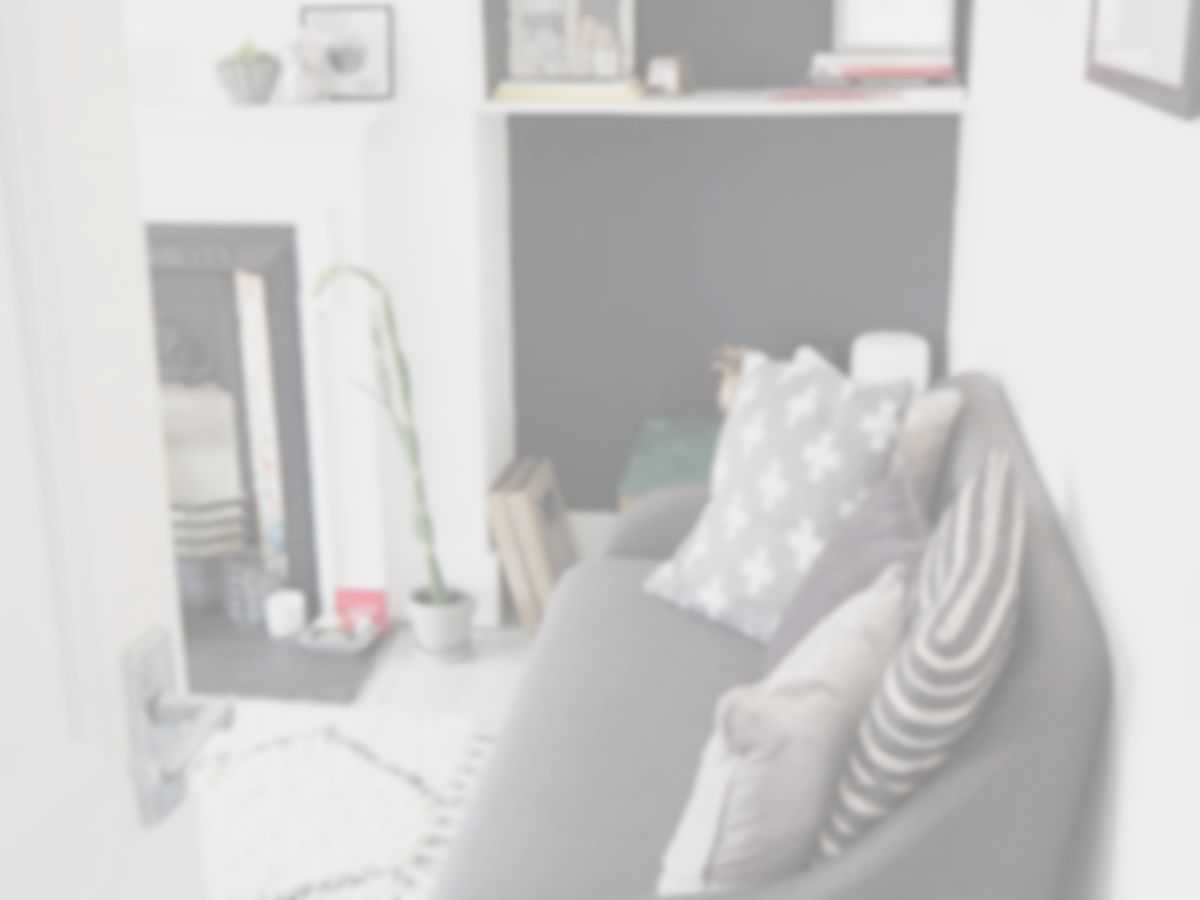} \\
     7.80 & 7.63 (-0.17) & 7.70 & 6.92 (-0.78)
     \end{tabular}
    \caption{When two images have the same content, technical distortions have a negative impact on content appeal scores (predicted by our model and shown below each image) as aesthetics and content appeal are not orthogonal axes.}
    \label{fig:effect_of_techinical_distortion}
\end{figure}

\ourwork{} can be easily adapted to different domains, of which we demonstrate two new ones here: \emph{vehicles} and \emph{landscapes}, where we illustrate the process of creating synthetic datasets. This involves gathering 50 appealing and 50 unappealing images for each domain, which are used to train appealing and unappealing textual inversion embeddings, $z_V^+$/$z_L^+$ and $z_V^-$/$z_L^-$, following the same methodology used for food and rooms. This allowed us to manipulate the relative appeal of images to generate synthetic datasets (\Cref{fig:generalize_to_vehicles_landscapes}). As can be seen, our method does a reasonable job at increasing/decreasing image appeal in these very different domains.

\section{Image Content Appeal Estimator Details\label{sec:estimator_details}}

\subsection{IAA baseline comparison\label{sec:iaa_baseline_comparison}}

To further show the difference between content appeal and image aesthetics, we visualize the correlation between them (\cref{fig:appeal_vs_aesthetic}) on above strided images, where we observe little correlation between content appeal and image aesthetics (for coefficient values, please refer to the paper). Furthermore, we visualize the relationship between content appeal and DIAA ``interesting content'' attribute (\cref{fig:appeal_vs_aesthetic} Row.4), where little correlation is presented as well. This means that DIAA `interesting content'' attribute cannot substitute \IPA{} either.

\subsection{Performance on amateur-taken images\label{sec:amateur_generalizability}}

Although our estimator is trained on professionally-taken images, it can be generalized to amateur-taken images during inference time and accurately distinguish content appealing (predicted scores in blue and bold) and content-unappealing (predicted scores in red and boxes) images (\cref{fig:generalize_to_amateur}).

\subsection{Effect of technical distortions\label{sec:technical_distortion_effect}} 

When two images have the same content, their content appeal should be affected by technical distortions, which is correctly reflected in our models (\cref{fig:effect_of_techinical_distortion}). However, these distortions should not overshadow the inherent appeal of the image content. As illustrated in Fig. 1 and \cref{fig:generalize_to_amateur}, images with unappealing content yet high aesthetic quality still receive low content appeal scores.

\section{Content Appeal Enhancer Details\label{sec:content_appeal_enhancer_details}}

\subsection{Implementation details}
We use Stable Diffusion v2.1 inpainting with depth-guided ControlNet for image content appeal enhancement. Specifically, here are some parameter values we use:
\begin{itemize}
    \item prompt: ``<$z_D^+$> <object\_type>''
    \item negative prompt: ``out of frame, lowres, text, error, cropped, worst quality, low quality, jpeg artifacts, ugly, duplicate, morbid, mutilated, out of frame, extra fingers, mutated hands, poorly drawn hands, poorly drawn face, mutation, deformed, blurry, dehydrated, bad anatomy, bad proportions, extra limbs, cloned face, disfigured, gross proportions, malformed limbs, missing arms, missing legs, extra arms, extra legs, fused fingers, too many fingers, long neck, username, watermark, signature,''
    \item ``Sampling method'': ``DPM++ 2M Karras''
    \item ``CGF scale'' $= 7$
    \item ``denoising strength'' $= 0.6$
    \item ControlNet ``Preprocessor'': ``depth\_midas'',
\end{itemize}
where the prompt is constructed by concatenating the appealing embedding with the type of the object in the input image (e.g. burger, kitchen), and ControlNet preprocessor use MiDaS [Ranftl et al. 2020] to estimate a depth map from the input image. 

Note that not all phrases in the negative prompt are directly related to the image domain the input image is from. Instead, we use this generic negative prompt for all image domains.

\subsection{More results and ablation studies\label{sec:image_content_appeal_enhancement}}

\begin{figure*}
\scriptsize
    \centering
    \begin{tabular}{cccc@{\hspace{0.1in}}cccc}
     Input & Enhanced & $M_F^H$ & Depth &  Input & Enhanced & $M_F^H$ & Depth \\
    \includegraphics[width=0.11\linewidth]{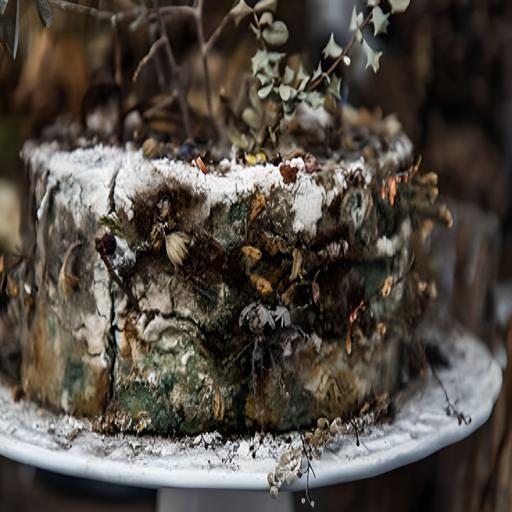} & \includegraphics[width=0.11\linewidth]{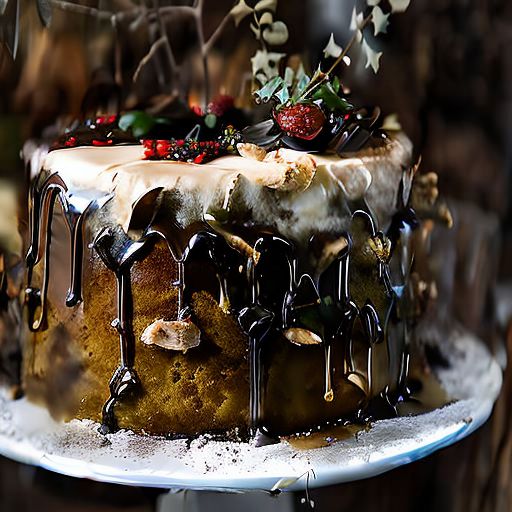} &  \includegraphics[width=0.11\linewidth]{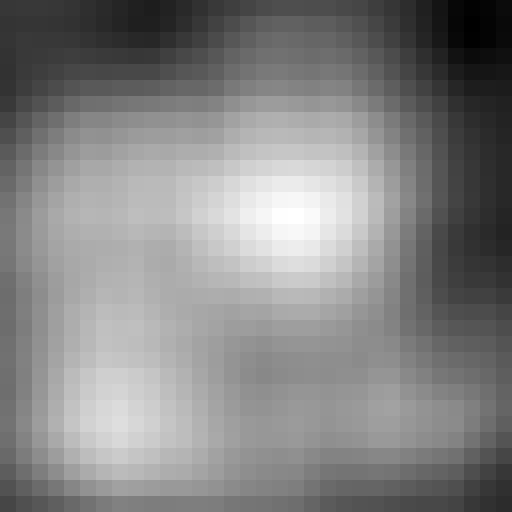} & \includegraphics[width=0.11\linewidth]{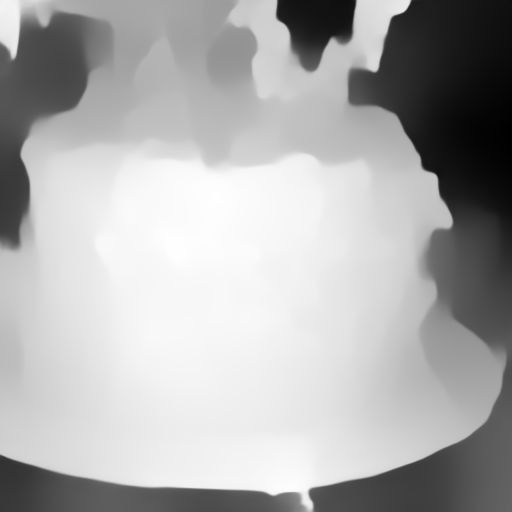} & 
    \includegraphics[width=0.11\linewidth]{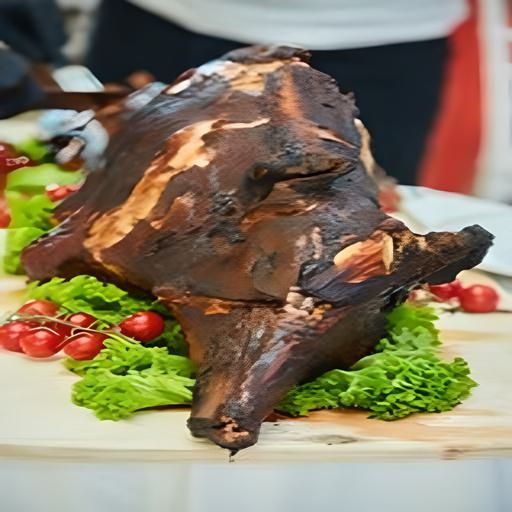} & \includegraphics[width=0.11\linewidth]{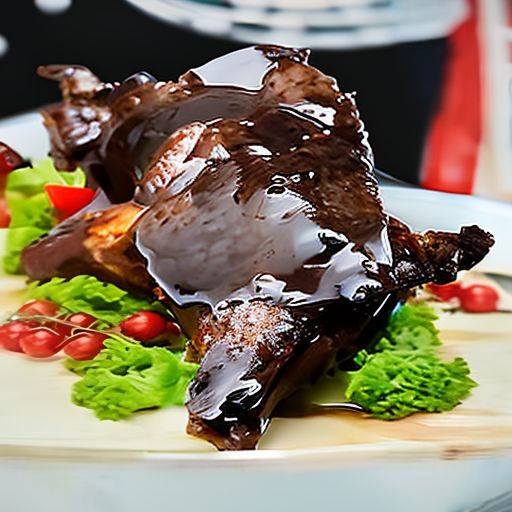} &  \includegraphics[width=0.11\linewidth]{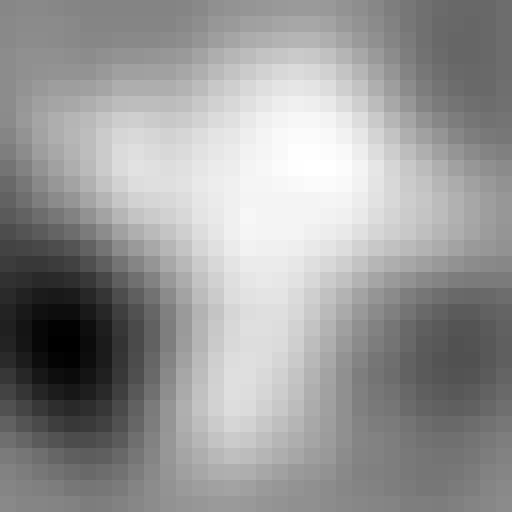} & \includegraphics[width=0.11\linewidth]{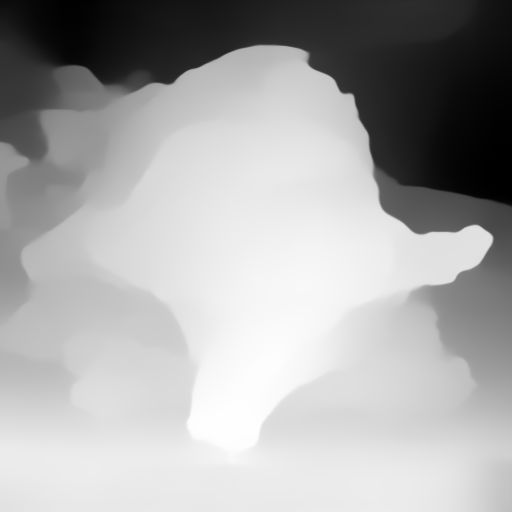} \\
    4.77 & 7.30 (+2.53) & & & 6.24 & 7.35 (+1.11) \\
    \includegraphics[width=0.11\linewidth]{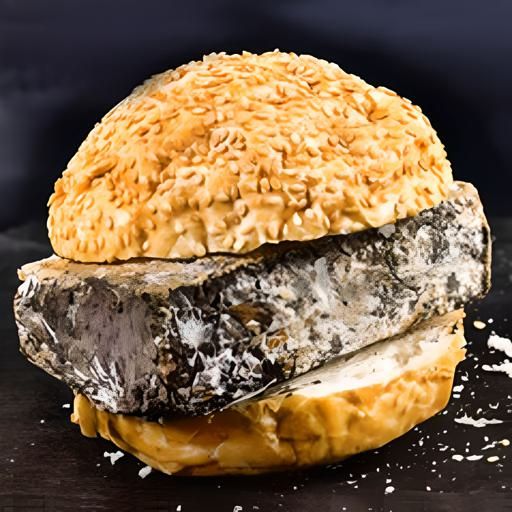} & \includegraphics[width=0.11\linewidth]{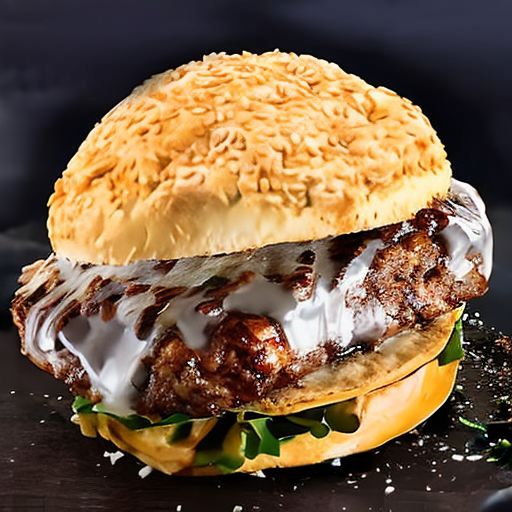} &  \includegraphics[width=0.11\linewidth]{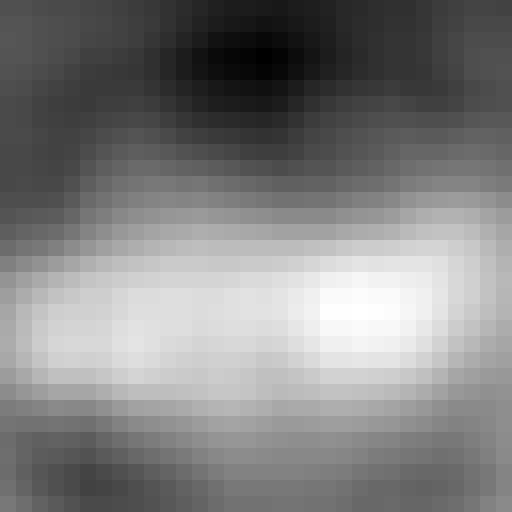} & \includegraphics[width=0.11\linewidth]{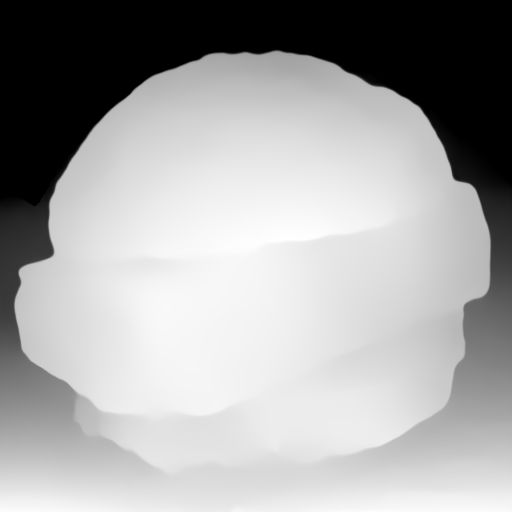} & 
    \includegraphics[width=0.11\linewidth]{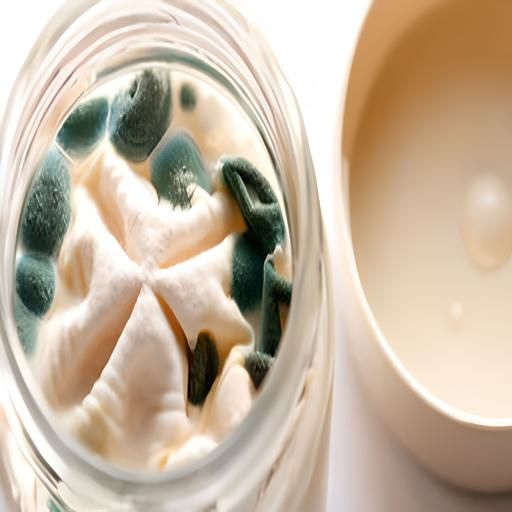} & \includegraphics[width=0.11\linewidth]{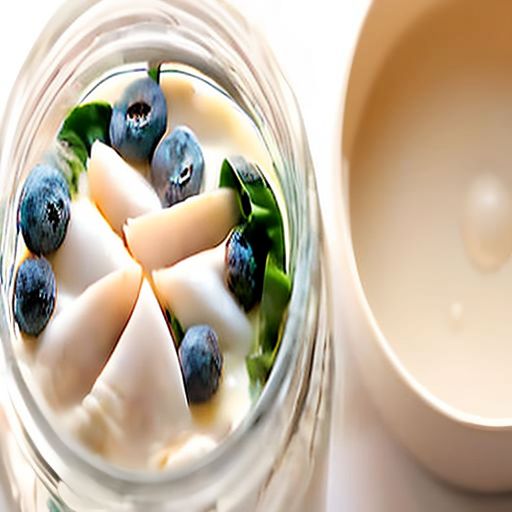} &  \includegraphics[width=0.11\linewidth]{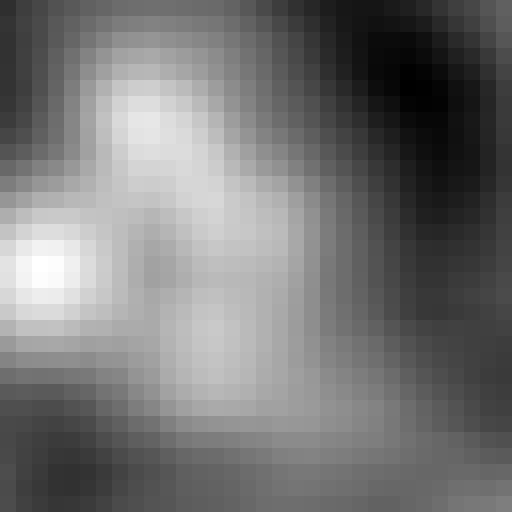} & \includegraphics[width=0.11\linewidth]{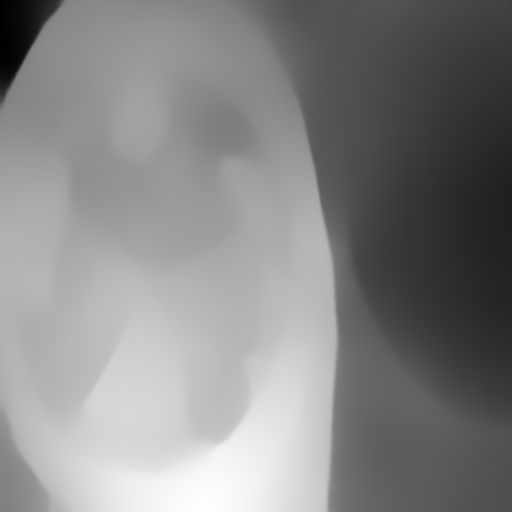} \\
    5.35 & 7.82 (+2.47) & & & 6.11 & 7.39 (+1.28) & & \\
    \includegraphics[width=0.11\linewidth]{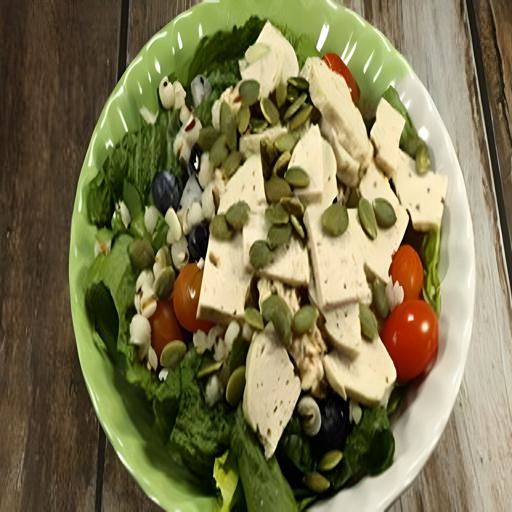} & \includegraphics[width=0.11\linewidth]{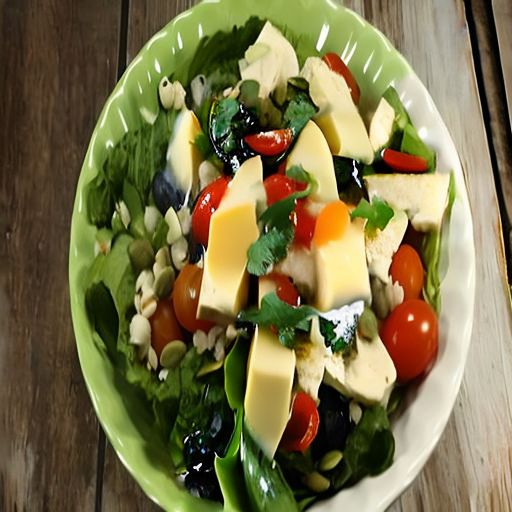} &  \includegraphics[width=0.11\linewidth]{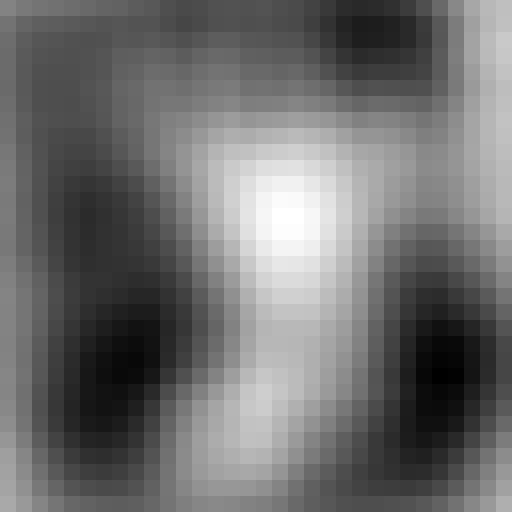} & \includegraphics[width=0.11\linewidth]{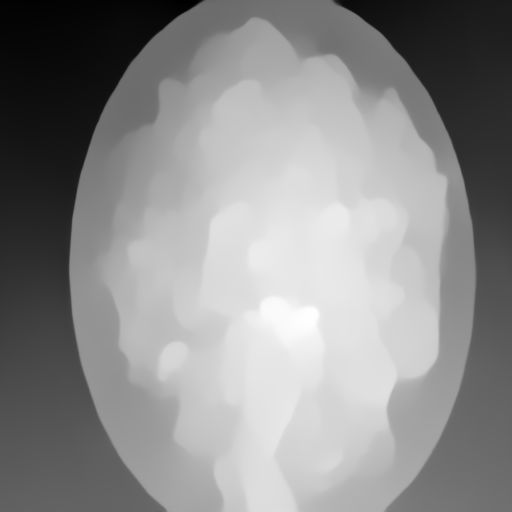} & 
    \includegraphics[width=0.11\linewidth]{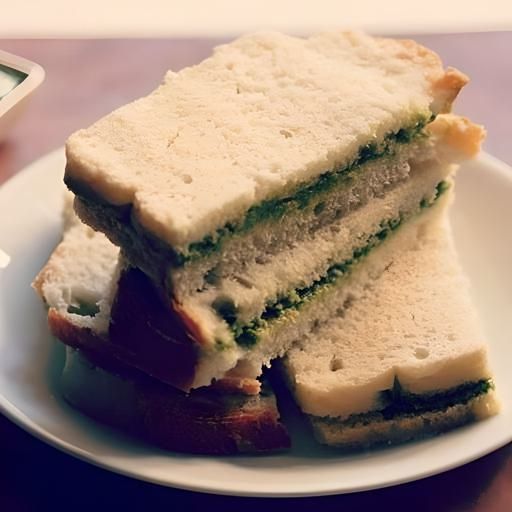} & \includegraphics[width=0.11\linewidth]{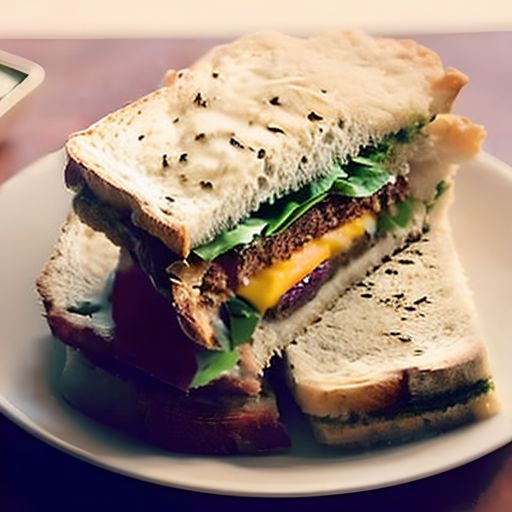} &  \includegraphics[width=0.11\linewidth]{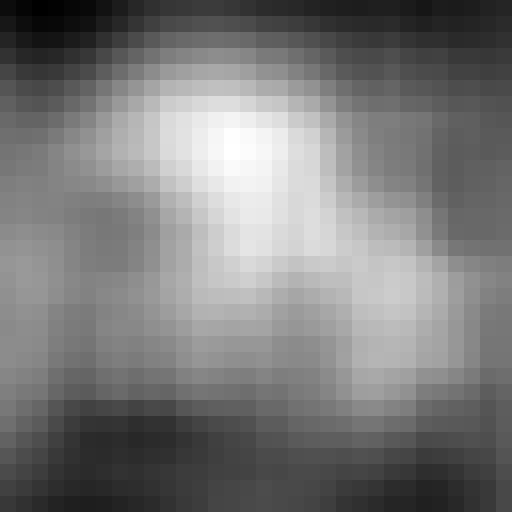} & \includegraphics[width=0.11\linewidth]{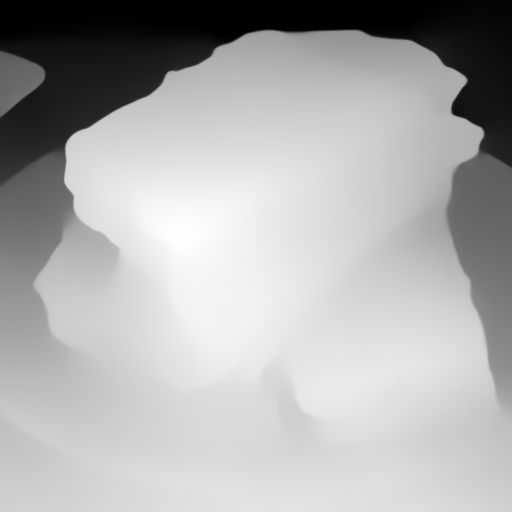} \\
    7.33 & 8.13 (+0.80) & & & 6.67 & 7.21 (+0.54) & & \\
    \includegraphics[width=0.11\linewidth]{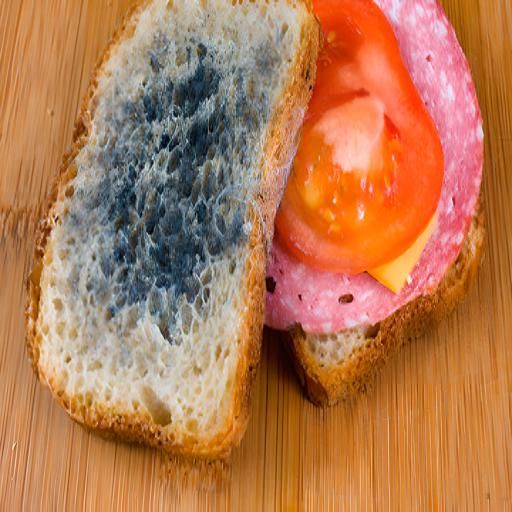} & \includegraphics[width=0.11\linewidth]{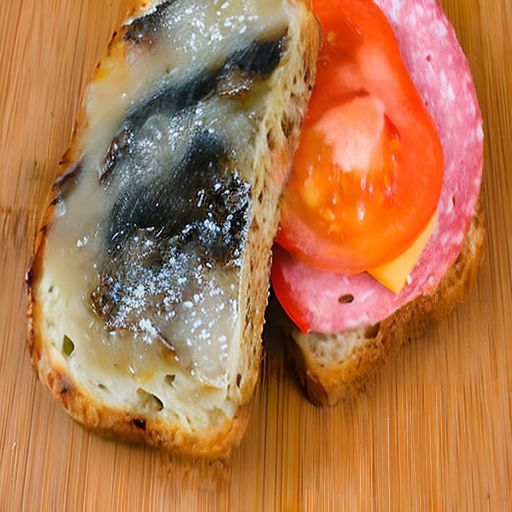} &  \includegraphics[width=0.11\linewidth]{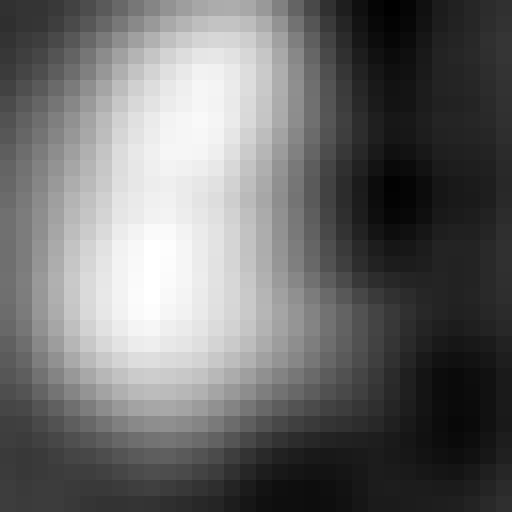} & \includegraphics[width=0.11\linewidth]{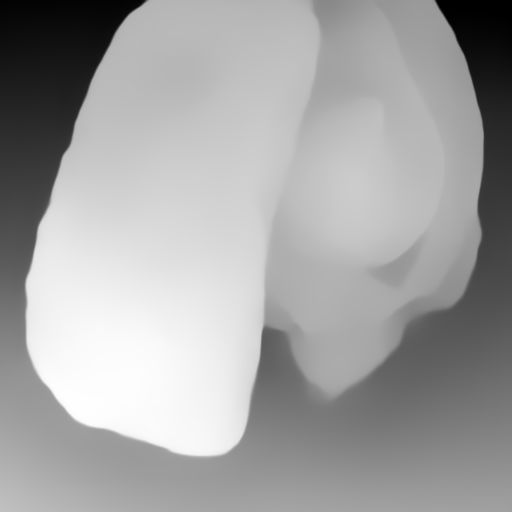} & 
    \includegraphics[width=0.11\linewidth]{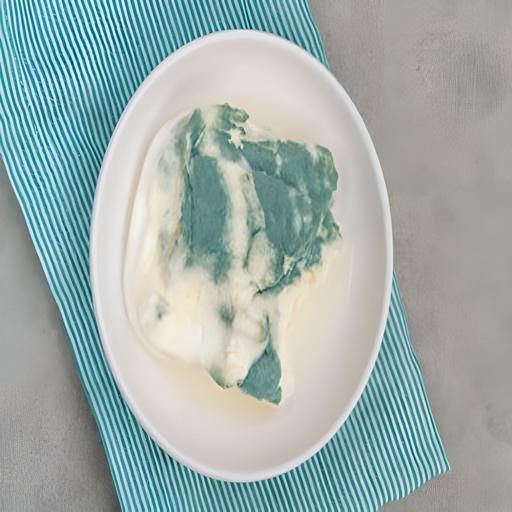} & \includegraphics[width=0.11\linewidth]{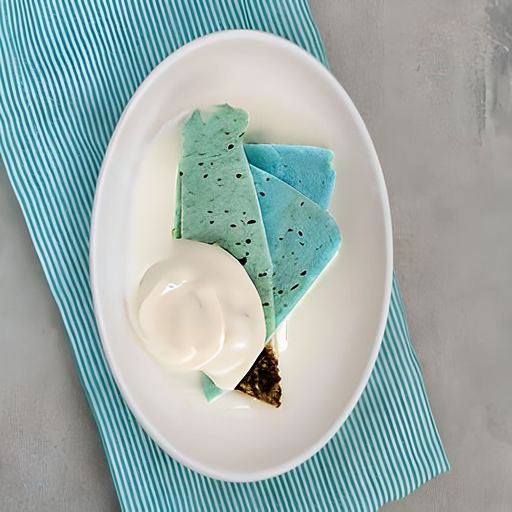} &  \includegraphics[width=0.11\linewidth]{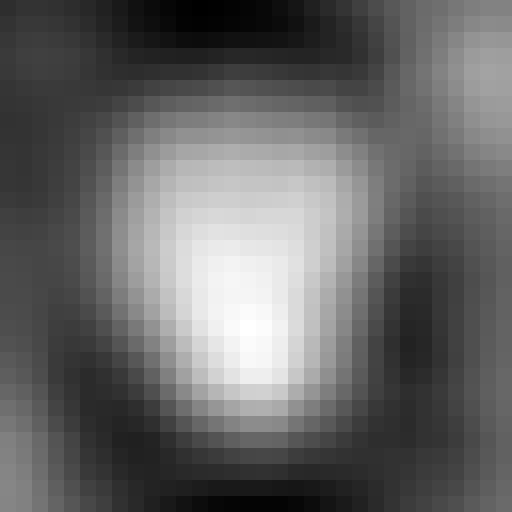} & \includegraphics[width=0.11\linewidth]{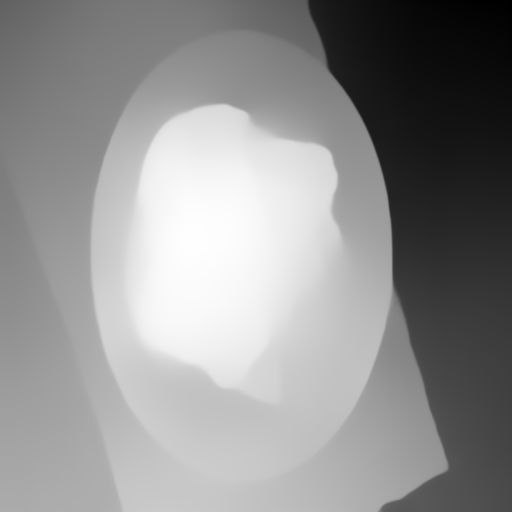} \\
    5.11 & 6.75 (+1.64) & & & 4.79 & 6.47 (+1.68) & & \\ \hline\hline & & & & & & & \\
    \includegraphics[width=0.11\linewidth]{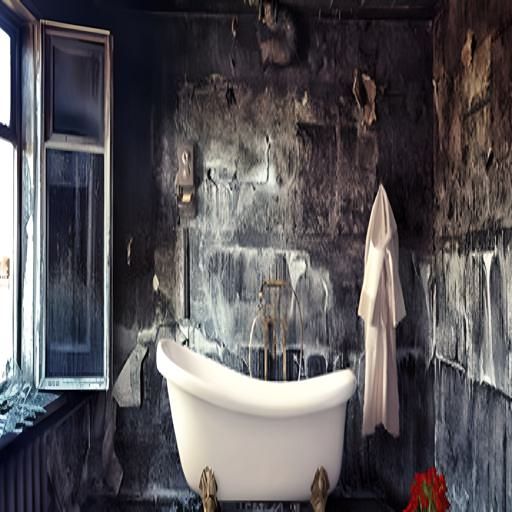} & \includegraphics[width=0.11\linewidth]{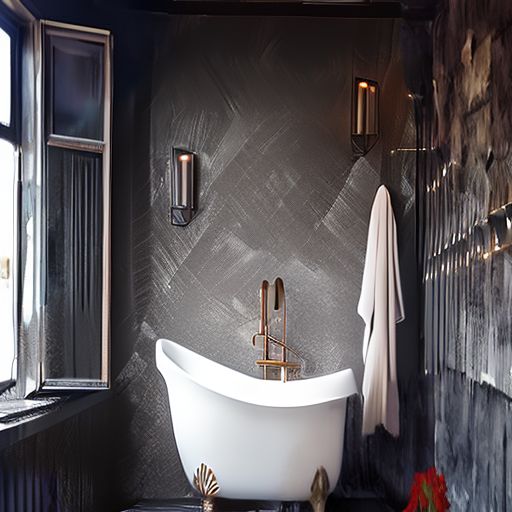} &  \includegraphics[width=0.11\linewidth]{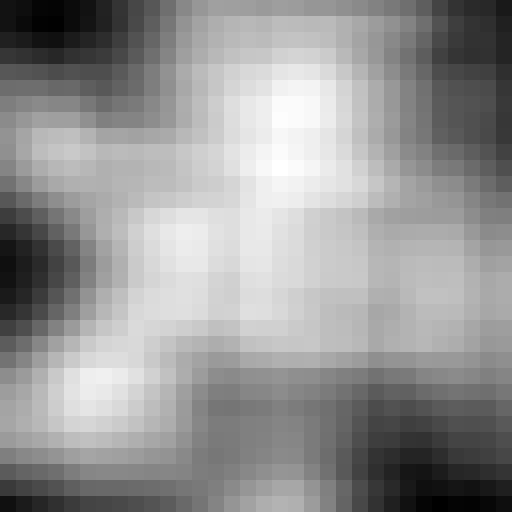} & \includegraphics[width=0.11\linewidth]{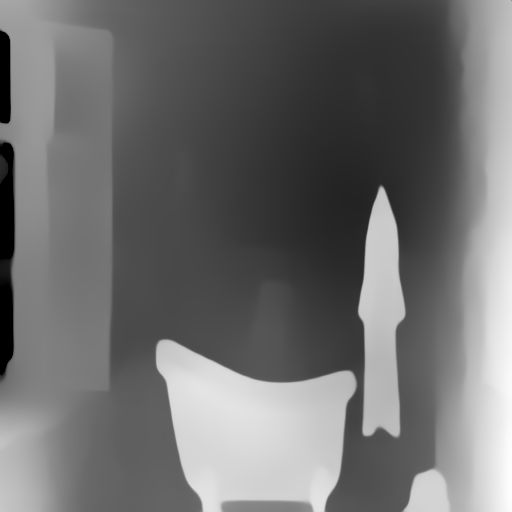} & 
    \includegraphics[width=0.11\linewidth]{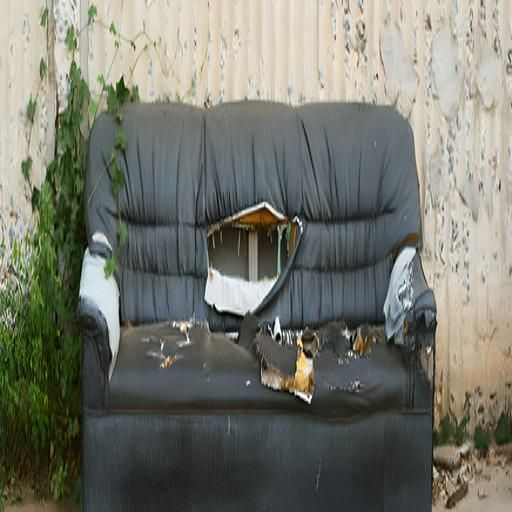} & \includegraphics[width=0.11\linewidth]{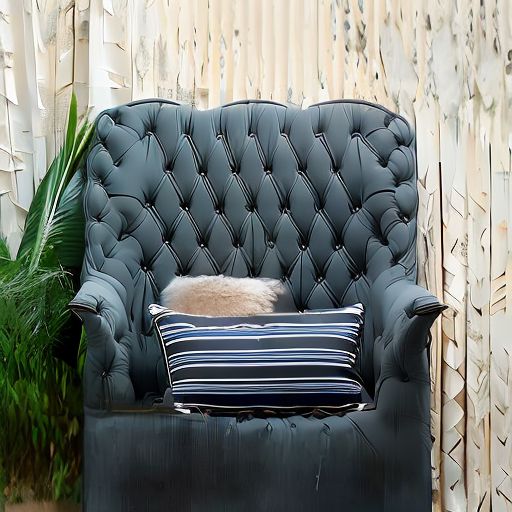} &  \includegraphics[width=0.11\linewidth]{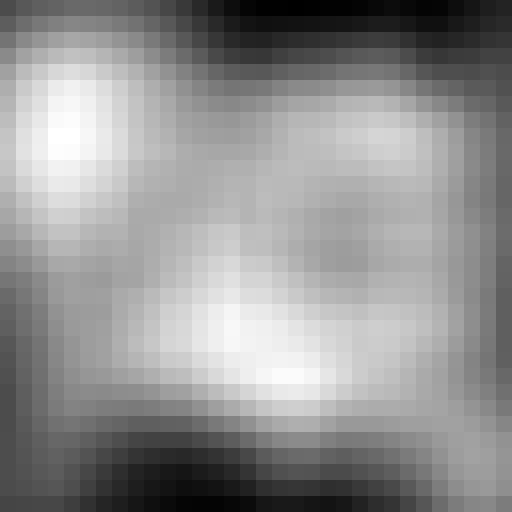} & \includegraphics[width=0.11\linewidth]{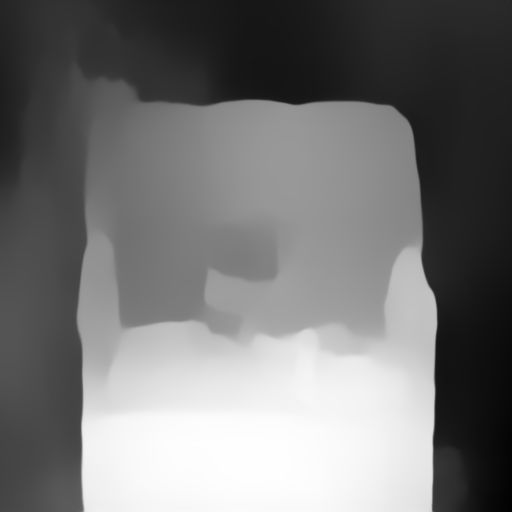} \\
    5.56 & 7.25 (+1.69) & & & 3.49 & 8.05 (+4.56) & & \\
    \includegraphics[width=0.11\linewidth]{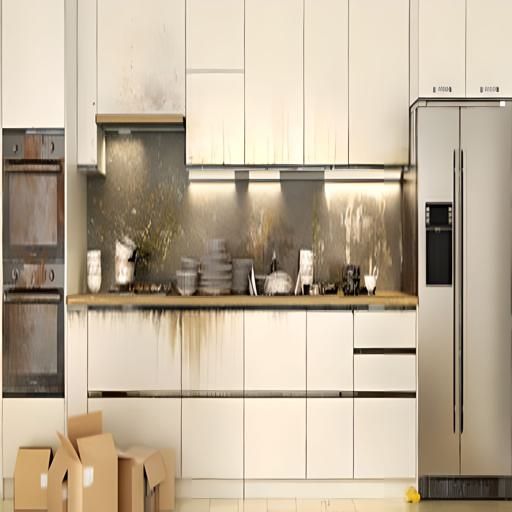} & \includegraphics[width=0.11\linewidth]{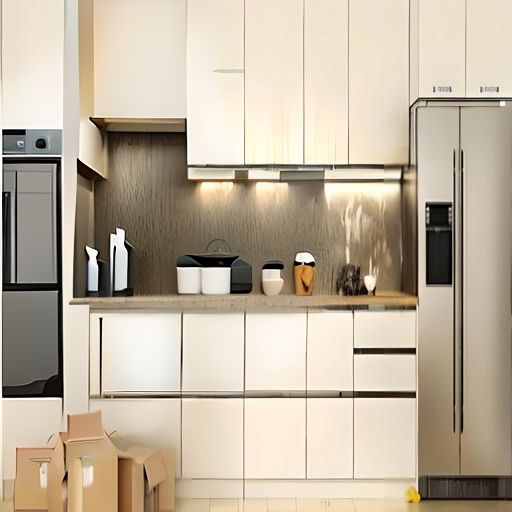} &  \includegraphics[width=0.11\linewidth]{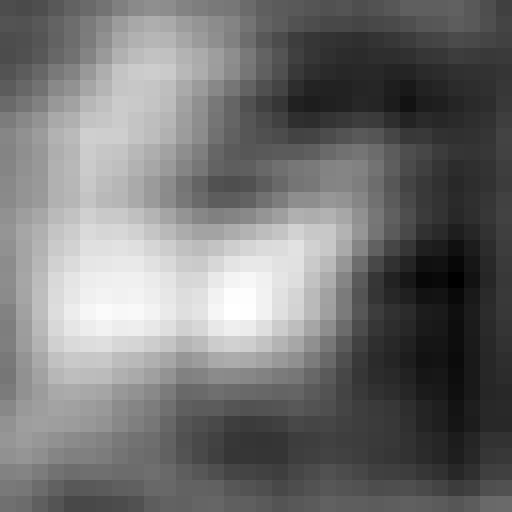} & \includegraphics[width=0.11\linewidth]{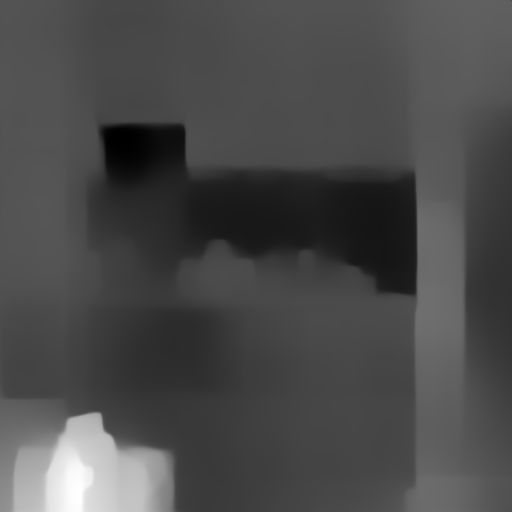} & 
    \includegraphics[width=0.11\linewidth]{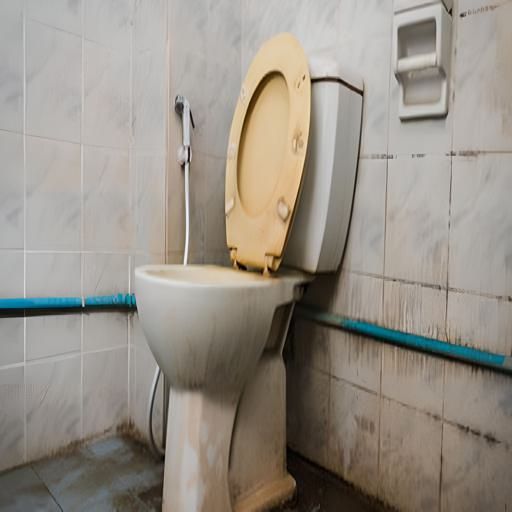} & \includegraphics[width=0.11\linewidth]{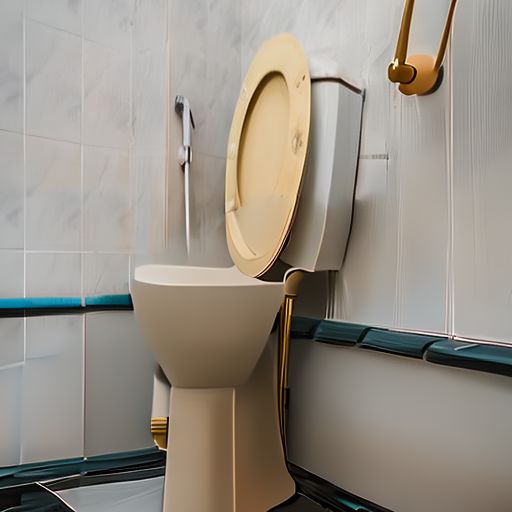} &  \includegraphics[width=0.11\linewidth]{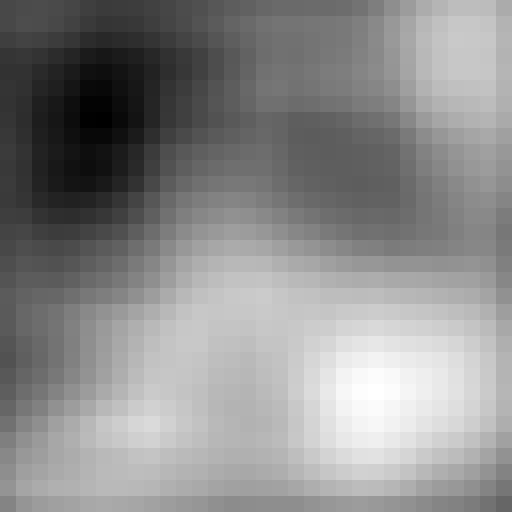} & \includegraphics[width=0.11\linewidth]{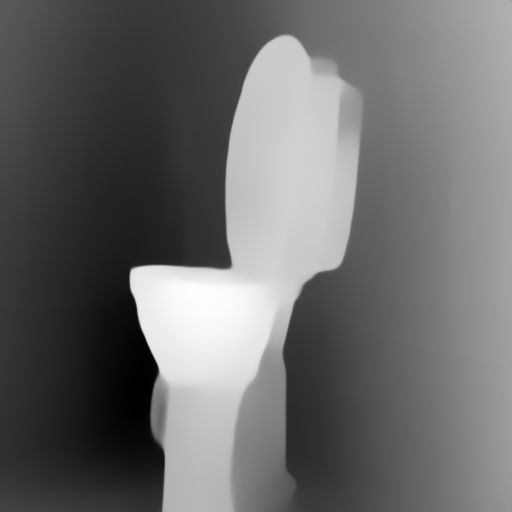} \\
    4.93 & 6.78 (+1.85) & & & 2.16 & 5.89 (+3.73) & & \\
    \includegraphics[width=0.11\linewidth]{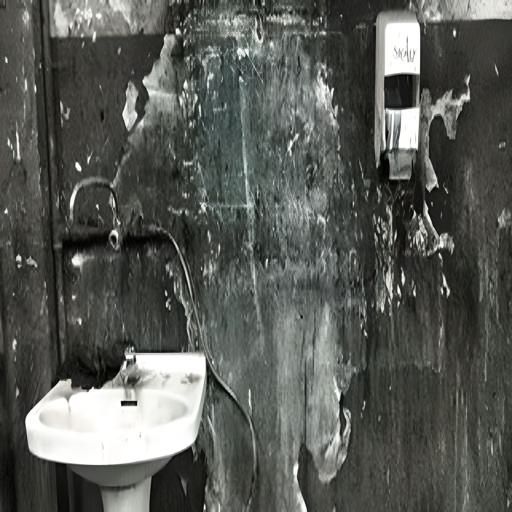} & \includegraphics[width=0.11\linewidth]{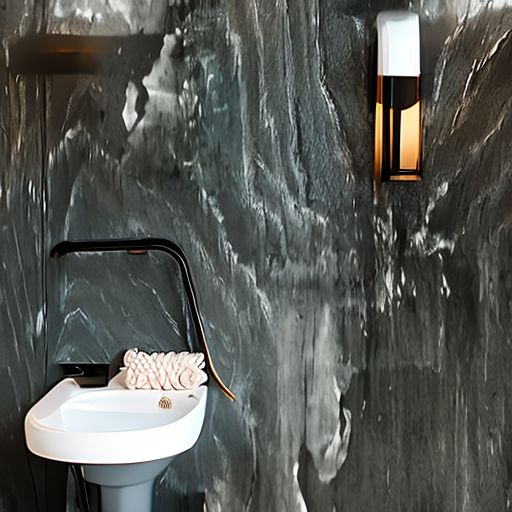} &  \includegraphics[width=0.11\linewidth]{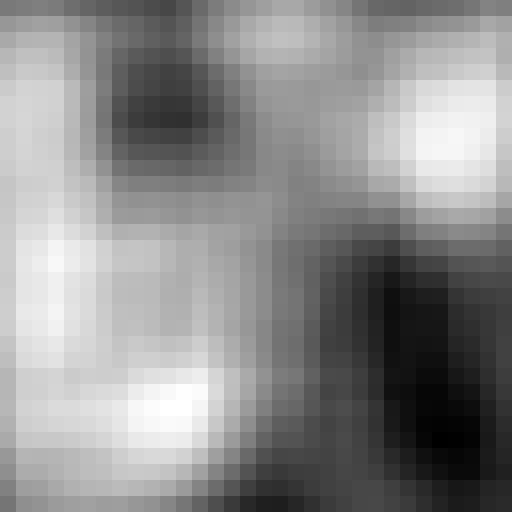} & \includegraphics[width=0.11\linewidth]{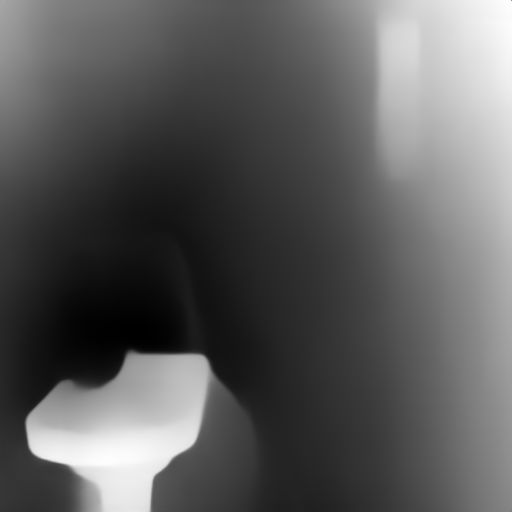} & 
    \includegraphics[width=0.11\linewidth]{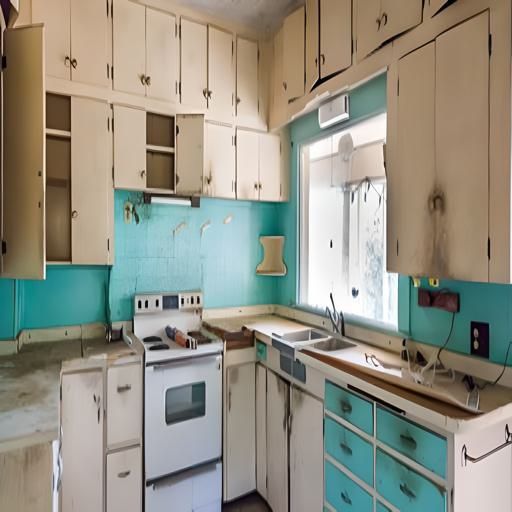} & \includegraphics[width=0.11\linewidth]{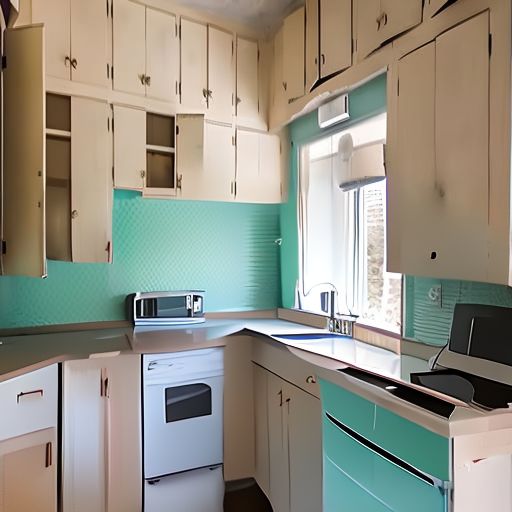} &  \includegraphics[width=0.11\linewidth]{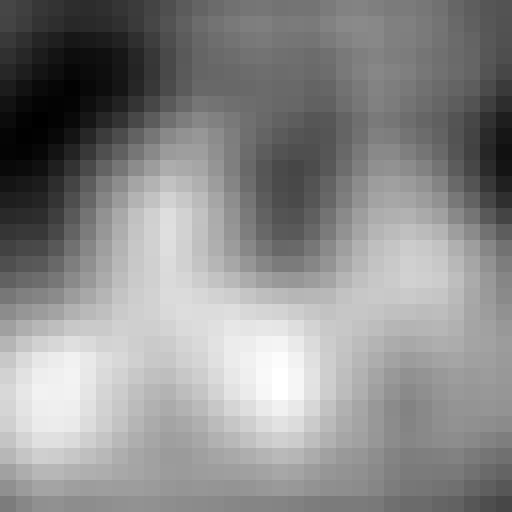} & \includegraphics[width=0.11\linewidth]{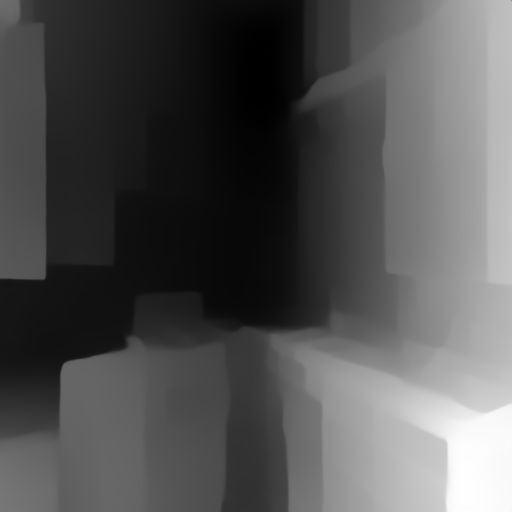} \\
    4.03 & 7.43 (+3.40) & & & 4.07 & 6.19 (+2.12) & & \\
    \includegraphics[width=0.11\linewidth]{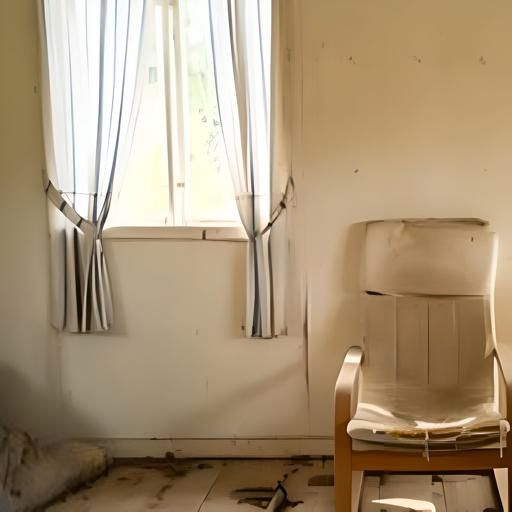} & \includegraphics[width=0.11\linewidth]{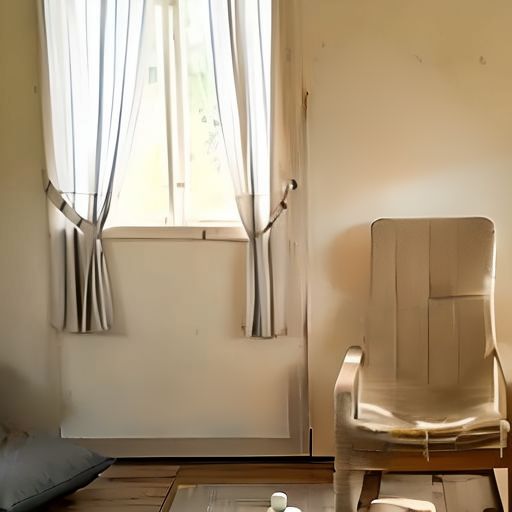} &  \includegraphics[width=0.11\linewidth]{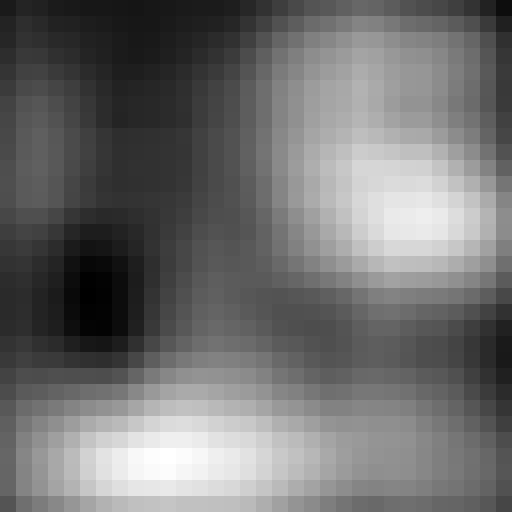} & \includegraphics[width=0.11\linewidth]{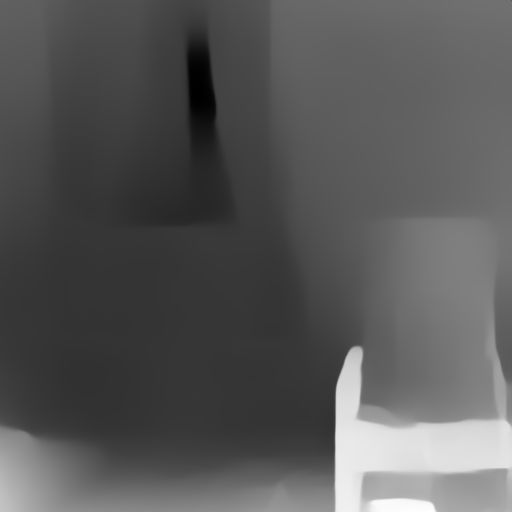} & 
    \includegraphics[width=0.11\linewidth]{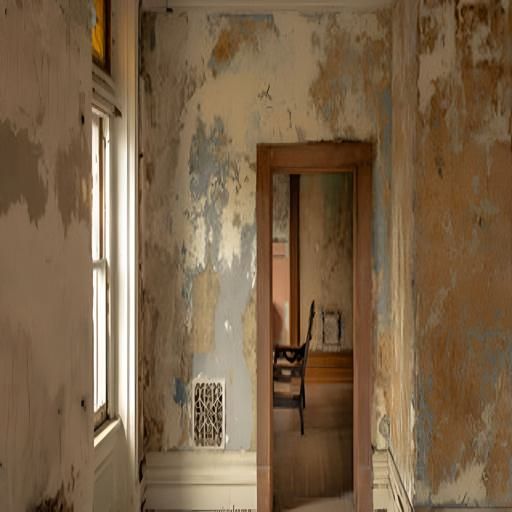} & \includegraphics[width=0.11\linewidth]{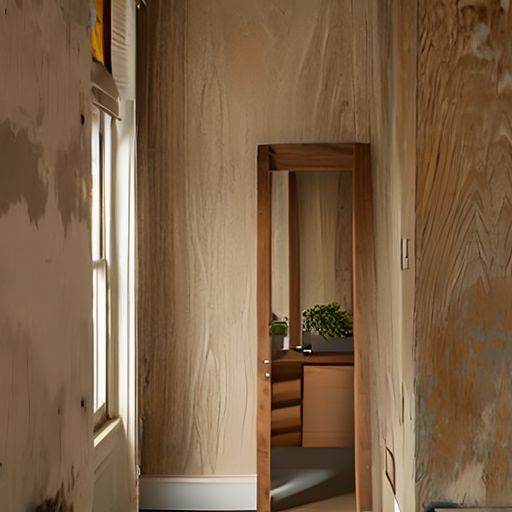} &  \includegraphics[width=0.11\linewidth]{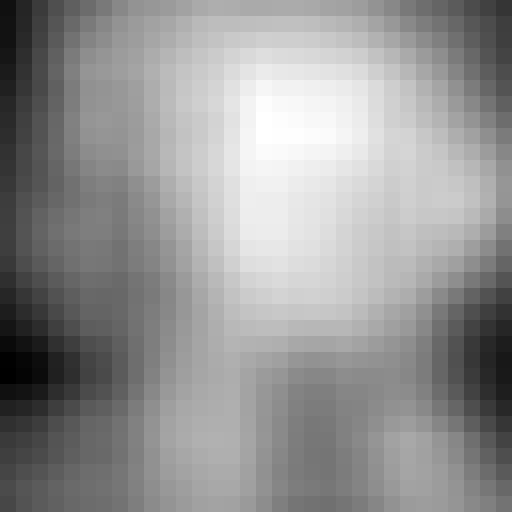} & \includegraphics[width=0.11\linewidth]{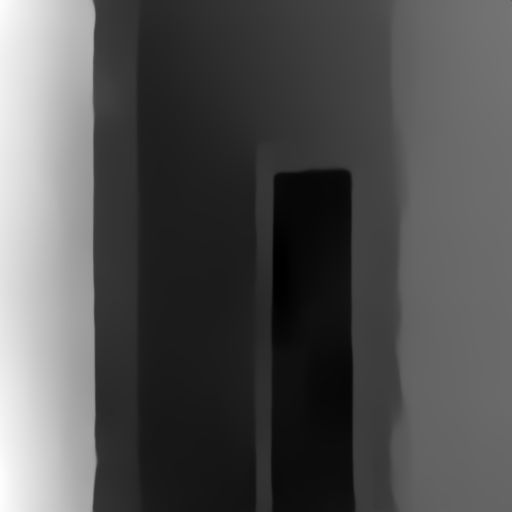} \\
    4.41 & 5.78 (+1.37) & & & 4.58 & 6.56 (+1.98) & & \\
    \end{tabular}
    \caption{\textbf{Image content appeal enhancement.} Corresponding to Fig. 9, we show images before/after enhancement (Col. 1/5 vs. Col. 2/6) with estimated appeal scores below each image. We use both the appeal heatmap $M_F^H$ (Col. 3/7) and the depth map (Col. 4/8) to guide the enhancement process.}
    \label{fig:appeal_enhancement_visual}
\end{figure*}

We present a comparative display of images before and after enhancement, accompanied by their content appeal scores as determined by our absolute appeal estimator (\cref{fig:appeal_enhancement_visual}). We also show the input image appeal heatmap and the estimated depth that guided the enhancement process. The visual and quantitative evidence from the increase in appeal scores clearly demonstrates that our methodology not only elevates the content appeal of images but also meticulously preserves the original color palette and structural integrity of the content.

\begin{figure}
\centering
\begin{tabular}{cccccc}
$ds =$ & 0.3 & 0.45 & \textbf{0.6} & 0.75 & 0.9 \\ 
 \includegraphics[width=0.13\linewidth]{figures_supp/more_baselines_comparison_figures/food_00020_input.jpeg}  & \includegraphics[width=0.13\linewidth]{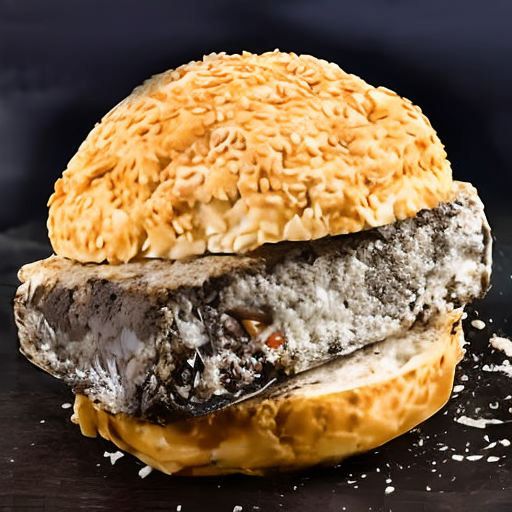} & \includegraphics[width=0.13\linewidth]{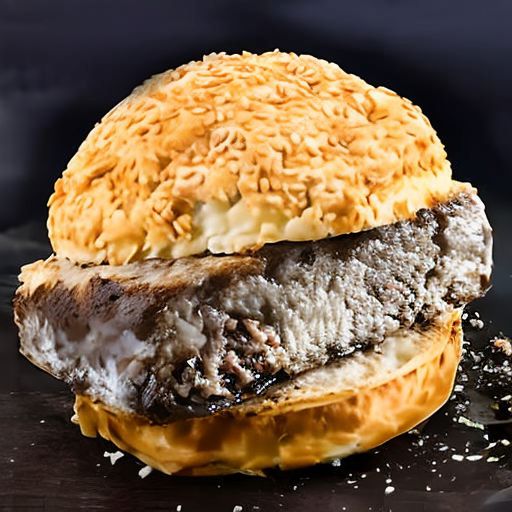} & \includegraphics[width=0.13\linewidth]{figures_supp/more_baselines_comparison_figures/food_00020_result.jpeg} & \includegraphics[width=0.13\linewidth]{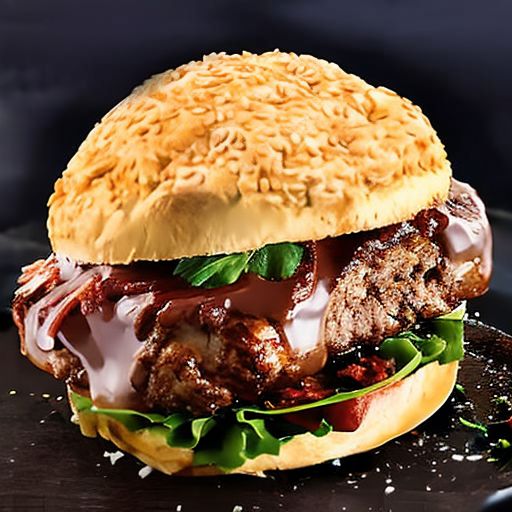} & \includegraphics[width=0.13\linewidth]{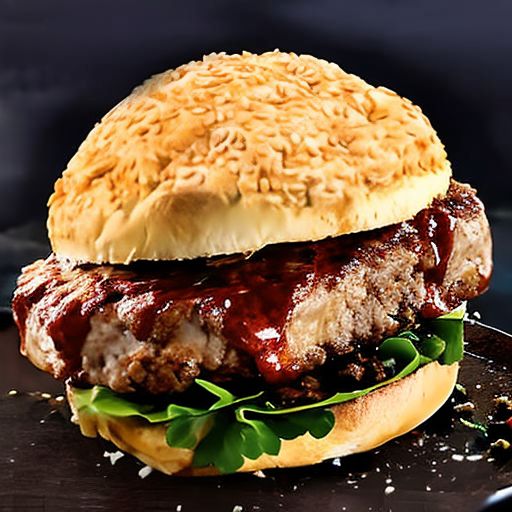} \\
    w/o $M_D^H$ & \includegraphics[width=0.13\linewidth]{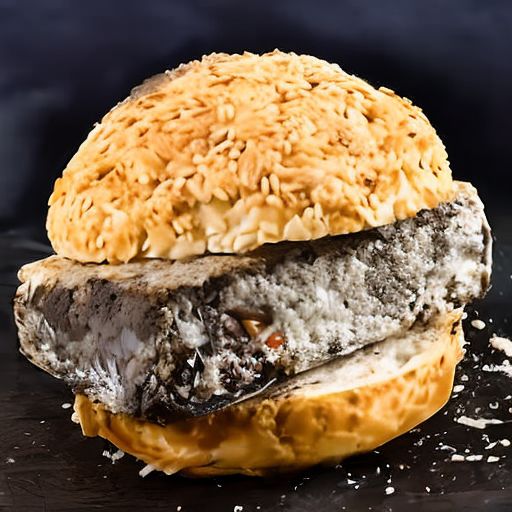} & \includegraphics[width=0.13\linewidth]{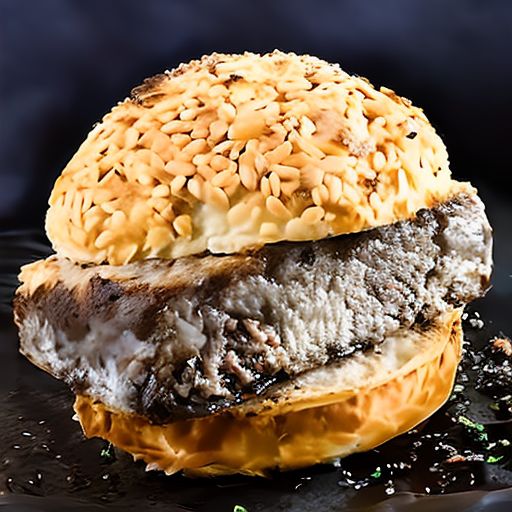} & \includegraphics[width=0.13\linewidth]{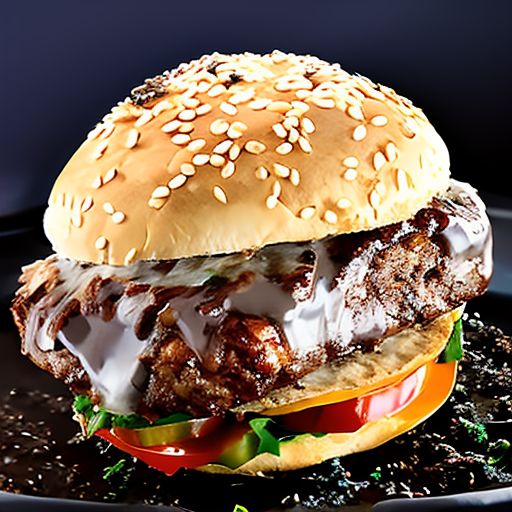} & \includegraphics[width=0.13\linewidth]{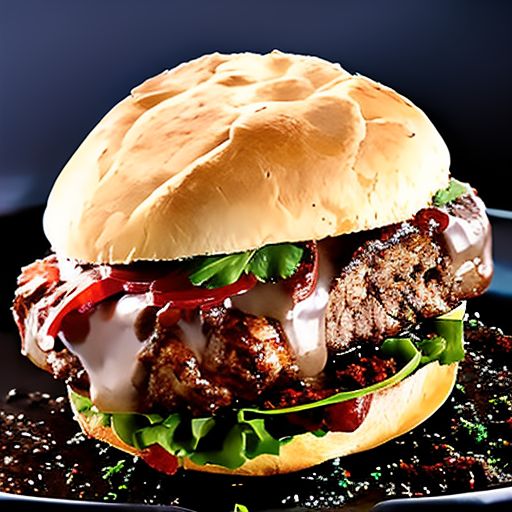} & \includegraphics[width=0.13\linewidth]{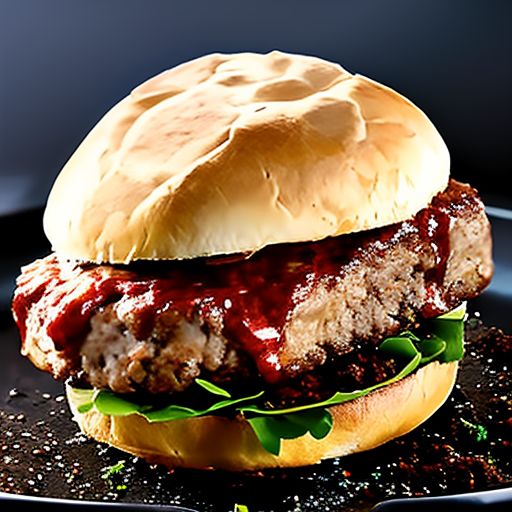} \\
    w/o depth & \includegraphics[width=0.13\linewidth]{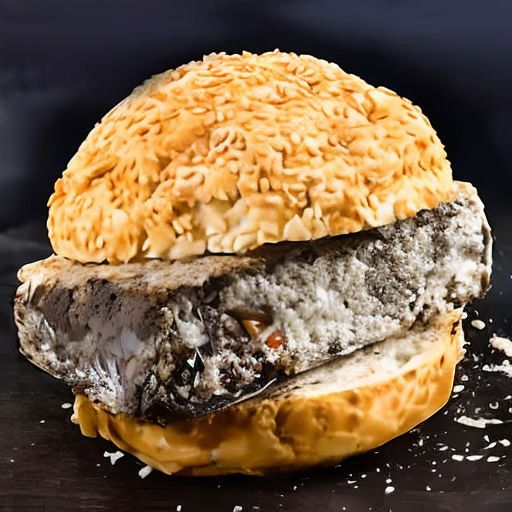} & \includegraphics[width=0.13\linewidth]{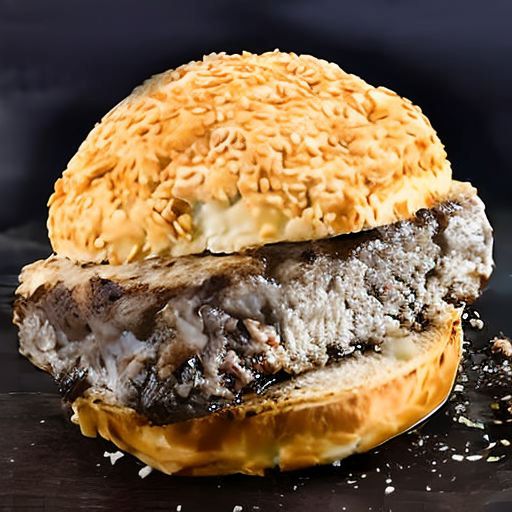} & \includegraphics[width=0.13\linewidth]{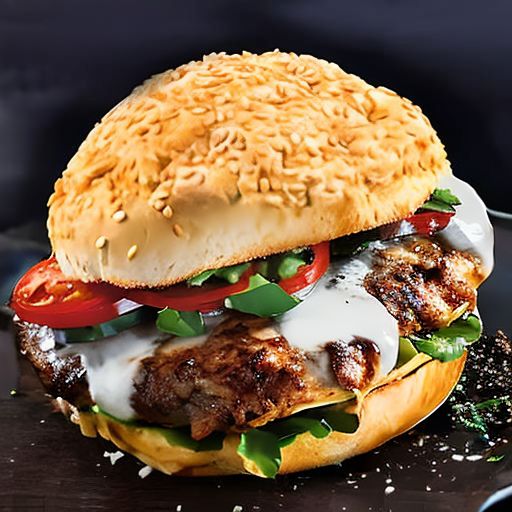} & \includegraphics[width=0.13\linewidth]{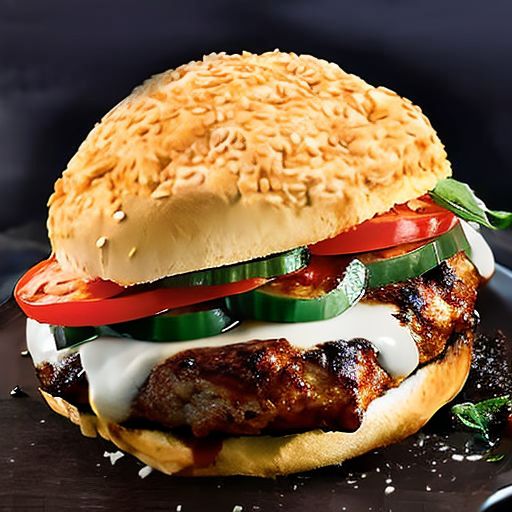} & \includegraphics[width=0.13\linewidth]{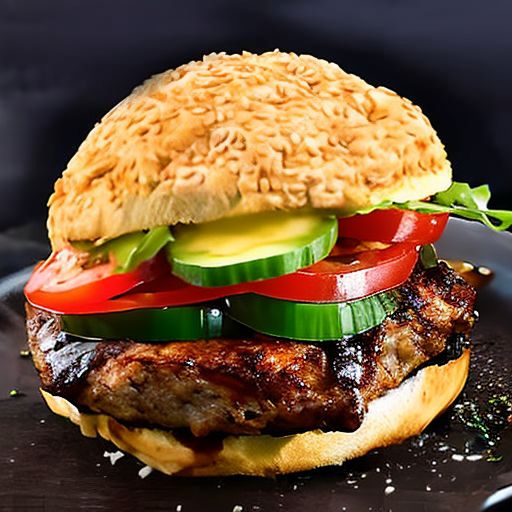} \\
    
\hline\hline

    $ds =$ & 0.3 & 0.45 & \textbf{0.6} & 0.75 & 0.9 \\ 
 \includegraphics[width=0.13\linewidth]{figures_supp/more_baselines_comparison_figures/room_00017_input.jpeg}  & \includegraphics[width=0.13\linewidth]{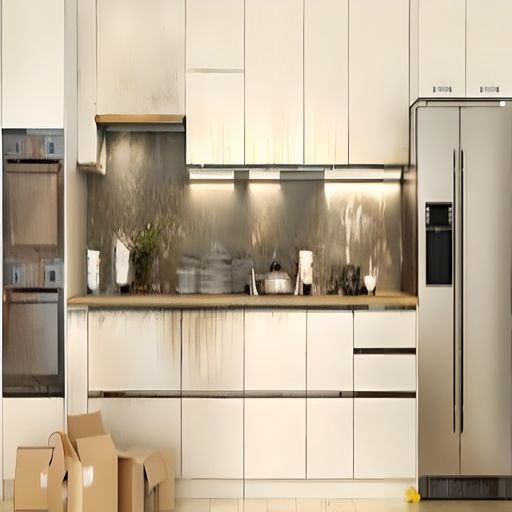} & \includegraphics[width=0.13\linewidth]{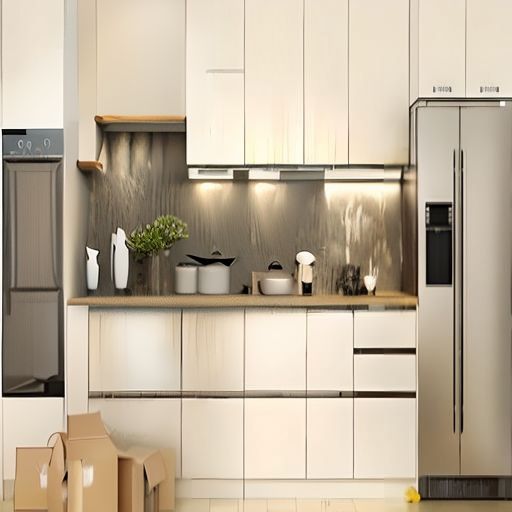} & \includegraphics[width=0.13\linewidth]{figures_supp/more_baselines_comparison_figures/room_00017_result.jpeg} & \includegraphics[width=0.13\linewidth]{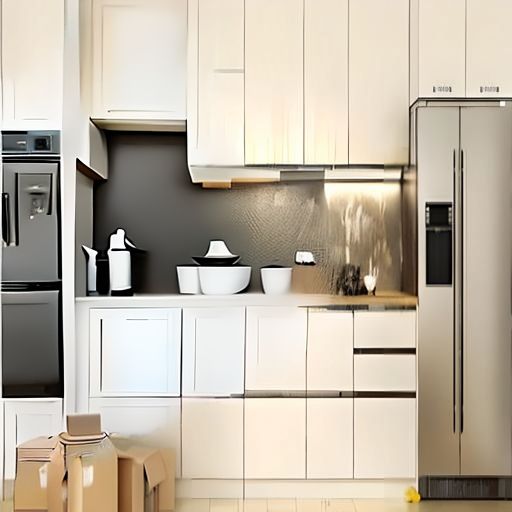} & \includegraphics[width=0.13\linewidth]{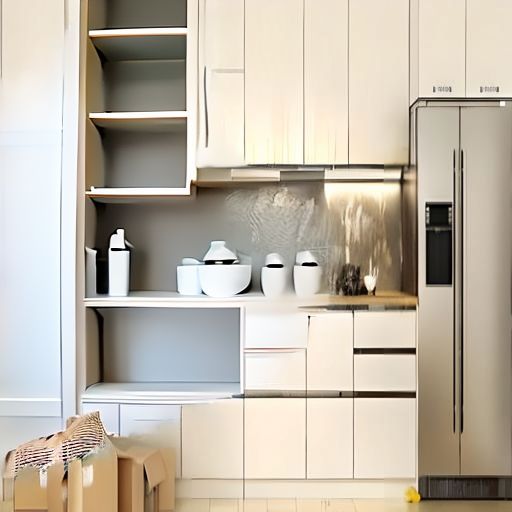} \\
    w/o $M_D^H$ & \includegraphics[width=0.13\linewidth]{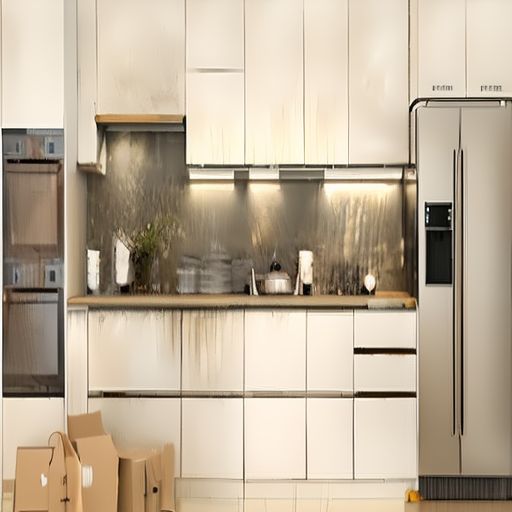} & \includegraphics[width=0.13\linewidth]{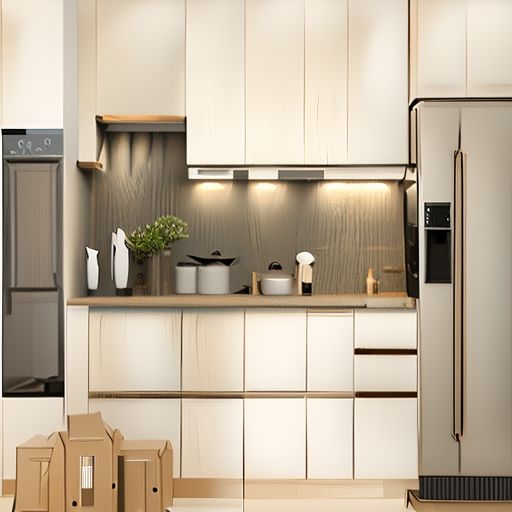} & \includegraphics[width=0.13\linewidth]{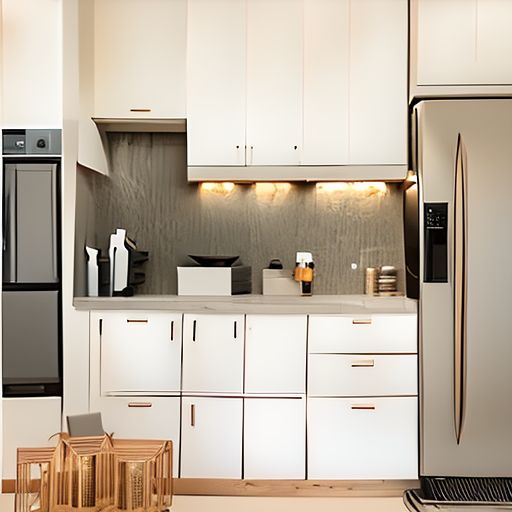} & \includegraphics[width=0.13\linewidth]{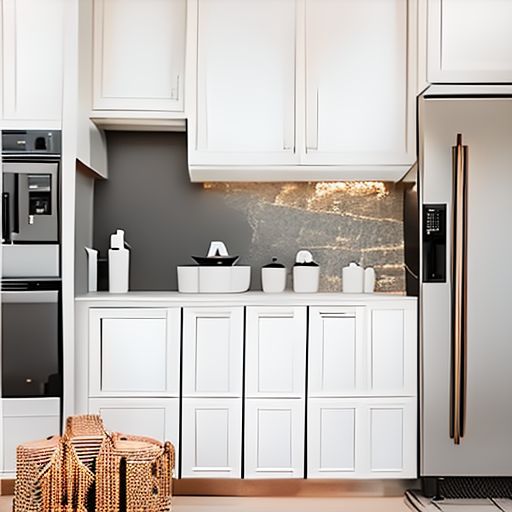} & \includegraphics[width=0.13\linewidth]{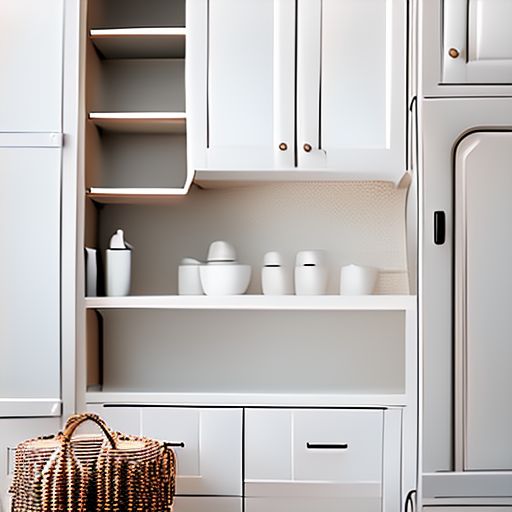} \\
    w/o depth & \includegraphics[width=0.13\linewidth]{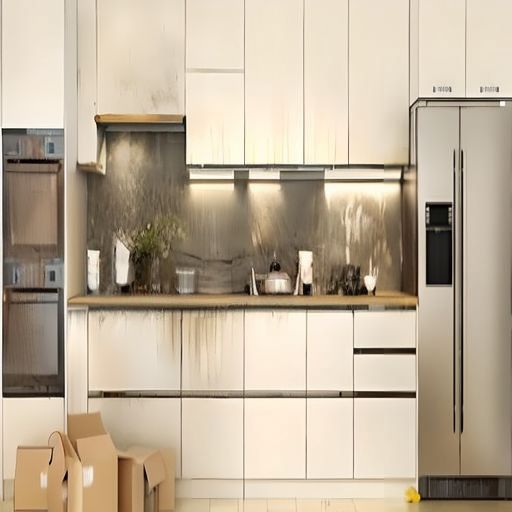} & \includegraphics[width=0.13\linewidth]{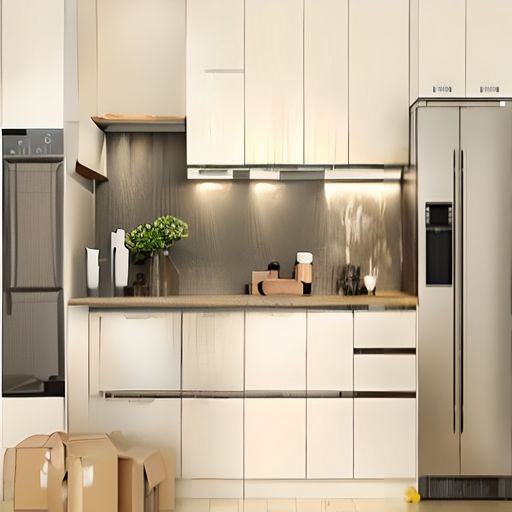} & \includegraphics[width=0.13\linewidth]{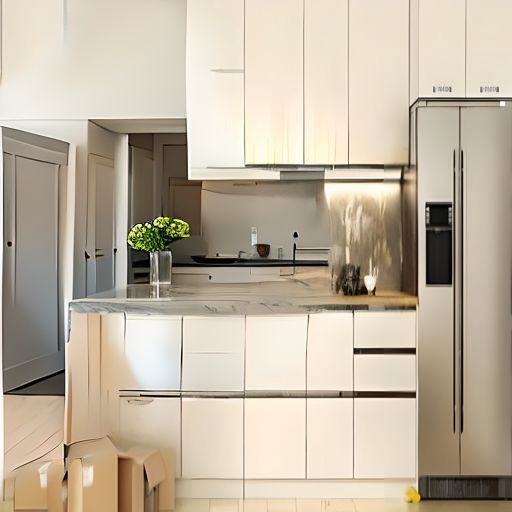} & \includegraphics[width=0.13\linewidth]{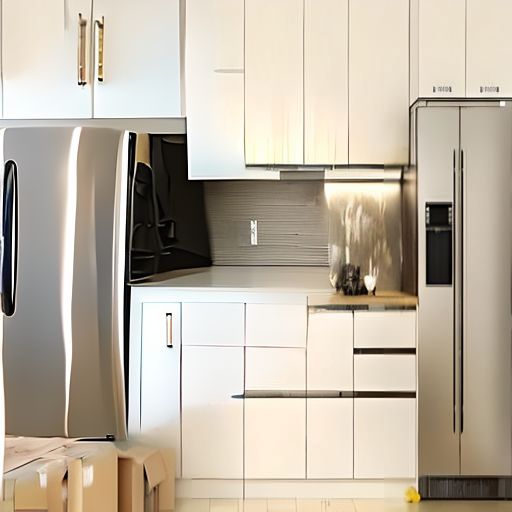} & \includegraphics[width=0.13\linewidth]{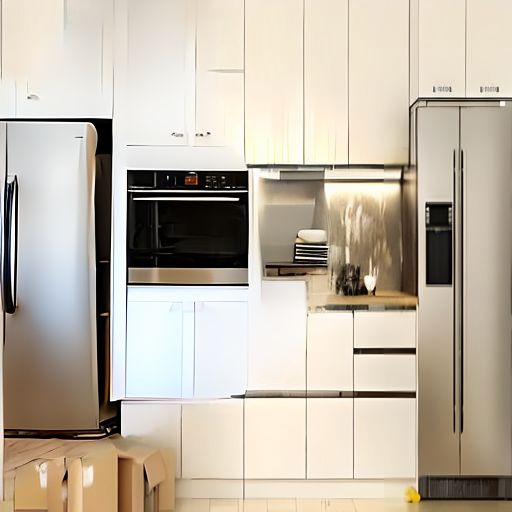} \\
\end{tabular}
\caption{\textbf{Effect of different denoising strength (ds) values, appeal heatmap, and depth on content appeal enhancement.} By enhancing the original image (the leftmost image in Cols.1 and 4 respectively) with different configurations, this analysis reveals that lower denoising strength values (e.g., 0.3, 0.45) result in marginal improvements in content appeal, indicating that such settings are insufficient for effective enhancement. Conversely, excessively high ds values (e.g., 0.75, 0.9) risk creating noticeable discontinuities in color and style between enhanced and non-enhanced areas, as delineated by the appeal heatmap $M_D^H$. Consequently, we opted for a denoising strength of 0.6 (highlighted in bold), balancing enhancement impact with visual coherence. Although omitting $M_D^H$ can ostensibly further augment overall content appeal, it also introduces undesired modifications, such as altering the appearance of the burger buns or cabinet drawers next to the fridge. Employing $M_D^H$ serves to mitigate unwarranted changes in color and structure, and the integration of a depth map further ensures the preservation of these attributes throughout the enhancement process.}
\label{fig:ablation}
\end{figure}

 We demonstrate the effect of different denoising strength, appeal heatmap $M_D^H$, and the depth map on the enhancement result in \cref{fig:ablation}, where lower denoising strength values (e.g., 0.3, 0.45) result in marginal improvements in content appeal, indicating that such settings are insufficient for effective enhancement. Excessively high denoising strength values (e.g., 0.75, 0.9) can cause noticeable color and style discontinuities between enhanced and non-enhanced areas, as shown by the appeal heatmap $M_D^H$. We chose a denoising strength of 0.6 to balance enhancement impact with visual coherence. Omitting $M_D^H$ can increase overall content appeal but may undesirably alter appealing objects. Using $M_D^H$ helps prevent unwanted changes, and incorporating a depth map ensures the preservation of these attributes during enhancement.

\subsection{Baselines details\label{sec:enhancement_baselines_details}}

We use the following text-guided localized image editing models as baselines for image enhancement comparisons:
\begin{itemize}
    \item InstructPix2Pix (IP2P): It takes text instructions as inputs to manipulate images. For food image, we use ``turn it into a delicious $[item]$,'' where $[item]$ is the name of the food in the image; for room images, we use ``turn it into a clean $[item]$,'' where $[item]$ is the name of the room in the image. In both cases, $[item]$ is parsed from the image text description generated by BLIP.
    \item Null-text Inversion (N-TI): This method takes an image and its text description as inputs, inverts the image based on the description, and allows edits by inserting new words or adjusting attention weights of existing words. We use BLIP to generate text descriptions of images. For editing, we decrease the attention weight of negative adjectives to -100 and insert positive adjectives like ``delicious,'' ``tasty,'' ``clean,'' or ``tidy,'' increasing their attention weight to 100. These values were set experimentally for optimal appeal improvement with minimal artifacts.
    \item pix2pix-zero (P2P-0): This method enables image manipulation using a specified edit direction. We generated two sets of 1,000 captions each for unappealing (burnt, moldy, rotten food) and appealing food images. The edit direction is the mean difference between the CLIP text embeddings of these sets. Similarly, for rooms, we created two sets of 1,000 captions describing unappealing (abandoned, dirty) and appealing (clean) rooms, following the same steps as for food images to define the edit direction.
    \item Text2LIVE (T2L): This method takes two prompts $(p_O, p_T)$ as inputs, where $p_O$ describes the input image and $p_T$ describes the target(desired) output. We take the search query that is used to retrieve the corresponding input image as $p_O$. For the \textsc{Food} dataset, we use $p_T = ``delicious [item]$; for the \textsc{Room} dataset, we use  $p_T = \text{``clean [item]''}$, where $[item]$ is obtained in the same manner as in IP2P. 
\end{itemize}

\begin{figure*}[t]
\centering
    \subfloat[]{\includegraphics[width=0.3\linewidth]{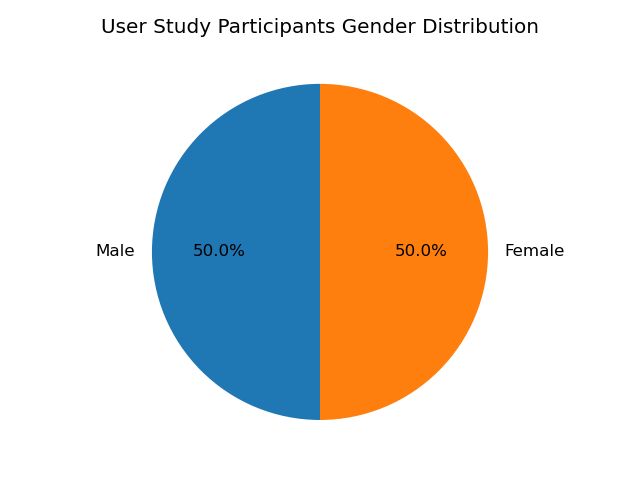}\label{fig:user_study_stats_gender}} \hfill
    \subfloat[]{\includegraphics[width=0.3\linewidth]{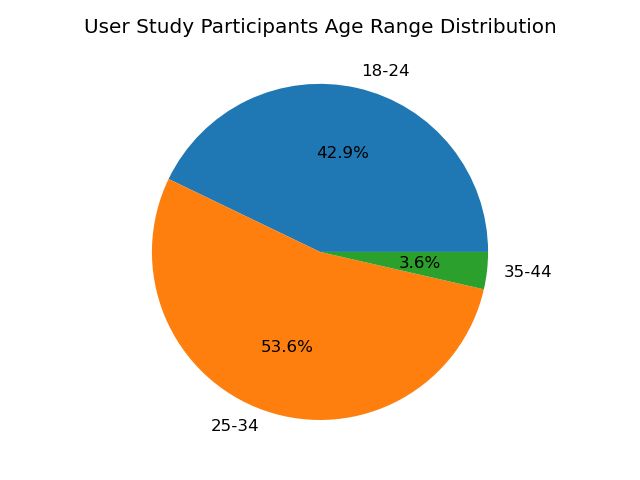}\label{fig:user_study_stats_age}} \hfill
    \subfloat[]{\includegraphics[width=0.3\linewidth]{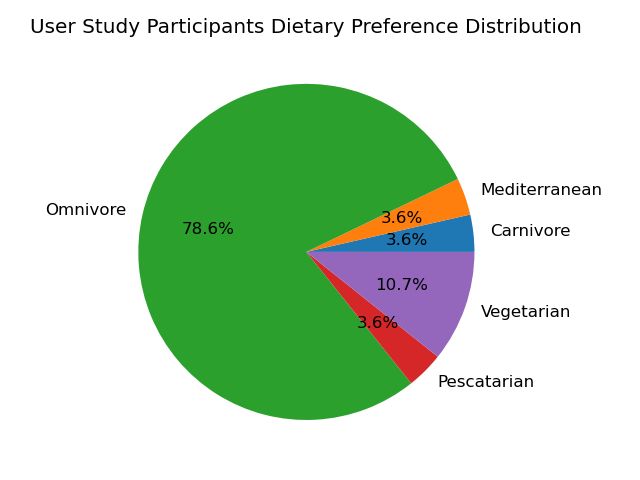}\label{fig:user_study_stats_dietary}}
    \caption{\textbf{User Study Questionnaire Answers Statistics.} Out of all the participants, there is an even split between males and females (\cref{fig:user_study_stats_gender}). The ages of most participants (27 out of 28; 96.4\%) are below 35, with 12 (42.9\%) of them aged between 18-24 and 15 (53.6\%) between 25-34 (\cref{fig:user_study_stats_age}). From \cref{fig:user_study_stats_dietary}, we can see that the majority of participants are omnivores (22 out of 28; 78.6\%); the second most common dietary preference among participants is Vegetarian (3 out of 28; 10.7\%).}
    \label{fig:user_study_stats}
\end{figure*}

\begin{figure*}
    \captionsetup[subfigure]{labelformat=empty}
    \subfloat[]{\includegraphics[width=0.19\linewidth]{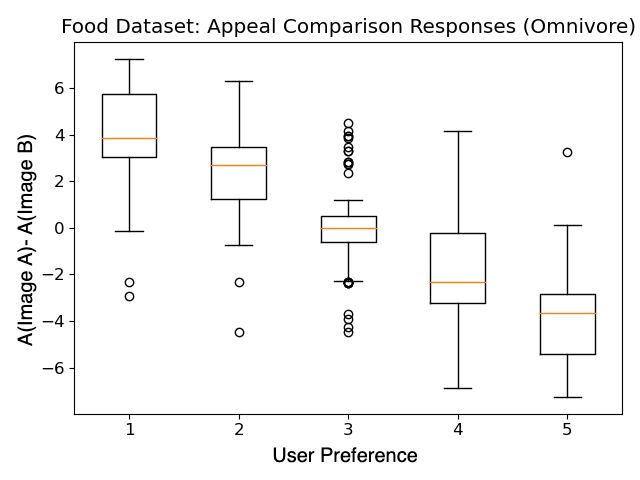}\label{fig:user_study_stats_omnivore_real}}
    \hfill
    \subfloat[]{\includegraphics[width=0.19\linewidth]{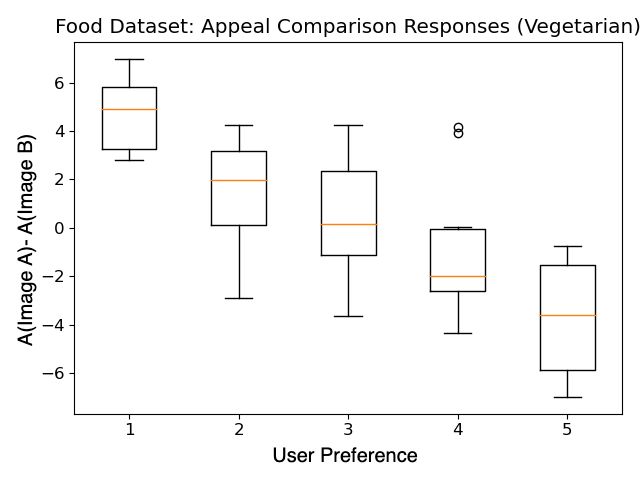}\label{fig:user_study_stats_vegetarian_real}}
    \hfill
    \subfloat[]{\includegraphics[width=0.19\linewidth]{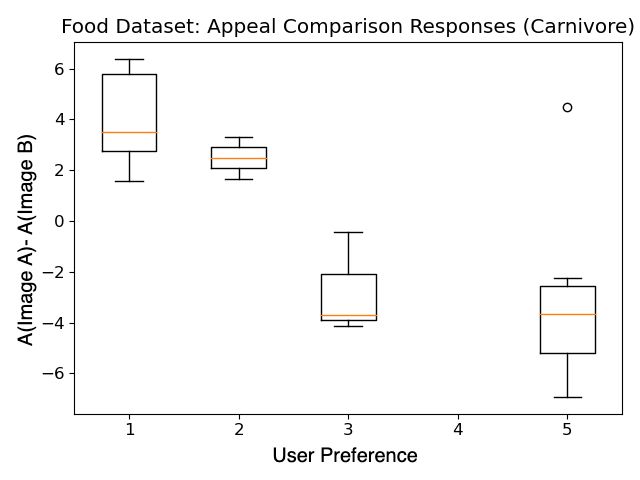}\label{fig:user_study_stats_carnivore_real}}
    \hfill
    \subfloat[]{\includegraphics[width=0.19\linewidth]{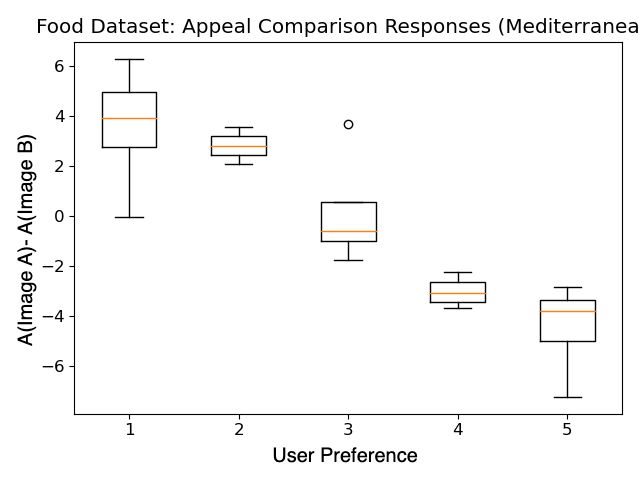}\label{fig:user_study_stats_mediterranean_real}}
    \hfill
    \subfloat[]{\includegraphics[width=0.19\linewidth]{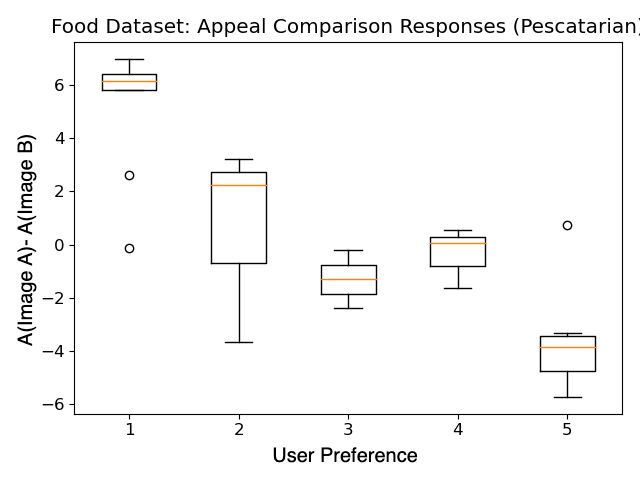}\label{fig:user_study_stats_pescatarian_real}}

    \subfloat[]{\includegraphics[width=0.19\linewidth]{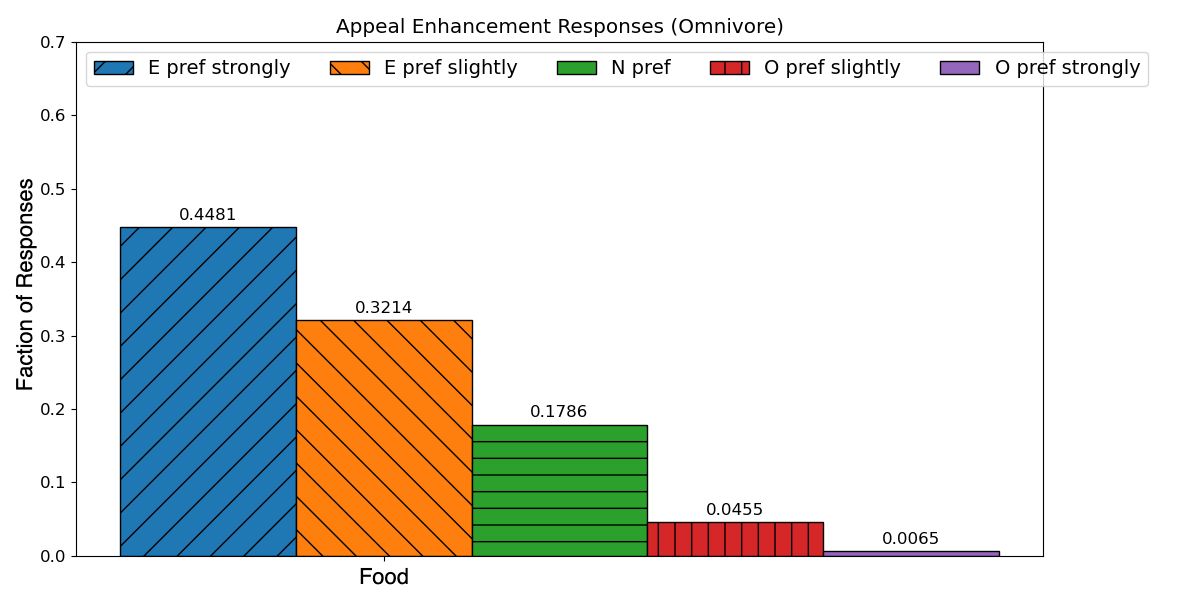}\label{fig:user_study_stats_omnivore_enhance}}
    \hfill
    \subfloat[]{\includegraphics[width=0.19\linewidth]{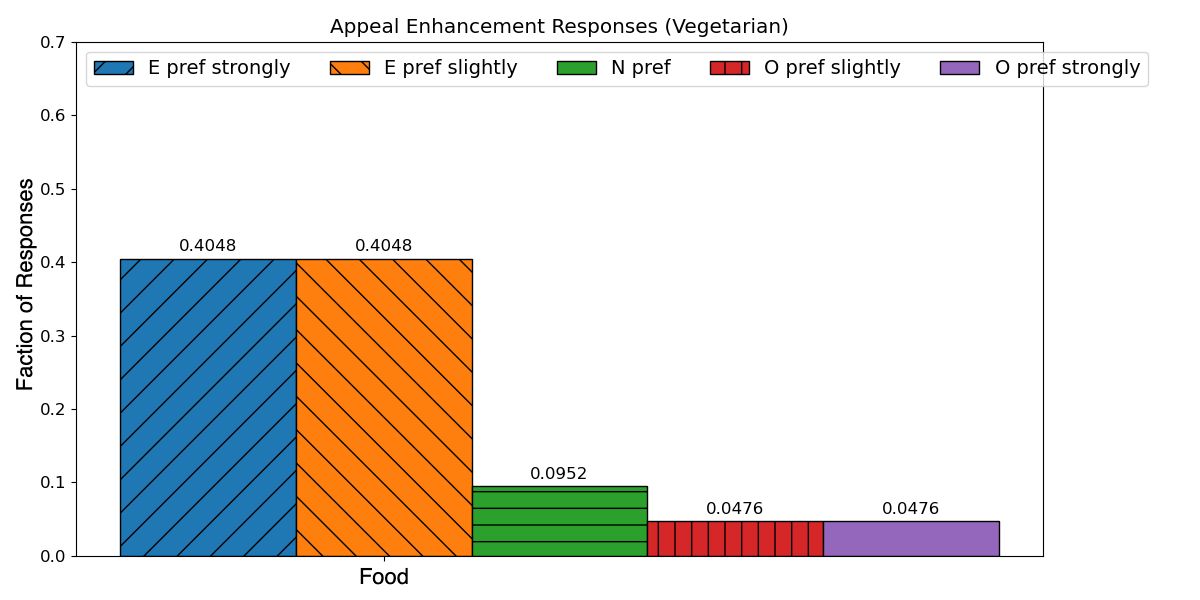}\label{fig:user_study_stats_vegetarian_enhance}}
    \hfill
    \subfloat[]{\includegraphics[width=0.19\linewidth]{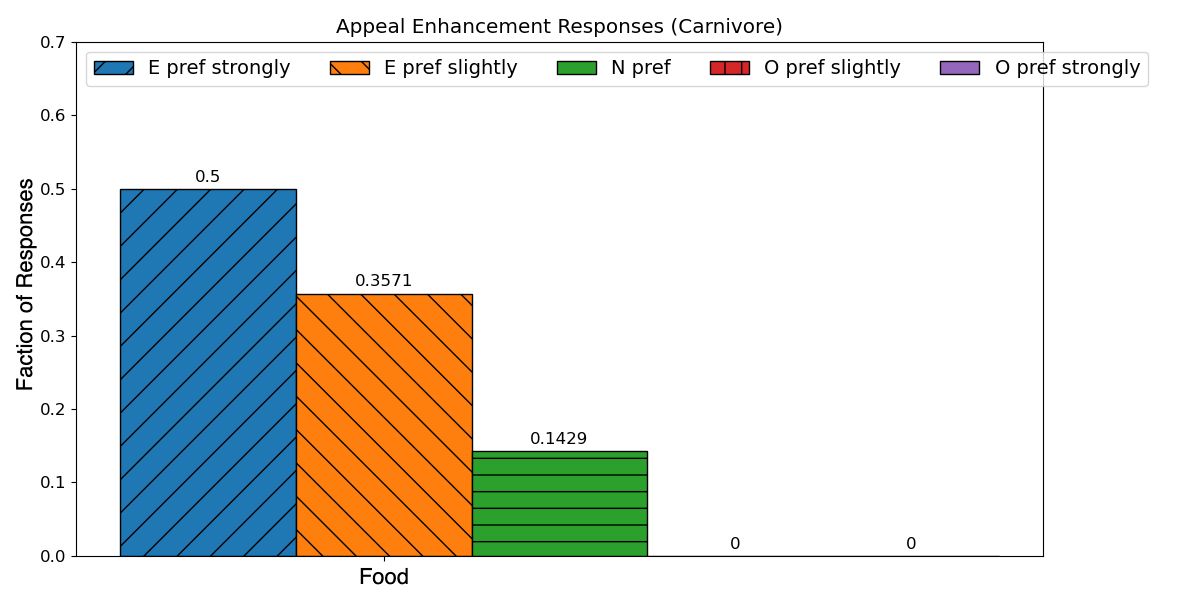}\label{fig:user_study_stats_carnivore_enhance}}
    \hfill
    \subfloat[]{\includegraphics[width=0.19\linewidth]{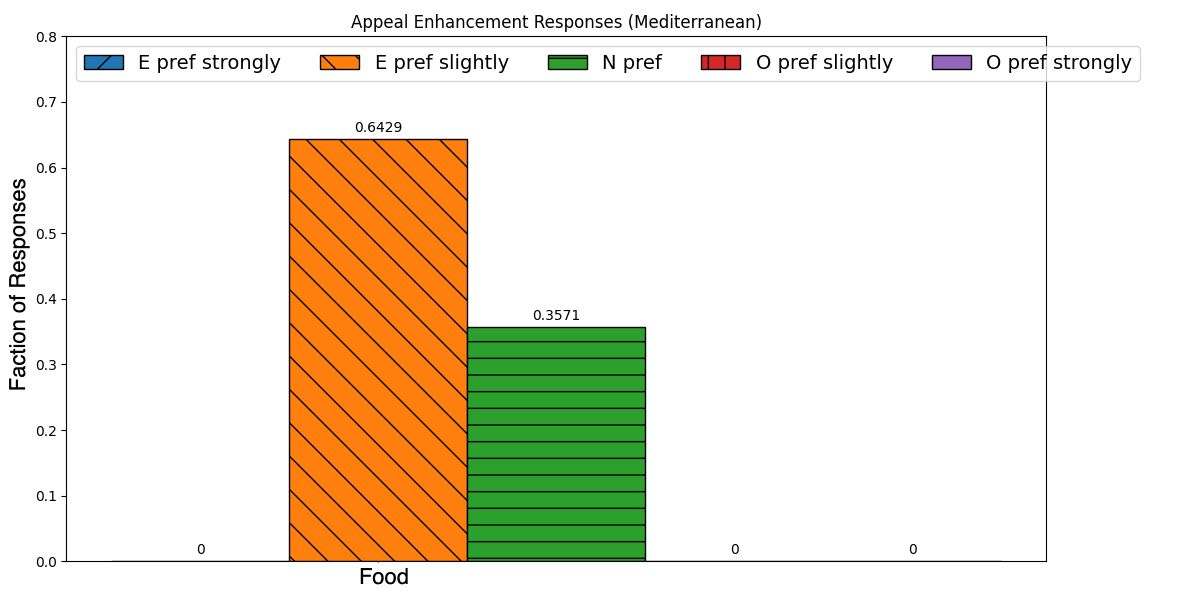}\label{fig:user_study_stats_mediterranean_enhance}}
    \hfill
    \subfloat[]{\includegraphics[width=0.19\linewidth]{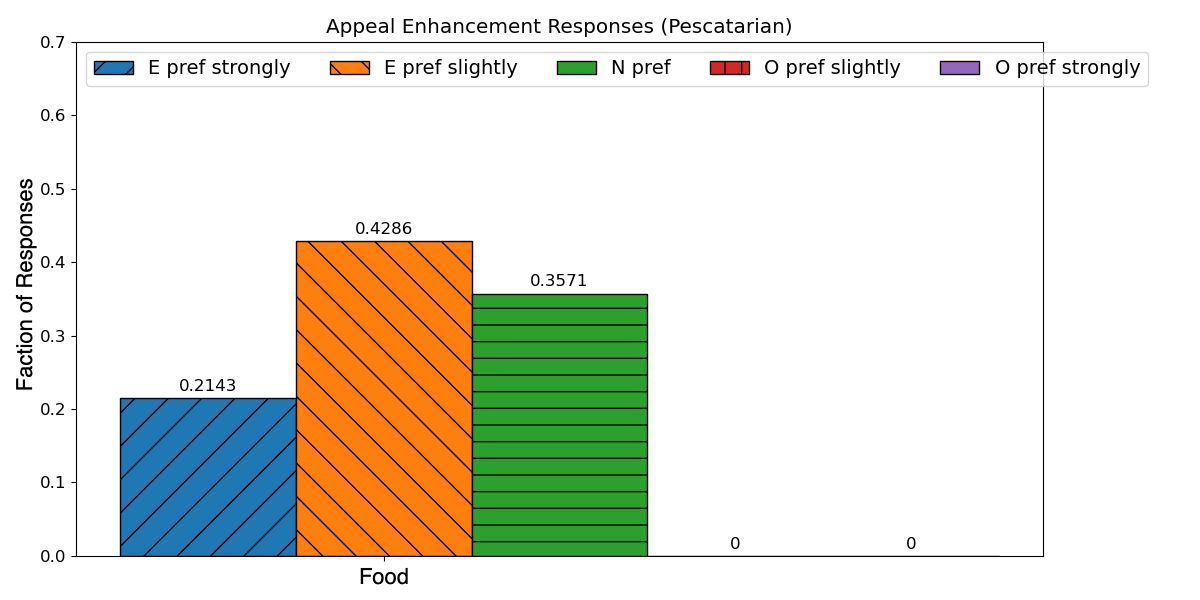}\label{fig:user_study_stats_pescatarian_enhance}}

    \caption{\textbf{Image Appeal Response Statistics by Dietary Preference.} Top row is the distribution of the appeal score difference for each of the five response options in the
user study. Bottom row is the percentage of image enhancement preference responses for each category, where E represents
the enhanced image, O is the original image, N is neither,
and “pref” stands for “is preferred.” From left to right are responses from participants whose dietary preference is Omnivore, Vegetarian, Carnivore, Mediterranean, and Pescatarian. We observe no major distribution change in responses across participants with different dietary preferences.}
    \label{fig:user_study_stats_by_dietary}
\end{figure*}

\section{User Study Questionnaire and Statistics\label{sec:supp_user_study}}

Here is the pre-survey questionnaire we ask participants to fill out:
\begin{itemize}
\item Gender: M/F/Other/Prefer not to say
\item Age range: 18-24, 25-34, 35-44, 45-54, 54+
\item Dietary preference: Vegan, Vegetarian,
    Omnivore, Carnivore, Mediterranean,
    Keto, Paleo, Other (please specify):
\end{itemize}

Out of all 28 participants, there is an even split between males and females (\cref{fig:user_study_stats_gender}). The ages of most participants (27 out of 28; 96.4\%) are below 35, with 12 (42.9\%) of them aged between 18-24 and 15 (53.6\%) between 25-34 (\cref{fig:user_study_stats_gender}). The majority of participants are omnivores (22 out of 28; 78.6\%); the second most common dietary preference among participants is Vegetarian (3 out of 28; 10.7\%). 

To see how participants' personal dietary preference may affect their responses, we visualize responses by dietary preference (\cref{fig:user_study_stats_by_dietary}), where we observe no major distribution change of user preference in terms of image appeal across participants with different dietary preference. This suggests that the question we ask in the user study, ``Which item in the image do you think the majority of the people would prefer'', helps leverage individual preference. 

\end{document}